\documentclass{article}

\usepackage[final]{neurips_2024}

\usepackage[utf8]{inputenc} 
\usepackage[T1]{fontenc}    
\usepackage{hyperref}       
\usepackage{url}            
\usepackage{booktabs}       
\usepackage{amsfonts}       
\usepackage{nicefrac}       
\usepackage{microtype}      
\usepackage{xcolor}         

\hypersetup{
    colorlinks=true,
    linkcolor=blue,
    filecolor=magenta,      
    urlcolor=cyan,
    pdfpagemode=FullScreen,
    citecolor=violet,
}

\usepackage{amsmath}
\usepackage{amssymb}
\usepackage{mathtools}
\usepackage{amsthm}
\usepackage{enumitem}

\usepackage{subcaption}
\usepackage{soul}
\usepackage{algorithm}
\usepackage{algpseudocode}

\usepackage{tcolorbox}
\usepackage{colortbl}
\newtcolorbox[auto counter, number freestyle={\noexpand\arabic{\tcbcounter}}]{mycolorbox}[3][]{%
    fonttitle=\bfseries,
    title=Example~#2~\thetcbcounter: #3,
    #1
}

\usepackage[capitalize]{cleveref}
\Crefname{figure}{Fig.}{Figs.}
\Crefname{tabular}{Tab.}{Tabs.}
\Crefname{algorithm}{Alg.}{Algs.}
\Crefname{section}{Sec.}{Secs.}
\Crefname{appendix}{App.}{Apps.}
\Crefname{proposition}{Prop.}{Props.}
\crefname{lemma}{Lemma}{Lemmas}
\crefname{observation}{Obs.}{Obs.}

\theoremstyle{definition}
\newtheorem{theorem}{Theorem}[section]

\theoremstyle{definition}

\theoremstyle{remark}

\usepackage[textsize=tiny]{todonotes}

\let\citet\citep 

\def\eqref#1{Eq.~\ref{#1}}

\def\1{\bm{1}}

\DeclareMathAlphabet{\mathsfit}{\encodingdefault}{\sfdefault}{m}{sl}
\SetMathAlphabet{\mathsfit}{bold}{\encodingdefault}{\sfdefault}{bx}{n}

\DeclarePairedDelimiterX\braket[2]{\langle}{\rangle}{#1\,\delimsize\vert\,\mathopen{}#2}

\newcommand{\squishlisttwo}{
 \begin{list}{$\bullet$}
  { \setlength{\itemsep}{1pt}
     \setlength{\parsep}{0pt}
    \setlength{\topsep}{0pt}
    \setlength{\partopsep}{0pt}
    \setlength{\leftmargin}{1em}
    \setlength{\labelwidth}{1.5em}
    \setlength{\labelsep}{0.5em} } }

\newcommand{\squishend}{
  \end{list}  }

\newcommand{\el}{\end{flushleft}}
\newcommand{\bl}{\begin{flushleft}}

\newcommand{\cI}{\mathcal{I}}

\newcommand{\cN}{\mathcal{N}}
\newcommand{\cO}{\mathcal{O}}

\newcommand{\cS}{\mathcal{S}}

\theoremstyle{definition}

\setcitestyle{numbers,square,comma}

\title{DETAIL: Task \underline{DE}mons\underline{T}ration \underline{A}ttribution for Interpretable \underline{I}n-context \underline{L}earning}

\author{%
  Zijian Zhou$^{13}$
  \quad
  Xiaoqiang Lin$^1$
  \quad
  Xinyi Xu$^{12}$
  \quad
  Alok Prakash$^3$ \\
  \textbf{Daniela Rus}$^{34}$
  \quad
  \textbf{Bryan Kian Hsiang Low}$^1$
  \\
  $^{1}$Department of Computer Science, National University of Singapore, Singapore \\
  $^{2}$Institute for Infocomm Research, A*STAR, Singapore \\
  $^{3}$Singapore-MIT Alliance for Research and Technology Centre, Singapore \\
  $^{4}$CSAIL, MIT, USA \\
  \texttt{\{zhzijian,xiaoqiang.lin,xuxinyi,lowkh\}@comp.nus.edu.sg} \\
  \texttt{\{zijian.zhou,alok.prakash\}@smart.mit.edu} \\
  \texttt{rus@csail.mit.edu}
}

\begin{document}

\maketitle

\begin{abstract}
In-context learning (ICL) allows transformer-based language models that are pre-trained on general text to quickly learn a specific task with a few ``task demonstrations''  without updating their parameters, significantly boosting their flexibility and generality. ICL possesses many distinct characteristics from conventional machine learning, thereby requiring new approaches to interpret this learning paradigm. Taking the viewpoint of recent works showing that transformers learn in context by formulating an internal optimizer, 
we propose an influence function-based attribution technique, \texttt{DETAIL}, that addresses the specific characteristics of ICL. We empirically verify the effectiveness of our approach for demonstration attribution while being computationally efficient. Leveraging the results, we then show how \texttt{DETAIL} can help improve model performance in real-world scenarios through demonstration reordering and curation. Finally, we experimentally prove the wide applicability of \texttt{DETAIL} by showing our attribution scores obtained on white-box models are transferable to black-box models in improving model performance.

\end{abstract}

\section{Introduction}

The rapid development of transformer-based language models~\citep{bommasani2022opportunities,brown2020,chowdhery2022palm,xu2024datacentricaiagelarge} has inspired a new in-context learning (ICL) paradigm~\citep{brown2020}, which allows a language model sufficiently pre-trained on general text to quickly adapt to specific tasks. This lightweight approach of customizing a general model for specific tasks is in contrast to fine-tuning~\citep{hu2022lora,xia2024less} that necessitates both access to the model parameters and resource-intensive step of tuning these parameters for model adaptation. In ICL, a few task demonstrations are included in the input text (i.e., prompt) together with a query to help the language model better understand how to answer the query. It has been shown that including task demonstrations in the prompt can enhance the capability of language models to apply common sense and logical reasoning~\citep{wei2023chainofthought,yao2023} and learn patterns from the supplied demonstrations~\citep{brown2020}, significantly enhancing the flexibility and generality of language models. In the ICL paradigm, each demonstration can be viewed as a ``training data point'' for ICL. Analogous to how the performance of a conventional supervised machine learning (ML) model depends on the quality of training data, the performance of ICL depends on the quality of task demonstrations~\citep{liu2023prompt}. 
A research question naturally arises: How to attribute and interpret ICL demonstrations that are helpful or harmful for model prediction?





Though there are many prior works on interpreting and attributing model prediction for conventional ML models~\citep{ghorbani2019,koh2017,lime2016}, these methods are not readily applicable to ICL due to its unique characteristics. Firstly, many existing attribution techniques require either computing the gradients~\citep{sundararajan17} or multiple queries to the model~\citep{cook1977}, both of which are slow and computationally expensive. In contrast, ICL is often applied in \emph{real-time} to a large foundation model~\citep{bommasani2022opportunities} that necessitates the attribution approaches for ICL to be fast and efficient. Secondly, ICL is known to be sensitive to ordering: The same set of demonstrations can result in significantly different model performance under different permutations~\citep{liu2023lost,lu2022fantastically}. However, conventional methods do not explicitly consider the ordering of training examples. Thirdly, ICL demonstration is usually supplied as a sentence comprising a sequence of tokens, rendering conventional token-level attribution methods ineffective, as they do not capture contextual information of each ICL demonstration~\citep{bahdanau2016neural,sundararajan17}. Lastly, ICL does not update model parameters, rendering conventional techniques that analyze model parameter change~\citep{koh2017} not applicable. Moreover, the absence of the need to update model parameters also allows a good attribution result for ICL to be transferable across different language models.

To address these challenges, we propose \texttt{DETAIL}, a novel technique that takes advantage of a classical attribution approach while tackling the unique characteristics of ICL. 
We adopt the perspective that transformers formulate an internal optimizer~\citep{akyürek2023learning,chen2024training,oswald2023,xie2022incontext} for ICL. Based on this internal optimizer, we design a method to understand the impact of each demonstration on the transformer's prediction. Notably, this approach allows us to leverage powerful existing analysis tools for transformer-based models in ICL, where otherwise the characteristics of ICL make applying these tools difficult.

Specifically, we describe an intuitive (re-)formulation of the influence function~\citep{koh2017}, a popular attribution method for conventional ML, on the internal optimizer and show that \texttt{DETAIL} addresses the challenges of computational cost, sensitivity to order, and attribution quality. Then, we empirically verify that our formulation can identify demonstrations helpful for model prediction and outlier demonstrations. Additionally, we apply our method to tasks with real-world implications including prompt reordering, noisy demonstration detection, and demonstration curation to show its effectiveness. We demonstrate that \texttt{DETAIL} achieves improved performance when applied to typical white-box large language models (LLMs). Furthermore, as many powerful LLMs are currently closed-source (thus black-box), we show that our \texttt{DETAIL} score obtained on a white-box LLM (e.g., Vicuna-7b~\citep{zheng2023judging}) exhibits transferable characteristics to the performance on a popular black-box model (ChatGPT).

\section{Related Work}

\paragraph{Understanding in-context learning.} Prior works have attempted to understand the ICL capability of transformer-based language models~\citep{ahn2023transformers,akyürek2023learning,bai2023transformers,bhattamishra2023understanding,brown2020,dai-etal-2023-gpt,garg2023transformers,li2023transformer,olsson2022incontext,panwar2024incontext,oswald2023,xie2022incontext,zhang2024icl}. \citet{brown2020} empirically demonstrated that language models can act as few-shot learners for unseen NLP tasks. \citet{olsson2022incontext} explained ICL by viewing attention heads as implementing simple algorithms for token sequence completion. \citet{panwar2024incontext,xie2022incontext} studied the ICL capability of transformers by casting it as an implicit Bayesian inference. \citet{akyürek2023learning,bhattamishra2023understanding,garg2023transformers} further showed that transformers can learn (regularized) linear and discrete functions in context. \citet{bai2023transformers} showed that transformers can adaptively implement appropriate algorithms for ICL. \citet{li2023transformer} provided a statistical generalization bound for ICL. \citet{dai-etal-2023-gpt,oswald2023,vonoswald2023uncovering,zhang2024icl} mathematically showed that transformers with specific parameters can perform gradient descent on parameters of an internal optimizer given the demonstrations. \citet{ahn2023transformers,zhang2024icl} further proved that the parameters (of the internal optimizer) can converge during forward passing. Inspired by the theoretical grounding, which is the focus of these works, we design our novel attribution technique for task demonstrations by adopting a similar view (i.e. transformers learn in context by implementing an optimization algorithm internally).

\paragraph{Data attribution.} Past works have focused on explaining and attributing model performance to training data of conventional ML~\citep{barshan20a,cook1977,ghorbani2019,grosse2023studying,koh2017,lin2024distributionally,lime2016,pmlr-v235-sim24a,xu2024datadistributionvaluation,zhou2022probablyapproximateshapleyfairness}. The rise of LLMs has inspired research efforts on attribution w.r.t.~prompts with a focus on task demonstrations~\citep{bohnet2023attributed,liu2023evaluating,machiraju2024prospector,inseq2023,yue2023automatic}, which is distinct from training data attribution since demonstrations are provided in context. Specifically, \citet{liu2023evaluating} used human annotators to evaluate the verifiability of attributing model answers to a prompt. \citet{bohnet2023attributed,yue2023automatic} relied on LLMs to evaluate attribution errors. These prior works are either computationally heavy (requiring additional queries of LLMs) or time-consuming (requiring human annotators). \citet{inseq2023} proposed an interpretability toolkit for sequence generation models using gradient- and perturbation-based methods. \citet{machiraju2024prospector}, a contemporary work, proposed to use a decoder module on the token embeddings for per-token attribution but requires costly training to learn the decoder weights.
Moreover, these methods do not specifically target demonstration attribution. Some prior techniques~\citep{cook1977,ghorbani2019} can be adapted for attributing ICL in LLMs but may be costly or ineffective. We empirically compare our method with those and the attribution methods consolidated in \citet{inseq2023}.

\paragraph{Attribution in LLMs.} Past works have attempted to apply various methods including influence in attributing language models~\citep{grosse2023studying,hu2024localizedzerothorderpromptoptimization,kwon2023datainf,lin2024promptoptimizationhumanfeedback,lin2024useinstinctinstructionoptimization,nguyen2023incontext,wang2024helpfulharmfuldatafinetuningfree,s2024incontext,wu2024promptoptimizationeaseefficient,xia2024less,zhang2024ideal}. \citet{nguyen2023incontext} considered a simplification of the influence function for task demonstration curation. \citet{grosse2023studying,kwon2023datainf,xia2024less} applied influence to pre-training and fine-tuning data of LLMs. \citet{zhang2024ideal} used influence to select demonstration inputs for annotation. \citet{s2024incontext} builds a classifier on the embeddings of demonstrations using a small LLM and computes influence w.r.t.~the classifier for demonstration selection. In contrast, we demonstrate various use cases of our method including on-the-fly demonstration curation, reordering, and noisy demonstration detection. 
A contemporary work that shares technical similarity~\citet{s2024incontext} focuses on demonstration selection whereas we focus on attribution and ~\citet{s2024incontext} is shown to be less effective than our method in \cref{sec:exp_application}. Additionally, compared to prior works leveraging influence to address specific problems, we apply influence function to provide a \textit{general attribution} for demonstrations, with many applications that we empirically show.




\section{Preliminaries}
\paragraph{In-context learning (ICL).} ICL is a learning paradigm that provides a few task demonstrations with formatted input and output for a pre-trained transformer (e.g., an LLM) to learn the mapping function from inputs to outputs in context (i.e., via the forward pass)~\citep{brown2020}. Formally, a transformer model $f$ takes in a prompt $p(\cS, x_{\text{query}})$ comprising a formatted sequence of demonstrations $\mathcal{S}=(z_1, z_2, \dots, z_n)$ where $z_i=\{x_i,y_i\}$ from a specific downstream task along with a query input $x_{\text{query}}$ (i.e., prompt) to predict its label as $\hat{y}_{\text{query}}=f(p(\cS, x_{\text{query}}))$. An visual for an example prompt is provided in \cref{colorbox:subj}. We wish to attribute the model prediction $\hat{y}_{\text{query}}$ to each $z_i \in \text{set}(\cS)$.

\paragraph{ICL as implementing an internal optimizer. } With the growing interest in the internal mechanism of transformers, previous works~\citep{akyürek2023learning,oswald2023,vonoswald2023uncovering,zhang2024icl} have theoretically shown that ICL can be treated as the transformer implementing an internal optimizer on the ICL demonstrations. Specifically, \citep{oswald2023,vonoswald2023uncovering,zhang2024icl} formulated the objective of ICL optimizer with (regularized) mean-squared error on a linear weight applied to the token itself in linear self-attentions (LSAs)~\citep{zhang2024icl} or a transformation of tokens (i.e., kernelized mean-squared error) if an extra multi-layered perception is attached before the LSA~\citep[Proposition 2]{oswald2023} in a recurrent transformer architecture. The transformer layers then function as performing gradient descent on the weight to minimize the objective~\citep{oswald2023,zhang2024icl}.



\paragraph{Influence function.} Influence function~\citep{koh2017} approximates the change of the loss of a test data point $z_{\text{test}}$ when up-weighting a training data point $z_i$. Formally, the influence of $z_i$ on predicting $z_{\text{test}}$ is\footnote{Following \citep{barshan20a} and the experiment implementation in \citep{koh2017}, we drop the negative sign in our influence definition.
The interpretation is that higher values imply a more positive impact.}
\begin{equation}
    \textstyle \cI(z_i, z_{\text{test}}) \coloneqq \nabla_{\theta}L(z_{\text{test}}, \hat{{\theta}})^\top
    \cI_{\text{reg}}(z_i) = \nabla_{\theta}L(z_{\text{test}}, \hat{{\theta}})^\top H^{-1}_{\hat{\theta}}\nabla_{\theta}L(z_i, \hat{\theta})
\label{eq:influence}
\end{equation}
where $L(z_{\text{test}}, \hat{\theta})$ refers to the loss (function) on a test point $z_{\text{test}}$ of the model parameterized by $\hat{{\theta}}$ and $H_{\hat{\theta}} \coloneqq 1/n \sum_{i=1}^n \nabla_{\theta}^2L(z_{i}, \hat{\theta})$ is the Hessian. However, this definition cannot be directly applied to ICL since there is no model parameter change during ICL, unlike the learning settings in \citep{barshan20a,koh2017}. We show how to adapt the formulation of \cref{eq:influence} to ICL in our proposed method \texttt{DETAIL} next.

\section{Influence Function on Internal Kernel Regression}



Following the idea that transformers learn in context by implementing an internal kernelized least-square objective, we present our formulation of \texttt{DETAIL} by computing the influence function on a kernelized linear regression~\citep{HainmuellerJens2014KRLS}.
Specifically, we build the regression w.r.t.~the following kernel 
\begin{equation}
   \textstyle k(x, x') \coloneqq m(x)^\top m(x')
\label{eq:kernel}
\end{equation}
where $m(x) \in \mathbb{R}^{1\times d}$ refers to (the mapping of an ICL demonstration\footnote{For LLMs, each demonstration may consist of more than $1$ token. We discuss how to address this in \cref{sec:eval_llm}.} to) an internal representation of $x$ (e.g., hidden state of a transformer layer) with output dimension $d$. Let $X \coloneqq (x_1, x_2, \cdots, x_n)$ and $Y \coloneqq (y_1, y_2, \cdots, y_n)$ be the vectors of inputs and outputs in $\cS$ respectively. The equivalent kernel regression can be written as $\textstyle \hat{Y} \coloneqq m(X) \beta$
where $\beta \in \mathbb{R}^{d\times 1}$ is the weight vector over the kernelized feature space. In practice, the dimension $d$ of $m$ is usually much larger than the number of demonstrations, causing severe over-parameterization. Such over-parameterization renders the influence values fragile~\citep{basu2021influence}. As such, we follow \citet{basu2021influence} and adopt an $\ell_2$ regularization on $\beta$ controlled by a hyper-parameter $\lambda$, which forms a \textit{kernelized ridge regression}~\citep{murphy22} with loss:
\begin{equation}
   \textstyle L(x, y) = [m(x) \beta - y]^2 + \lambda \beta^\top \beta\ .
    \label{eq:loss_ridge}
\end{equation}
Taking the $2$nd derivative of \cref{eq:loss_ridge}, we obtain the hessian $H_\beta$ as
\begin{equation}
   \textstyle H_{\beta} \coloneqq (1/n)\sum_{i=1}^n \nabla_{\beta}^2 L(x_i,y_i) = (2/n) \sum_{i=1}^n \left(m(x_i)^\top m(x_i) + \lambda I \right) \ .
    \label{eq:hessian}
\end{equation}
Adopting a matrix multiplication form for the summation in \cref{eq:hessian}, we write the influence of training data on the model parameters $\beta$ as follows,
\begin{equation}
  \textstyle  \cI_{\text{reg}}(z) \coloneqq H_{\beta}^{-1} \nabla_{\beta} L(x, y) = n(K + \lambda I)^{-1}[m(x)^\top(m(x) \beta - y) + \lambda \beta]
    \label{eq:infl_reg}
\end{equation}
where $K \coloneqq m(X)^\top m(X)\in \mathbb{R}^{d \times d}$ is the Gram matrix and $\cI_{\text{reg}} \in \mathbb{R}^d$ refers to the influence of a particular demonstration $(x,y)$ w.r.t.~the kernel regression weights $\beta$. Then, combining \cref{eq:influence,eq:loss_ridge,eq:infl_reg}, we can express the \texttt{DETAIL} score as the influence of a demonstration $z$ on a query $z_{\text{test}}$:
\begin{equation}
\begin{aligned}
  \textstyle  &\cI(z_{\text{test}}, z) \coloneqq \nabla_{\beta} L(x_{\text{test}}, y_{\text{test}})^\top \cI_{\text{reg}}(z) \\
    &= n[m(x_{\text{test}})^\top(m(x_{\text{test}}) \beta - y_{\text{test}}) + \lambda \beta](K + \lambda I)^{-1}[m(x)^\top(m(x) \beta - y) + \lambda \beta]
\end{aligned}
\label{eq:infl_kernel}
\end{equation}
where $\beta$ has a closed-form expression (shown in \cref{algo:infl} in \cref{app:algo}). 
While inverting matrices in \cref{eq:infl_kernel} requires $\cO(d^3)$ time, $d$ is usually in the thousands: A typical LLM like Llama-2-7b~\citep{touvron2023llama} has an embedding size $d=4096$, allowing reasonable computation time of $\cI$ (e.g., a few seconds). In practice, this computation is accelerated by the techniques already implemented in existing scientific computation libraries admitting sub-cubic complexity for matrix inversion.

\paragraph{Computing self-influence.} One important application of the influence function in ML is identifying outliers via self-influence~\citep{barshan20a,koh2017}. The conventional definition of $\cI$ trivially admits computing self-influence simply by replacing $z_{\text{test}}$ in \cref{eq:influence} with $z_i$. While the same approach applies to \texttt{DETAIL}, there are two shortcomings: (i) As the embedding is sensitive to the position, placing the same demonstration at the end of the prompt (as a query) or in the middle (as a demonstration) results in different embeddings, leading to unreasonable influence score. (ii) For each demonstration, it needs one forward pass of the model to compute the self-influence, which can be costly when the ICL dataset size is large. Instead, we implement self-influence for \texttt{DETAIL} by \textit{reusing} the demonstration's embedding. This way, we keep the two sides of \cref{eq:influence} consistent and only require one forward pass of the model to compute the self-influence for all demonstrations.

\paragraph{Further speedup via random matrix projection.} While the current formulation in \cref{eq:infl_kernel} is already computationally cheap, a relatively large embedding size (e.g. $d=4096$ for Llama-2-7b) can become a bottleneck as inverting the matrix can be relatively slow. We apply an insight that for ICL, much of the information in the embedding $m$ is redundant (we do not need a $4096$-dimensional $\beta$ to fit $20$ demonstrations). Hence, we project $m$ to a much lower dimensional space via a random matrix projection while preserving the necessary information, following the Johnson-Lindenstrauss lemma~\citep{dasgupta2003elementary,jllemma84}, precisely represented as a projection matrix $P \in \mathbb{R}^{d \times d'}$ with each entry i.i.d.~from $\cN(0, 1/d')$. We provide a more detailed discussion in \cref{app:discussion}. Empirically, we show that we can compress $m$ to a much smaller dimension $d' \leq 1000$, resulting in a $10\times$ computation speedup on a typical 7B LLM on an NVIDIA L40 GPU (see \cref{sec:eval_llm}). 

\begin{figure}[!ht]
    \centering
    \includegraphics[width=\textwidth]{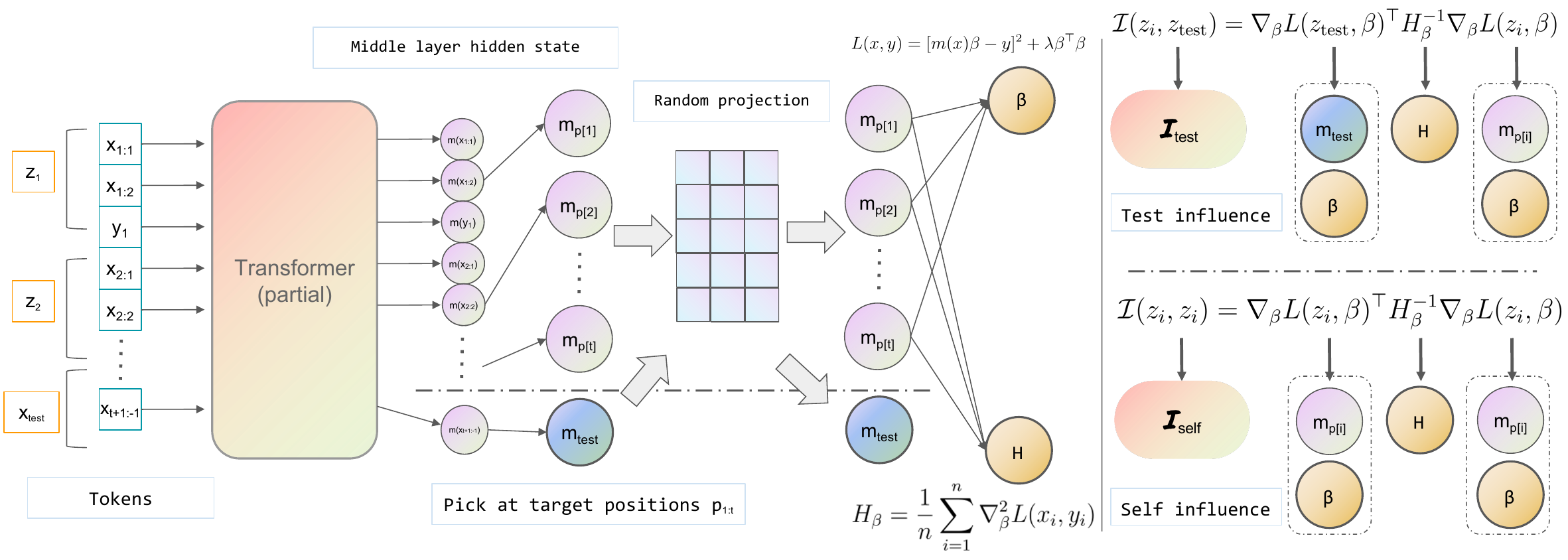}
    \caption{Illustration of computing \texttt{DETAIL} score for transformer-based ICL. Note that we use the same notation $m_{p[\cdot]}$ before and after the random projection since the projection is optional.}
    \label{fig:illustration}
\end{figure}

A visualization of our proposed method is in \cref{fig:illustration} and a pseudo-code implementation is in \cref{app:algo}.

\section{Empirical Evaluation}

We evaluate the effectiveness of \texttt{DETAIL} (i.e., \cref{eq:infl_kernel}) on two metrics: computational time (via logged system time required) and effectiveness in attribution (via performance metrics for tasks). We start by visualizing how the \texttt{DETAIL} scores, particularly $\cI(z_{\text{test}}, z_i)$ (test influence, abbreviated as $\cI_{\text{test}}$) and $\cI(z_i,z_i)$ (self influence, abbreviated as $\cI_{\text{self}}$) following \citep[Sections 5.1 \& 5.4]{koh2017}, attribute demonstrations to a query first on a custom transformer and then on LLMs. Note that the hyper-parameter $\lambda$ varies under different scenarios and we discuss some heuristics for setting $\lambda$ in \cref{app:discussion}.

\paragraph{Enforcing ICL behavior. } We consider tasks where transformers learn from the demonstrations and form an internal optimizer. To evaluate the effectiveness of our method, we enforce the ICL behavior by mapping the labels of the demonstrations to a token that carries no semantic meaning, This way, pre-trained transformers cannot leverage memorized knowledge to produce answers but have to learn the correct answer from the supplied demonstrations. Specifically, we map all labels to one of $\{A,B,C,D,E\}$ depending on the number of classes. More details are in each section.

\subsection{Evaluation on a Custom Transformer}
\label{sec:eval_custom_tf}

\begin{figure}[ht]
\centering
\begin{subfigure}[t]{0.6\textwidth}
    \includegraphics[width=\textwidth]{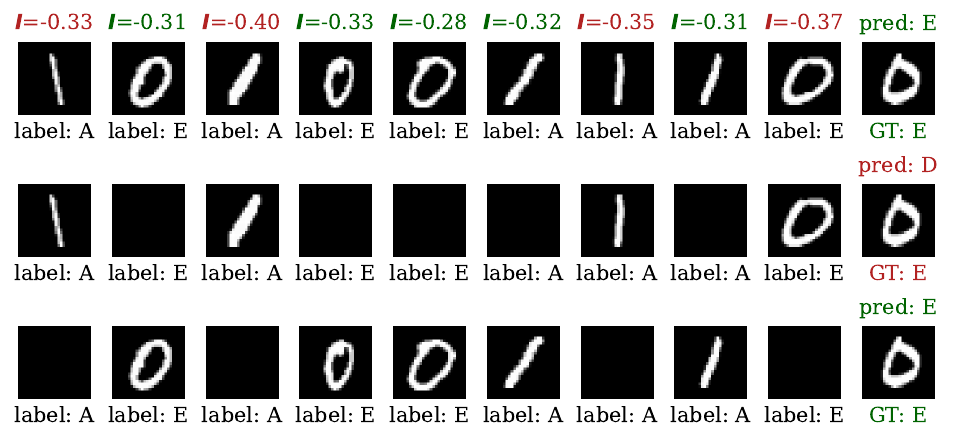}
    \end{subfigure}
\hfill
\begin{subfigure}[t]{0.38\textwidth}
    \includegraphics[width=\textwidth]{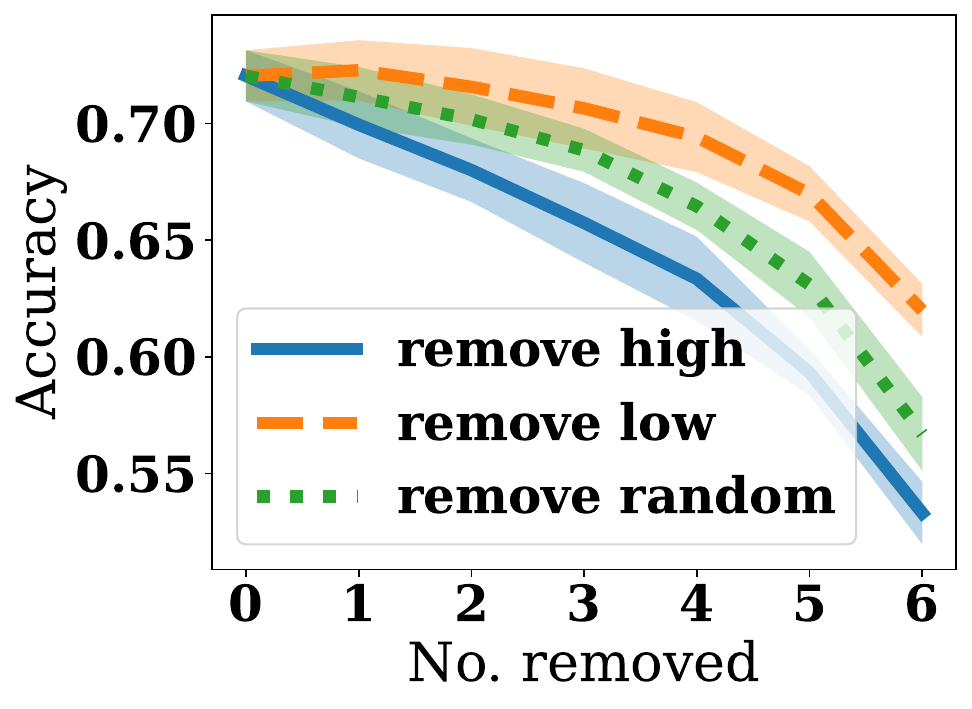}
    \end{subfigure}
\hfill
\caption{(Left) Visualization of learning label mapping of MNIST digits in context. The left $9$ images in each row are demonstrations while the \textit{right-most} one is a query image. Below each image shows its mapped label (``A'' to ``E''). Above each ICL image is its $\cI_{\text{test}}$ w.r.t.~the query image with high values highlighted in \textcolor{green!50!black}{green} and low values highlighted in \textcolor{red!70!black}{red}. Above the query image is the prediction (pred) made by the pre-trained transformer which is in \textcolor{green!50!black}{green} if consistent with the ground truth (GT) and \textcolor{red!70!black}{red} otherwise. Top row shows that using all $9$ demonstrations allows the transformer to learn the mapping in context as GT$=$pred$=$``E''. Middle shows removing $5$ demonstrations with the highest $\cI_{\text{test}}$ results in most digit $0$'s removed, leading to a wrong prediction. Bottom shows removing $5$ demonstrations with the lowest $\cI_{\text{test}}$ results in $3$ digit $0$'s remaining for the transformer to learn in context, leading to correct prediction. (Right) Average accuracy on $1013$ ICL datasets repeated over $10$ trials; $\lambda=0.01$; Lines and shades represent mean and standard error over $10$ independent trials.}
\label{fig:mnist}
\end{figure}

We use the MNIST dataset~\citep{deng2012mnist} to visualize how $\cI_{\text{test}}$ can be applied to attribute model prediction to each demonstration. We design a task where the transformer needs to learn a mapping of the digit image to a label letter in context. Specifically, for each image of $28 \times 28$ pixels, we flatten the pixels and concatenate them with a mapped label to form a $785$-dimensional token vector. For simplicity of illustration, we only use images of digits in $\{0, 1\}$. For each ICL dataset, we assign to each distinct digit a letter in $\{A,B,C,D,E\}$ randomly. We build a recurrent transformer based on the design in \citet{oswald2023} with $10$ recurrent layers each consisting of $15$ attention heads. We pre-train the transformer with random label mappings from randomly selected digits so that the transformer cannot memorize the mapping but has to infer from the demonstrations. We use the hidden state after the $1$st layer as $m$ to compute $\cI_{\text{test}}$. A qualitative visualization of $\cI_{\text{test}}$'s attribution and a quantitative plot showing how the test prediction varies by removing demonstrations with the highest (lowest) $\cI_{\text{test}}$ are in \cref{fig:mnist}. Left shows that removing tokens (represented by the image pixels and the corresponding label) with the largest $\cI_{\text{test}}$ makes the model make wrong predictions, whereas removing tokens with the lowest $\cI_{\text{test}}$ can retain the correct model predictions. Right shows that removing tokens with the lowest $\cI_{\text{test}}$ results in a slower decrease in prediction accuracy than removing the highest $\cI_{\text{test}}$, demonstrating that tokens with the highest $\cI_{\text{test}}$'s are more helpful and \textit{vice versa}.

\subsection{Evaluation on Large Language Models}
\label{sec:eval_llm}

With the insight obtained in \cref{sec:eval_custom_tf}, we now apply \texttt{DETAIL} to full-fledged LLMs. We start with demonstration perturbation and noisy demonstration detection to demonstrate that \texttt{DETAIL} can be used to interpret the quality of a demonstration. Then, leveraging these results, we further show how we can apply the \texttt{DETAIL} scores to tasks more closely related to real-world scenarios.

A distinction between LLMs and the custom transformer used above is that demonstrations in LLMs are usually \textit{a sequence of tokens}, whereas demonstrations in the custom transformer are single tokens representing the actual numerical values of the problems (see \cref{fig:mnist}). This distinction makes it difficult to find an appropriate internal representation for each demonstration (i.e., $m$). To overcome this challenge, we draw inspiration from prior works~\citep{wang-etal-2023-label,xie2022incontext} which suggest that information flows from input words to output labels in ICL. As an implementation detail, we take the embedding of the \textit{last token before the label token} (hereafter referred to as the target position) in the \textit{middle layer} where most of the information has flown to the target token positions. We include ablation studies on using different layers' embeddings in \cref{app:exp_layer} and using different target positions in \cref{app:exp_target_pos}.

\paragraph{Setup.} 
We consider (for white-box models) mainly a Vicuna-7b v1.3~\citep{zheng2023judging} and also a Llama-2-13b~\citep{touvron2023llama} on some tasks using greedy decoding to show that \texttt{DETAIL} works on models with different training data and of varying sizes. While our theoretical backing stands for transformer-based models, we experiment \texttt{DETAIL} on Mamba-2.8b~\citep{gu2023mamba}, a state-space model architecture that has received increased attention recently in \cref{app:exp_ssm}.
We primarily evaluate our method on AG News ($4$ classes)~\citep{Zhang2015CharacterlevelCN}, SST-2 ($2$ classes)~\citep{socher-etal-2013-recursive}, Rotten Tomatoes ($2$ classes)~\citep{pang05a}, and Subj ($2$ classes)~\citep{conneau2018senteval} datasets which all admit classification tasks. Due to space limits, some results are deferred to \cref{app:exp}.

\begin{figure}[!ht]
\centering
\begin{subfigure}[t]{0.22\textwidth}
    \includegraphics[width=\textwidth]{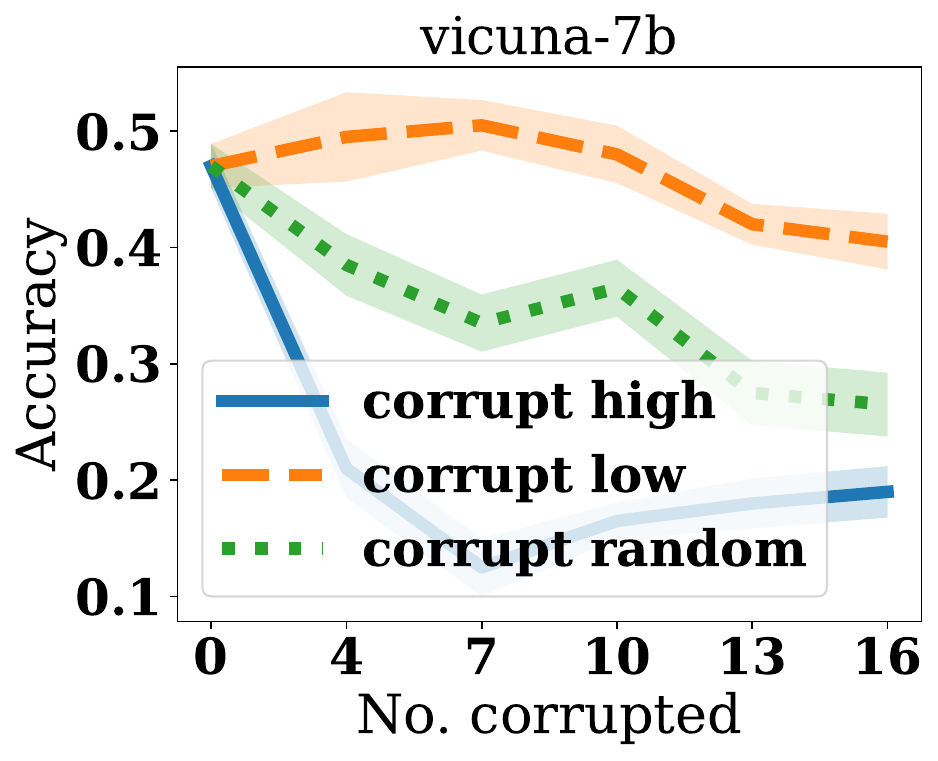}
    \end{subfigure}
\hfill
\begin{subfigure}[t]{0.22\textwidth}
    \includegraphics[width=\textwidth]{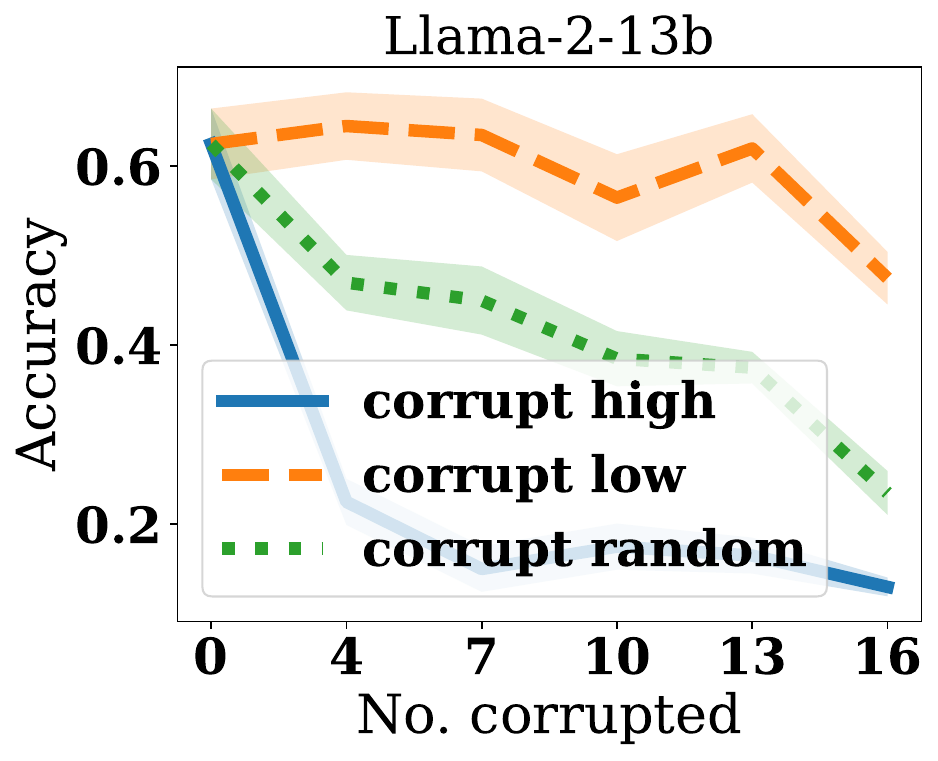}
    \end{subfigure}
\hfill
\begin{subfigure}[t]{0.22\textwidth}
    \includegraphics[width=\textwidth]{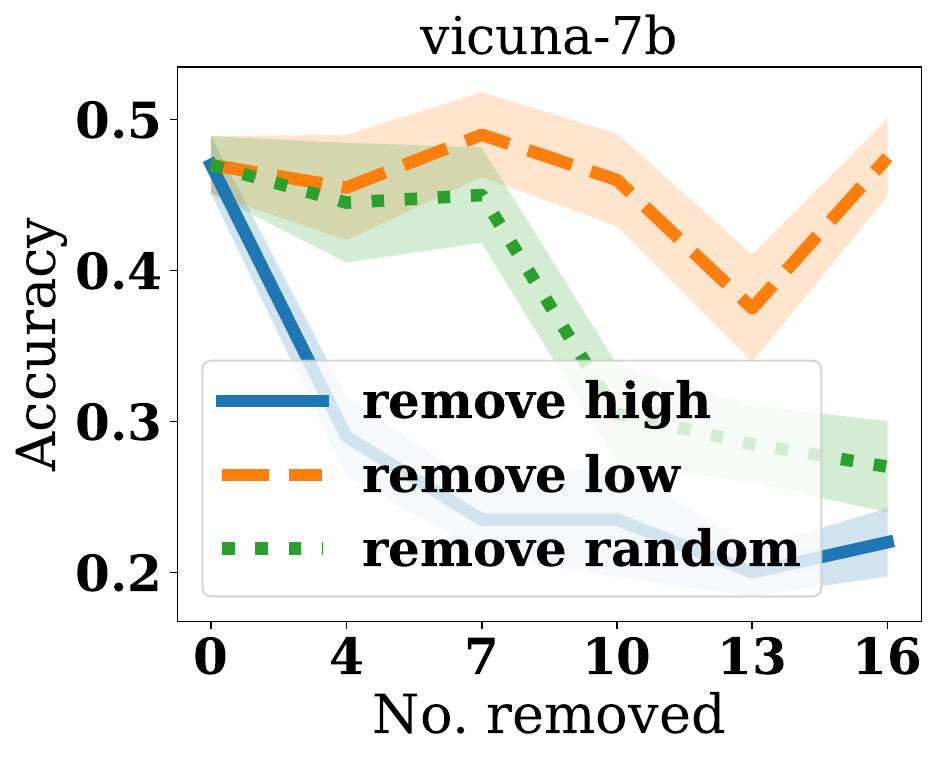}
    \end{subfigure}
\hfill
\begin{subfigure}[t]{0.22\textwidth}
    \includegraphics[width=\textwidth]{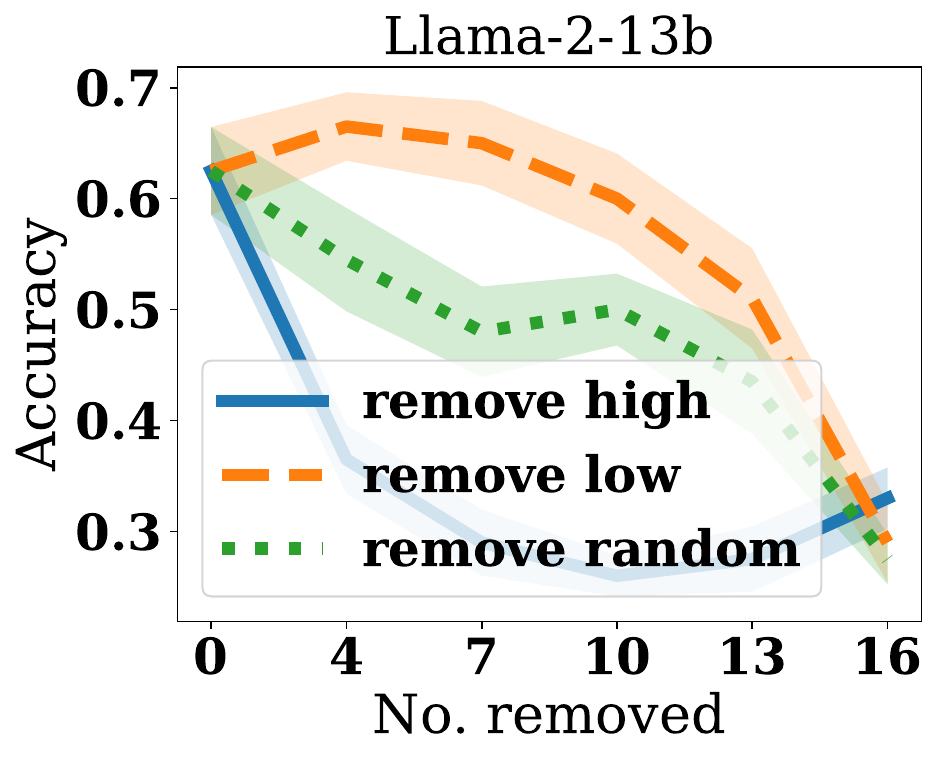}
    \end{subfigure}
\hfill
\caption{($1$st and $2$nd) Corrupting labels of demonstrations and ($3$rd and $4$th) removing demonstrations  with high/low \texttt{DETAIL} scores ($\cI_{\text{test}}$) on AG News. Perturbing demonstrations randomly result in an accuracy in the middle as expected. All experiments are repeated $10$ trials. $\lambda=1.0$. Lines and shades represent the mean and standard error respectively.}
\label{fig:llm_label_perturbation}
\end{figure}

\paragraph{Demonstration perturbation.} We show that \texttt{DETAIL} can explain LLM predictions by showing how perturbation (i.e., corrupting the labels of some demonstrations to an incorrect class or removing some demonstrations) with the high/low $\cI_{\text{test}}$ affects the model's predictive power. We randomly pick $20$ ICL datasets each comprising $20$ demonstrations and $1$ query from AG News and find the average and standard error of the accuracy of predicting the query after perturbation using Vicuna-7b and Llama-2-13b, shown in \cref{fig:llm_label_perturbation} (results for other datasets deferred to \cref{app:exp_sample_perturb_additional}). It can be observed that perturbing demonstrations with low $\cI_{\text{test}}$ results in a slower drop (or even improvement) in accuracy and \textit{vice versa}, similar to the trend observed in \cref{fig:mnist}, showcasing the applicability of \texttt{DETAIL} to LLMs. We additionally include the results using Falcon-7b~\citep{almazrouei2023falcon} and Llama-2-7b~\citep{touvron2023llama} in \cref{app:exp_sample_perturb_additional} where we perturb $10$ demonstrations and observe a similar accuracy gap.

\begin{figure}[ht]
\centering
\begin{subfigure}[t]{0.23\textwidth}
    \includegraphics[width=\textwidth]{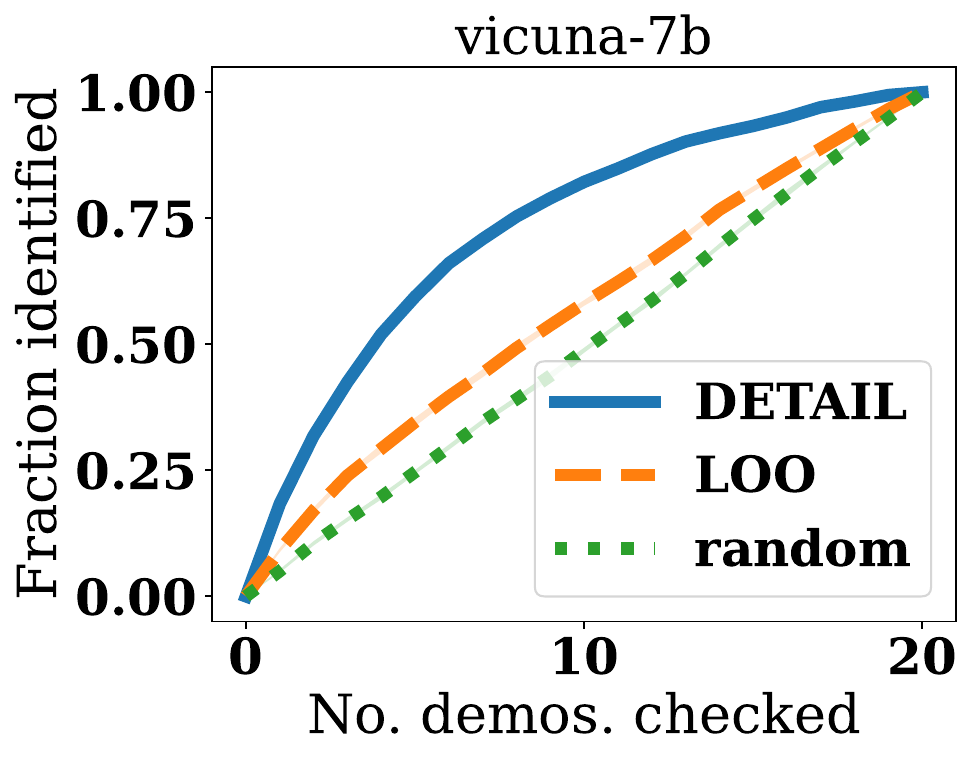}
    \end{subfigure}
\hfill
\begin{subfigure}[t]{0.23\textwidth}
    \includegraphics[width=\textwidth]{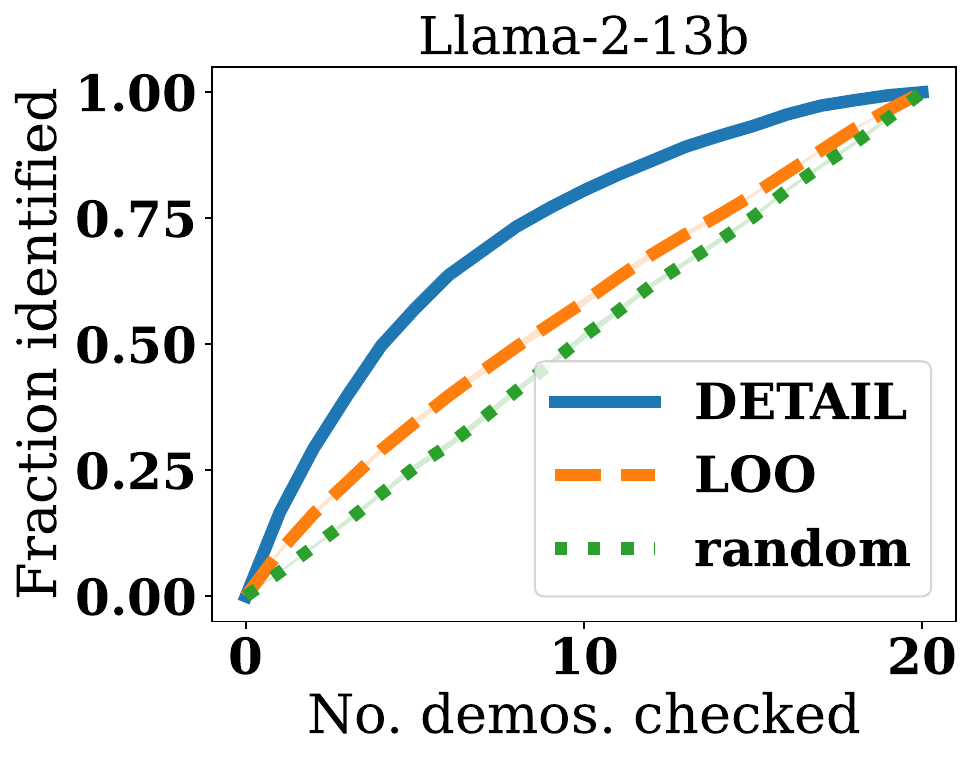}
    \end{subfigure}
\hfill
\begin{subfigure}[t]{0.23\textwidth}
    \includegraphics[width=\textwidth]{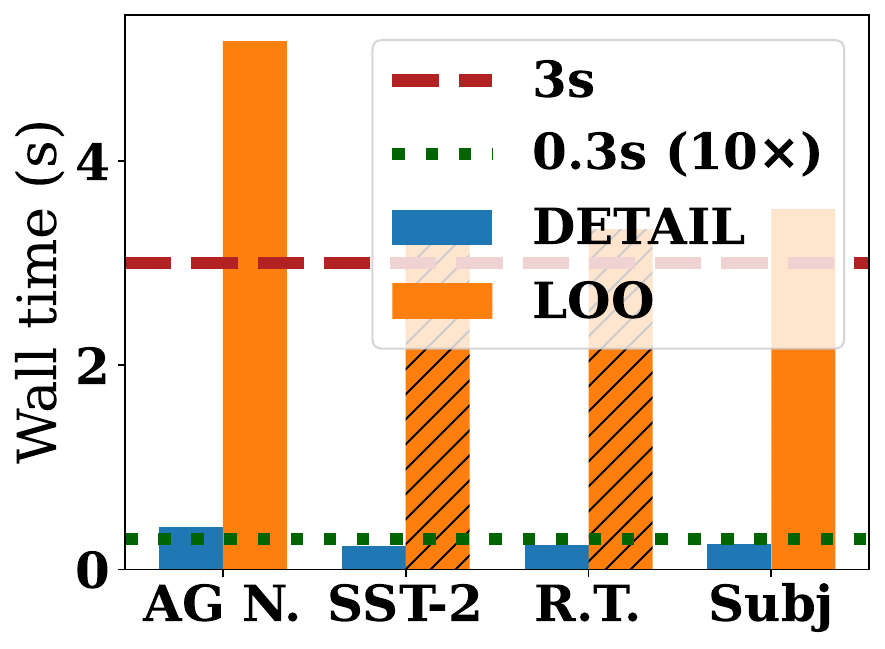}
    \end{subfigure}
\hfill
\begin{subfigure}[t]{0.23\textwidth}
    \includegraphics[width=\textwidth]{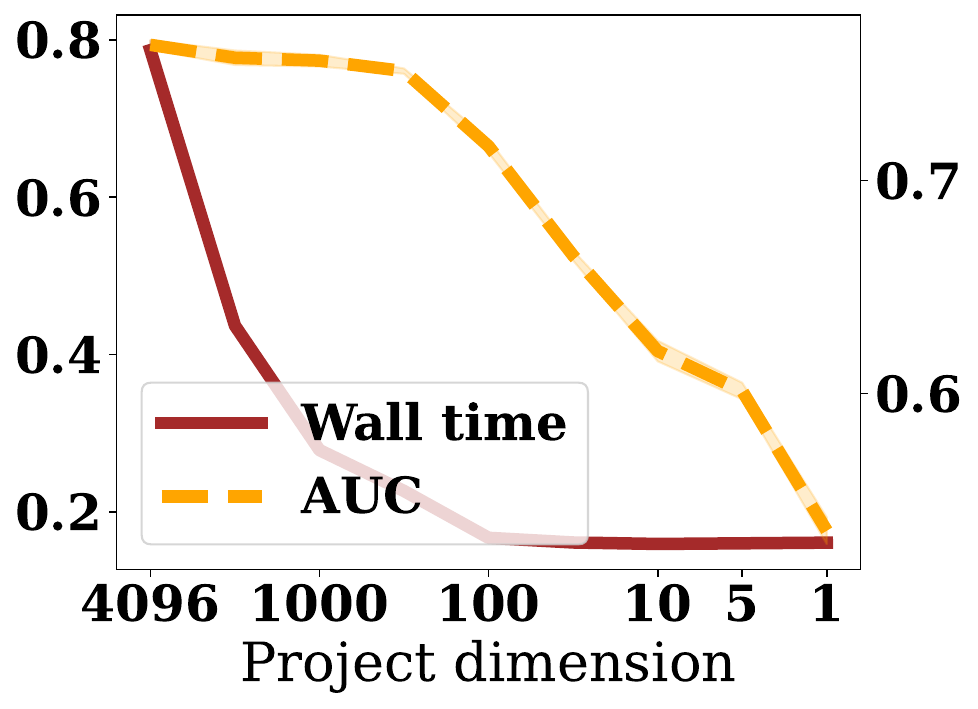}
    \end{subfigure}
\hfill
\caption{($1$st and $2$nd) Fraction of noisy labels identified vs.~number of demonstrations ranked by \texttt{DETAIL} (with $d' = 1000$) and LOO checked on Subj using Vicuna-7b and Llama-2-13b respectively. ($3$rd)  Wall time comparison between \texttt{DETAIL} and LOO on all datasets. ($4$th) wall time in seconds (left $y$-axis) and  AUCROC (right $y$-axis) vs.~projection dimension on Subj using Vicuna-7b. All experiments are repeated $10$ trials. $\lambda=10^{-9}$. Lines and shades represent the mean and std.~error.
}
\label{fig:llm_noisy_label_detection}
\end{figure}

\paragraph{Noisy demonstration detection.} We utilize the \texttt{DETAIL} score to detect noisy demonstrations with corrupted labels. The experiment setup largely follows \citep[Section 5.4]{koh2017}. We randomly draw $100$ ICL datasets each consisting of $20$ demonstrations and $1$ query. For each ICL dataset, we randomly corrupt the labels of $4$ demonstrations (i.e., flipping the label to an incorrect class). The demonstrations are then ranked in descending order of their $\cI_{\text{self}}$. The fraction of noisy demonstrations detected is plotted in the first $3$ figures of \cref{fig:llm_noisy_label_detection} (result for other datasets deferred to \cref{app:exp_noisy}). We compare our method with the leave-one-out (LOO) score~\citep{cook1977} where the difference in cross-entropy loss of the model output is used as the utility. It can be observed that LOO performs close to random selection, whereas our method has a much higher identification rate w.r.t.~the number of demonstrations checked. We also note that our method \textit{not only} outperforms LOO in effectiveness but is also around $10\times$ faster than LOO since LOO requires multiple LLM inferences for each demonstration in the ICL dataset.

\paragraph{Dimension reduction via random projection.} We analyze the impact of random projection on the effectiveness of \texttt{DETAIL}. Intuitively, dimension reduction trades off the effectiveness of \texttt{DETAIL} with computational efficiency, specifically the $\cO(d^3)$ cost for inverting $K_{\cI}$. To understand the trade-off, we follow the same setup as the noisy demonstration detection experiment and compare the change in AUC ROC of detection and system wall time as the dimension $d'$ of the projection matrix $P$ decreases. The result for Subj is in the last figure of \cref{fig:llm_noisy_label_detection} (results for other datasets deferred to \cref{app:exp_dim_reduction_additional}), showing that wall time stays minimal ($\approx 0.3$s) for project dimensions up to $1000$ generally before it exponentially increases. Effectiveness measured in terms of AUC reaches optimal with  $d' \geq 1000$. The results suggest a ``sweet spot'' \textendash\  $d' \approx 1000$ \textendash \  for a low running time and high performance.


\subsection{Applications of \texttt{DETAIL}}
\label{sec:exp_application}

With the two experiments above verifying the effectiveness of \texttt{DETAIL} ($\cI_{\text{self}}$ and $\cI_{\text{test}}$) and the experiment on random projection which ensures computational efficiency, we demonstrate next how \texttt{DETAIL}, with $\cI_{\text{self}}$ for noisy demonstration detection and $\cI_{\text{test}}$ for demonstration perturbation, can be applied to real-world scenarios, achieving superior performance and speed.

\begin{figure}[!ht]
  \begin{minipage}[b]{.3\linewidth}
    \centering
    \includegraphics[width=0.95\linewidth]{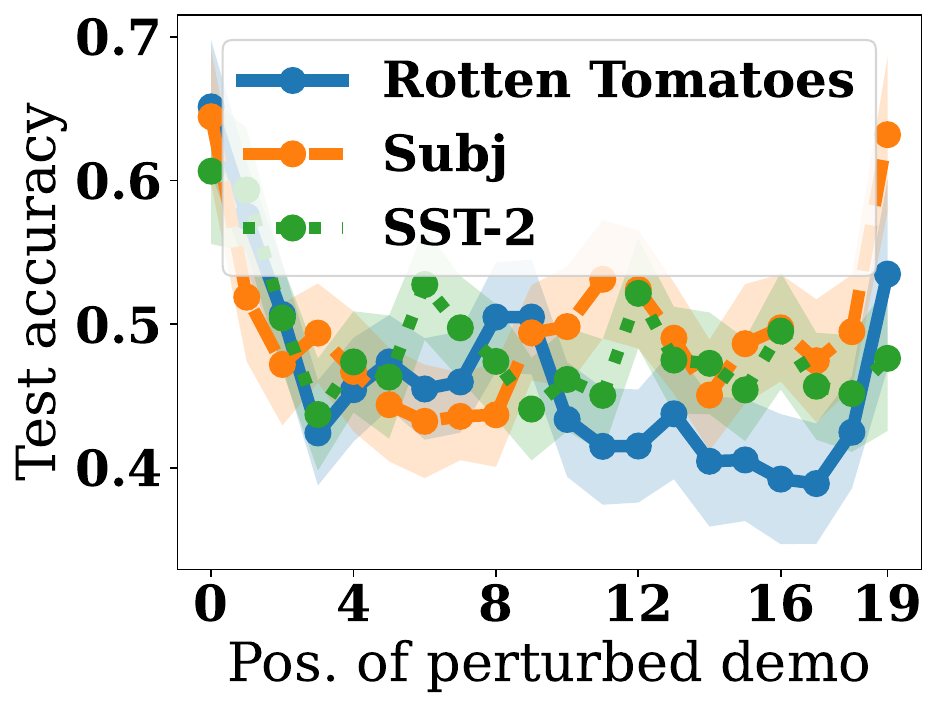}
    \captionof{figure}{Test accuracy (mean and std. error) vs.~position of the demonstration with the corrupted label over $50$ trials.}
    \label{fig:position}
  \end{minipage}\hfill
  \begin{minipage}[b]{.69\linewidth}
    \centering
    \captionof{table}{Predictive accuracy of demonstrations permuted randomly and based on $\cI_{\text{self}}$, respectively. The mean and standard error (in bracket) with $80$ repeated trials is shown.}
    \resizebox{\linewidth}{!}{
    \footnotesize
    \begin{tabular}{@{}lccc@{}}
    \toprule
         & Subj & SST-2 & Rotten Tomatoes \\ 
    \midrule
     \textbf{No corrupted demo}\\
        Baseline (random)  & \textcolor{red!50!black}{0.722 (7.22e-03)}  & \textcolor{red!50!black}{0.665 (5.24e-03)} &  \textcolor{red!50!black}{0.660 (1.08e-02)} \\
        Reorder (\texttt{DETAIL}) & \textcolor{green!50!black}{0.743 (7.10e-03)} &  \textcolor{green!50!black}{0.679 (5.42e-03)} & \textcolor{green!50!black}{0.684 (1.15e-02)}  \\
        Difference $\uparrow$ & \textbf{0.0206 (7.40e-03)} & \textbf{0.0139 (6.08e-03)} & \textbf{0.0244 (1.11e-02)} \\
      \midrule
     \textbf{Corrupt $3$ demos}\\
        Baseline (random) & \textcolor{red!50!black}{0.655 (8.54e-03)}  & \textcolor{red!50!black}{0.607 (7.61e-03)} &  \textcolor{red!50!black}{0.553 (1.10e-02)} \\
        Reorder (\texttt{DETAIL}) & \textcolor{green!50!black}{0.685 (9.39e-03)} &  \textcolor{green!50!black}{0.630 (7.04e-03)} & \textcolor{green!50!black}{0.582 (1.42e-02)}  \\
        Difference $\uparrow$ & \textbf{0.0300 (9.10e-03)} & \textbf{0.0230 (7.22e-03)} & \textbf{0.0291 (1.06e-02)} \\
      \bottomrule                          
    \end{tabular}
    }
    \label{tab:position}
  \end{minipage}
\end{figure}

\paragraph{ICL order optimization.} One distinctive trait of ICL compared to conventional ML is that the order of demonstrations affects the model's predictive performance~\citep{liu2023lost,lu2022fantastically}. We show that $\cI_{\text{self}}$ helps reorder the demonstrations with improved model predictive performance. We first show, using a Vicuna-7b model, that moving demonstrations with lower quality to the front (or back) of the prompt tends to improve the test accuracy of the model. To see this, we corrupt the label of a random demonstration and allocate this corrupted demonstration to different positions of the ICL dataset (each with $20$ demonstrations with permutations drawn from a Sobol sequence~\citep{mitchellsobol22} to capture the average performance better). A general trend with decreasing-then-increasing accuracy can be observed in \cref{fig:position}: Allocating noisy demonstrations to the front (or the back) results in much higher test accuracy. Leveraging this insight, we utilize $\cI_{\text{self}}$ to reorder a random permutation of ICL demonstrations and show the reordered prompt improves the test accuracy. For each randomly ordered prompt, $\cI_{\text{self}}$ for each demonstration is computed (note that this computation only requires $1$ pass of the LLM). Then, based on the trend observed in \cref{fig:position}, for Subj and Rotten Tomatoes datasets, the demonstrations are reordered by placing the two demonstrations with the largest $\cI_{\text{self}}$ in front followed by the rest in ascending order. For SST-2, the demonstrations are reordered in descending order of $\cI_{\text{self}}$. To simulate situations where demonstrations have varying quality, we additionally consider randomly perturbing $3$ demonstrations (and $6$ demonstrations in \cref{app:demo_reorder}) in each ICL dataset. We note a clear improvement in test accuracy of $1.4\% \sim 3.0\%$ via reordering demonstrations \emph{only}, as shown in \cref{tab:position}. The improvement demonstrates that $\cI_{\text{self}}$ can identify demonstrations that are low-quality or inconsistent with other demonstrations in the ICL dataset.

\begin{figure}[!ht]
\centering
\begin{subfigure}[t]{0.23\textwidth}
    \includegraphics[width=\textwidth]{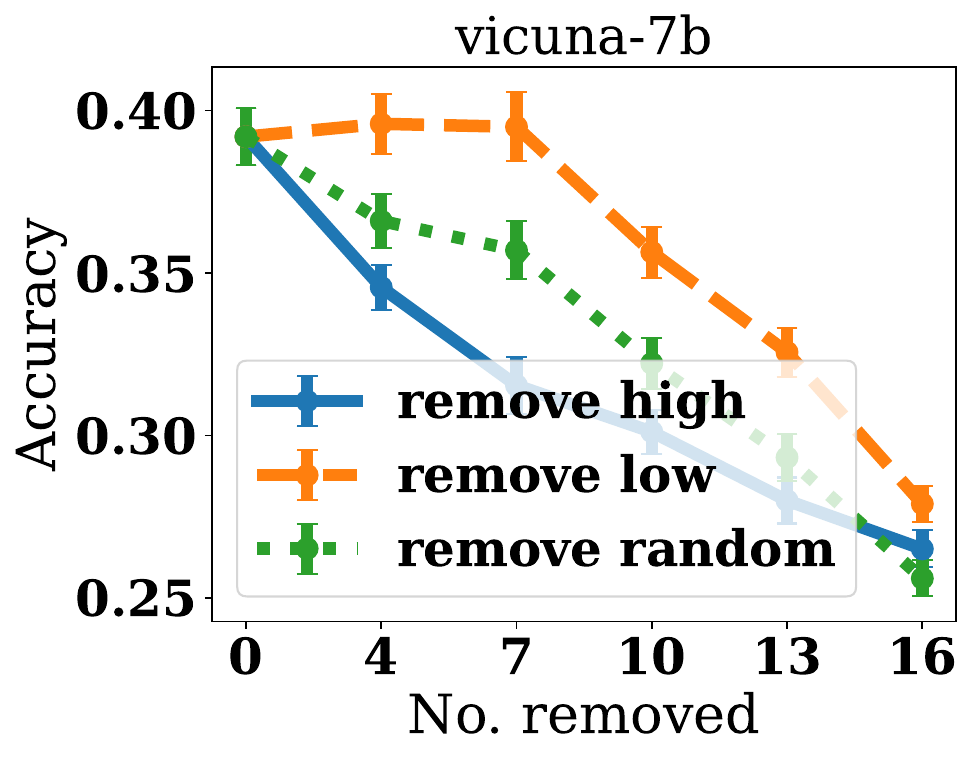}
    \end{subfigure}
\hfill
\begin{subfigure}[t]{0.22\textwidth}
    \includegraphics[width=\textwidth]{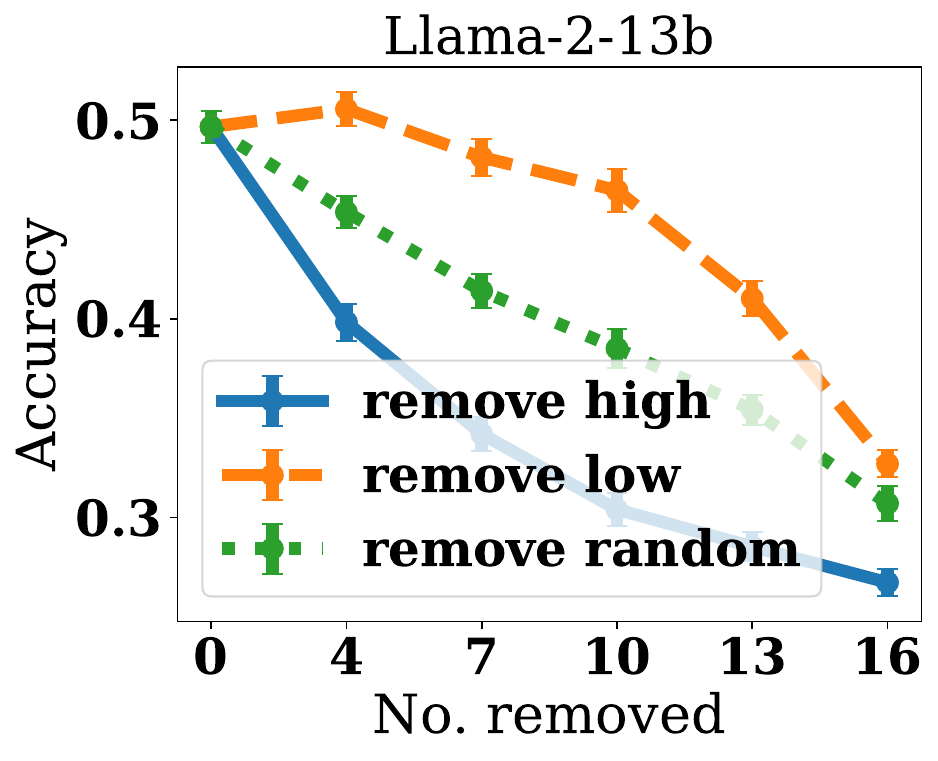}
    \end{subfigure}
\hfill
\begin{subfigure}[t]{0.23\textwidth}
    \includegraphics[width=\textwidth]{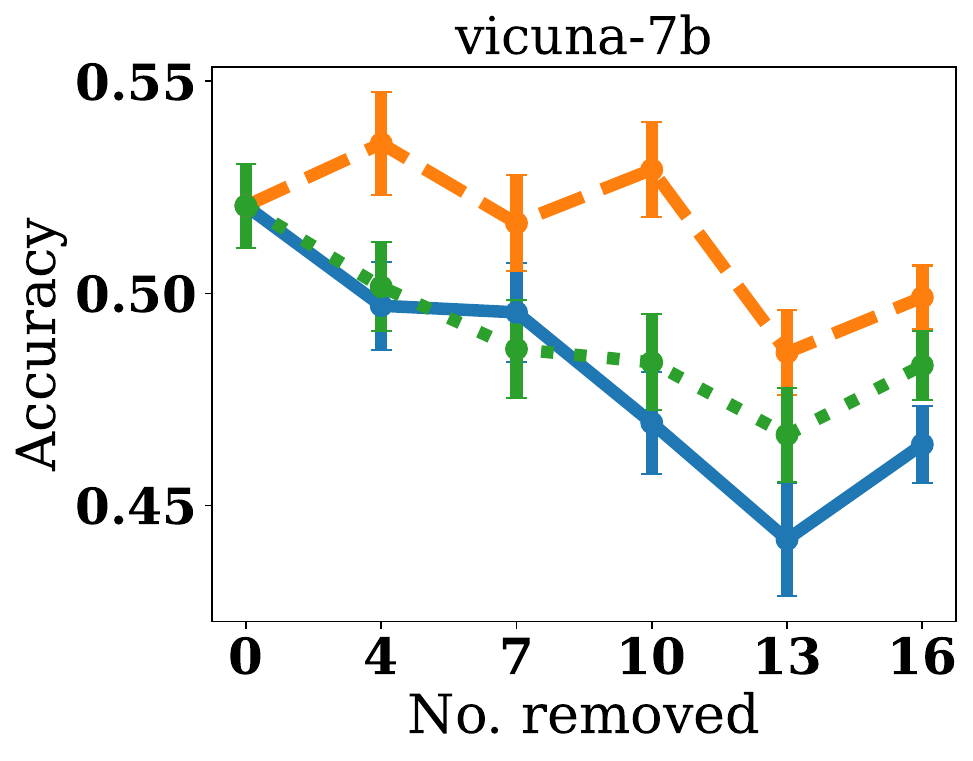}
    \end{subfigure}
\hfill
\begin{subfigure}[t]{0.22\textwidth}
    \includegraphics[width=\textwidth]{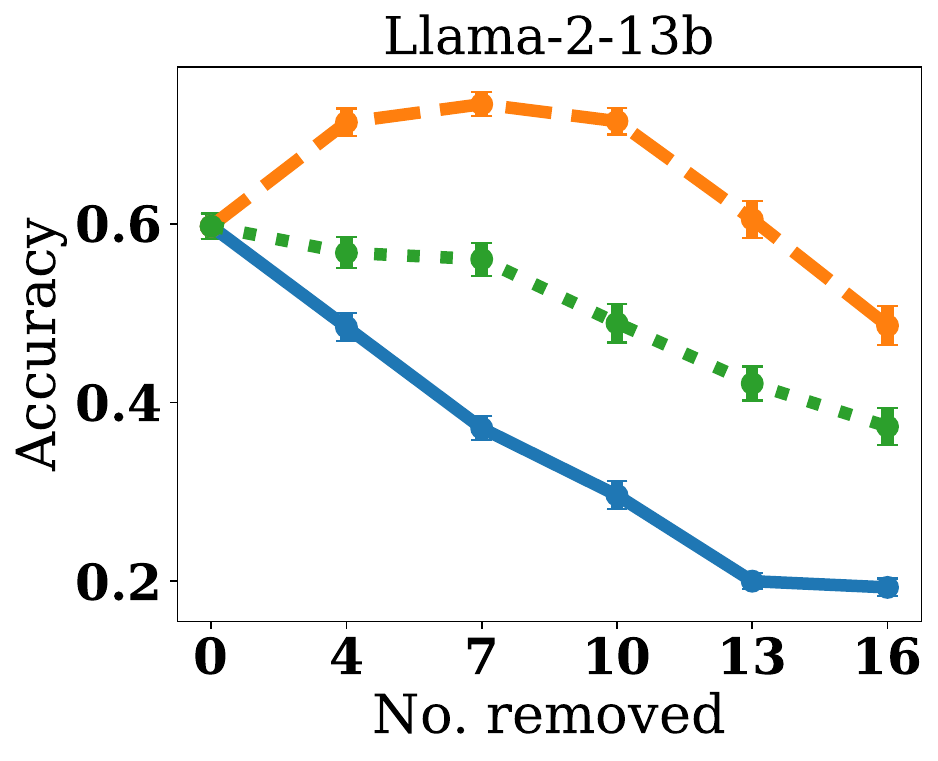}
    \end{subfigure}
\hfill
\caption{Test accuracy vs.~number of demonstrations removed using $\cI_{\text{test}}$ on ($1$st and $2$nd) AG news and ($3$rd and $4$th) Rotten Tomatoes using Vicuna-7b and Llama-2-13b. All experiments are repeated with $80$ trials. Lines and bars represent the mean and standard error.}
\label{fig:icl_data_curation}
\end{figure}

\paragraph{ICL demonstration curation.} In the demonstration perturbation experiment, we have verified that our $\cI_{\text{test}}$ can correctly attribute helpful demonstrations w.r.t.~a query. A direct application is demonstration curation where a subset of most helpful demonstrations are selected to prompt the LLM while maintaining accuracy \textit{on a test dataset}.\footnote{Note that a key difference between demonstration curation task and demonstration perturbation task is that for curation, the test dataset is unknown when computing the \texttt{DETAIL} scores.} This application is useful, especially for saving the cost of querying LLMs.\footnote{At the time of this writing, GPT-4 API costs \$10/1mln tokens. See \url{https://openai.com/pricing}.} For proprietary LLMs, reducing the prompt length can also significantly save inference time which scales quadratically in the prompt length. As a setup, we fix a randomly selected set of $120$ demonstrations as the test set. In each trial, we randomly pick $20$ demonstrations to form an ICL dataset and another $20$ demonstrations as the validation set. The individual $\cI_{\text{test}}$'s on each validation demonstration are summed as the final score. Then, demonstrations with the lowest scores are removed (in position). We randomly corrupt $5$ demonstrations in each ICL dataset to simulate prompts with varying qualities. The results are shown in \cref{fig:icl_data_curation} (results on other datasets deferred to \cref{app:other_attr_additional}). A clear gap between the test accuracy after removing demonstrations with high/low $\cI_{\text{test}}$ can be observed for both Vicuna-7b and Llama-2-13 on both binary (Rotten Tomatoes) and $4$-way classification (AG News). Removing demonstrations with lower $\cI_{\text{test}}$'s maintains (or even improves) the test accuracy. Moreover, the gap for the 13B model is wider and more certain (shorter error bars), signaling better curation. We attribute this phenomenon to the better capability of the larger model to formulate an ``internal optimizer'', which enhances the attributive power of  $\cI_{\text{test}}$.


\subsection{Comparison with Other Attribution Methods}

We compare our \texttt{DETAIL} score with other metrics proposed for demonstration attribution/selection or can be directly extended to attributing demonstrations. We analyze both the attributability via a demonstration curation experiment and the computational cost via recording the system wall time for performing the attribution. We select representative conventional approaches from different paradigms, including integrated gradients (IG)~\citep{sundararajan17} and LIME~\citep{lime2016} (Attention~\citep{bahdanau2016neural} in \cref{app:other_attr_additional}).  As these methods are originally designed for token-level attribution, we use the sum of the scores of all tokens in each demonstration as the attribution score. We also compare recent efforts on demonstration selection~\citep{chang2023data,nguyen2023incontext,s2024incontext}. In natural language processing, a popular choice for attribution has been using BERT-based scores~\citep{devlin-etal-2019-bert,Zhang2019BERTScoreET} to match the similarity between texts. These methods enjoy the benefit of fast inference since the scores are model-agnostic. However, the independence between the scores and the transformer being applied limits their interpretability and attribution accuracy. We compute the BERT score obtained on a popular sentence transformer model~\footnote{\url{https://huggingface.co/sentence-transformers/msmarco-bert-base-dot-v5}} and compare its performance against \texttt{DETAIL}. We select an ICL dataset of $20$ demonstrations, compute the attribution scores on a validation set of $20$ demonstrations, and record the accuracy after removing $10$ demonstrations in place on $120$ test queries. The results are tabulated in \cref{tab:other_attr}. For hyper-parameter, we choose $M=100$ and $k=1$ for \citet{nguyen2023incontext}, $5$ iterations of LiSSA update~\citep{agarwal17lissa} for \citet{s2024incontext}, and $M=10,K=4$ for datamodel~\citep{chang2023data}. We do not perform batch inference for a fair comparison of computational time as some approaches do not admit batch inference. We also use a projection of $d'=1000$ to compute $\cI_{\text{test}}$. It can be observed that \texttt{DETAIL} outperforms all other attribution methods in test accuracy. Our computation is efficient (with wall time of $4 \sim 10$s), achieving over $5\times$ speedup compared to other methods except \citet{s2024incontext} and BERT score which achieve a comparable computational time to ours but a lower accuracy. Notably, IG and LIME perform close to random removal, which is likely because these methods are designed for token-level attribution and generalize poorly to demonstration-level attribution. Interestingly, we find that combing \texttt{DETAIL} and BERT score achieves state-of-the-art performance. We discuss the experiment setup in detail in \cref{app:other_attr_additional}.

\begin{table}[!ht]
    \centering
    \setlength{\tabcolsep}{5pt}
    \caption{Test accuracy after curating the ICL dataset and the incurred wall time (in seconds on one L40 GPU). The mean and std. error (in bracket) is shown with $20$ repeated trials.}
    \resizebox{\linewidth}{!}{
    \footnotesize
    \begin{tabular}{@{}lcccccccc@{}}
    \toprule
         Metric & 
         \texttt{DETAIL} ($d'=1000$) &
         IG~\citep{sundararajan17} & LIME~\citep{lime2016} & \citet{nguyen2023incontext} & \citet{s2024incontext} & Datamodel~\citep{chang2023data} & BERT-Score~\citep{Zhang2019BERTScoreET} & Random  \\ 
    \midrule
     \textbf{Subj}\\
        Accuracy $\uparrow$  & \textbf{0.747 (2.60e-02)}  
        & 0.658 (2.22e-02) & 0.665 (2.41e-02) & 0.583 (2.75e-02) & 0.556 (1.38e-02)  & 0.658 (2.62e-02) & 0.671 (2.43e-02) & 0.654 (2.54e-02)  \\
        Wall time $\downarrow$    & \textbf{5.22 (1.17e-01)} 
        & 593 (1.20e+01) & 393 (2.44e+01) & 54.3 (3.78e-01) & 9.37 (4.19e-01) & 746 (3.42e+00) & \textbf{2.97 (3.05e-02)} &
        N.A.   \\
      \midrule
      \textbf{SST-2} \\
        Accuracy $\uparrow$  & \textbf{0.607 (2.12e-02)}  
        & 0.458 (2.06e-02) & 0.476 (1.87e-02) & 0.513 (1.88e-02) & 0.493 (1.34e-02) & 0.477 (2.54e-02) & 0.460 (2.36e-02) &
       0.469 (2.15e-02)  \\
        Wall time $\downarrow$    & \textbf{4.88 (1.35e-01)} 
        & 458 (7.99e+00) & 337 (1.69e+01) & 121 (4.79e+00) & 10.6 (7.80e-01) & 713 (1.96e+00) & \textbf{2.91 (2.78e-02)} &
        N.A.   \\
      \midrule
      \textbf{Rotten Tomatoes} \\
        Accuracy $\uparrow$  & \textbf{0.555 (1.94e-02)}  
        & 0.442 (2.13e-02) & 0.435 (1.39e-02) & 0.520 (2.17e-02) & 0.498 (1.72e-02) & 0.484 (1.87e-02) & 0.435 (1.33e-02) &
        0.457 (2.19e-02)  \\
        Wall time $\downarrow$    & \textbf{5.11 (1.06e-01)} 
        & 525 (1.23e+01) & 245 (6.32e+01) &
        122 (4.68e+00) & 9.74 (5.57e-01) & 732 (2.10e+00) & \textbf{2.95 (4.43e-02)} &
        N.A.   \\
      \midrule
      \textbf{AG News} \\
        Accuracy $\uparrow$  & \textbf{0.412 (1.35e-02)}  
        & 0.351 (1.65e-02) & 0.368 (1.73e-02) & 0.392 (1.42e-02) & 0.361 (1.83e-02) & 0.373 (1.31e-02) & 0.355 (1.58e-02) &
        0.379 (1.70e-02)  \\
        Wall time $\downarrow$    & 10.4 (1.07e-01) 
        & 1208 (2.16e+01) & 599 (1.03e+01) &
        81.3 (6.05e-01) & \textbf{6.94 (4.78e-02)} & 997 (7.55e+00) & \textbf{3.15 (5.87e-02)} &
        N.A.   \\
      \bottomrule                          
    \end{tabular}
    \label{tab:other_attr}
    }
\end{table}

\subsection{Transferability to Black-box Models}
\label{sec:transfer}
We evaluate the transferability of \texttt{DETAIL} on GPT-3.5,\footnote{We use gpt-3.5-turbo-1106. See \url{https://platform.openai.com/docs/models/gpt-3-5-turbo}.} a popular black-box model. We experiment with both the demonstration reordering and demonstration curation tasks where we compute the \texttt{DETAIL} scores on a Vicuna-7b model (white-box) and then test the performance on GPT. Our method produces promising results on both tasks as shown in \cref{tab:position_transfer} and \cref{tab:other_attr_transfer} respectively. Notably, curating demonstrations with our method achieves a $17.9\%$ average improvement in accuracy compared to random curating on the demonstration curation task (\cref{tab:other_attr_transfer}). We also note an over $2\%$ improvement in accuracy for reordering task if we corrupt $3$ demonstrations (\cref{tab:position_transfer}). With no corrupted demonstration, reordering with our approach does not improve performance on GPT-3.5, which we attribute to the stronger inference power of GPT-3.5, resulting in less variance w.r.t.~demonstration orders, consistent with the findings in \citep{lu2022fantastically}.

\begin{figure}[!ht]
  \begin{minipage}[b]{.53\linewidth}
    \centering
    \captionof{table}{Accuracy (on GPT-3.5) of demonstrations (demos) permuted randomly and based on $\cI_{\text{self}}$. Mean and std. error (in bracket) with $80$ trials is shown.}
    \resizebox{\linewidth}{!}{
    \footnotesize
    \begin{tabular}{@{}lccc@{}}
    \toprule
         & Subj & SST-2 & Rotten Tomatoes \\ 
    \midrule
     \textbf{No corrupted demo}\\
        Baseline (random)  & \textcolor{red!50!black}{0.708(7.79e-04)}  & \textcolor{green!50!black}{0.799(7.52e-04)} &  \textcolor{red!50!black}{0.901(6.01e-04)} \\
        Reorder (\texttt{DETAIL}) & \textcolor{green!50!black}{0.711(7.51e-04)} &  \textcolor{red!50!black}{0.792(9.01e-04)} & \textcolor{green!50!black}{0.909(4.84e-04) }  \\
        Difference $\uparrow$ & \textbf{0.002(7.52e-04)} & -0.007(6.49e-04) & \textbf{0.008(6.14e-04)} \\
      \midrule
     \textbf{Corrupt $3$ demos}\\
        Baseline (random)  & \textcolor{red!50!black}{0.628(8.21e-04)}  & \textcolor{red!50!black}{0.720(1.11e-03)} &  \textcolor{red!50!black}{0.788(1.44e-03)} \\
        Reorder (\texttt{DETAIL}) & \textcolor{green!50!black}{0.660(9.57e-04)} &  \textcolor{green!50!black}{0.742(1.20e-03)} & \textcolor{green!50!black}{0.816(1.60e-03)}  \\
        Difference $\uparrow$ & \textbf{0.032(8.61e-04)} & \textbf{0.022(8.92e-04)} & \textbf{0.028(1.10e-03)} \\
      \bottomrule                          
    \end{tabular}
    }
    \label{tab:position_transfer}
  \end{minipage}\hfill
  \begin{minipage}[b]{.46\linewidth}
    \centering
    \captionof{table}{Accuracy (on GPT-3.5) on a test dataset of size $20$ after curating $10$ demonstrations from the ICL dataset. The mean and std. error (in bracket) of accuracy after removal is shown with $20$ repeated trials.}
    \resizebox{\linewidth}{!}{
    \footnotesize
    \begin{tabular}{@{}lcc@{}}
    \toprule
         Dataset & 
         \texttt{DETAIL} ($d'=1000$) & Random \\ 
    \midrule
        Subj  & \textbf{0.842 (2.16e-02)}  & 0.660 (3.47e-02) \\
        SST-2 & \textbf{0.812 (1.96e-02)}  & 0.618 (5.51e-02) \\
        Rotten Tomatoes  & \textbf{0.690 (4.66e-02)}  & 0.420 (5.14e-02)\\
        AG News  & \textbf{0.515 (3.08e-02)}  & 0.447 (2.73e-02)\\
      \bottomrule                          
    \end{tabular}
    }
    \label{tab:other_attr_transfer}
  \end{minipage}
\end{figure}

\section{Conclusion, Limitation, and Future Work} \label{sec:conclusion}

We tackle the problem of attributing demonstrations in ICL for transformers. Based on the well-known influence function commonly used for attributing conventional ML, we propose \texttt{DETAIL}, an innovative adoption of the influence function to ICL through the lens of treating the transformer as implementing an internal kernelized ridge regression. Combined with a dimension reduction technique using random projection, \texttt{DETAIL} can be computed in real-time with an impressive performance on various real-world related tasks such as demonstration order optimization and demonstration curation. One limitation of our approach is the need to access the internal state of the transformer, which we mitigate by additionally showing that \texttt{DETAIL} scores are transferable to black-box models. As a first step toward attributing demonstrations w.r.t.~a transformer's internal optimizer, we hope this work serves as a building block for future research to develop attribution techniques for more generalized prompting settings such as chain-of-thought~\citep{wei2023chainofthought}.

\begin{ack}
This research is supported by the National Research Foundation Singapore and the Singapore Ministry of Digital Development and Innovation, National AI Group under the AI Visiting Professorship Programme (award number AIVP-$2024$-$001$).
This research is supported by the National Research Foundation (NRF), Prime Minister's Office, Singapore under its Campus for Research Excellence and Technological Enterprise (CREATE) programme. The Mens, Manus, and Machina (M3S) is an interdisciplinary research group (IRG) of the Singapore MIT Alliance for Research and Technology (SMART) centre.
\end{ack}

{
\bibliographystyle{natbib}
\bibliography{reference}

\begin{thebibliography}{79}
\providecommand{\natexlab}[1]{#1}
\providecommand{\url}[1]{\texttt{#1}}
\expandafter\ifx\csname urlstyle\endcsname\relax
  \providecommand{\doi}[1]{doi: #1}\else
  \providecommand{\doi}{doi: \begingroup \urlstyle{rm}\Url}\fi

\bibitem[Agarwal et~al.(2017)Agarwal, Bullins, and Hazan]{agarwal17lissa}
Naman Agarwal, Brian Bullins, and Elad Hazan.
\newblock Second-order stochastic optimization for machine learning in linear time.
\newblock \emph{Journal of Machine Learning Research}, 2017.

\bibitem[Ahn et~al.(2023)Ahn, Cheng, Daneshmand, and Sra]{ahn2023transformers}
Kwangjun Ahn, Xiang Cheng, Hadi Daneshmand, and Suvrit Sra.
\newblock Transformers learn to implement preconditioned gradient descent for in-context learning.
\newblock In \emph{Advances in Neural Information Processing Systems}, 2023.

\bibitem[Akyürek et~al.(2023)Akyürek, Schuurmans, Andreas, Ma, and Zhou]{akyürek2023learning}
Ekin Akyürek, Dale Schuurmans, Jacob Andreas, Tengyu Ma, and Denny Zhou.
\newblock What learning algorithm is in-context learning? investigations with linear models, 2023.

\bibitem[Almazrouei et~al.(2023)Almazrouei, Alobeidli, Alshamsi, Cappelli, Cojocaru, Debbah, Étienne Goffinet, Hesslow, Launay, Malartic, Mazzotta, Noune, Pannier, and Penedo]{almazrouei2023falcon}
Ebtesam Almazrouei, Hamza Alobeidli, Abdulaziz Alshamsi, Alessandro Cappelli, Ruxandra Cojocaru, Mérouane Debbah, Étienne Goffinet, Daniel Hesslow, Julien Launay, Quentin Malartic, Daniele Mazzotta, Badreddine Noune, Baptiste Pannier, and Guilherme Penedo.
\newblock The falcon series of open language models, 2023.

\bibitem[Ansel et~al.(2024)Ansel, Yang, He, Gimelshein, Jain, Voznesensky, Bao, Bell, Berard, Burovski, Chauhan, Chourdia, Constable, Desmaison, DeVito, Ellison, Feng, Gong, Gschwind, Hirsh, Huang, Kalambarkar, Kirsch, Lazos, Lezcano, Liang, Liang, Lu, Luk, Maher, Pan, Puhrsch, Reso, Saroufim, Siraichi, Suk, Suo, Tillet, Wang, Wang, Wen, Zhang, Zhao, Zhou, Zou, Mathews, Chanan, Wu, and Chintala]{pytorch2024}
Jason Ansel, Edward Yang, Horace He, Natalia Gimelshein, Animesh Jain, Michael Voznesensky, Bin Bao, Peter Bell, David Berard, Evgeni Burovski, Geeta Chauhan, Anjali Chourdia, Will Constable, Alban Desmaison, Zachary DeVito, Elias Ellison, Will Feng, Jiong Gong, Michael Gschwind, Brian Hirsh, Sherlock Huang, Kshiteej Kalambarkar, Laurent Kirsch, Michael Lazos, Mario Lezcano, Yanbo Liang, Jason Liang, Yinghai Lu, CK~Luk, Bert Maher, Yunjie Pan, Christian Puhrsch, Matthias Reso, Mark Saroufim, Marcos~Yukio Siraichi, Helen Suk, Michael Suo, Phil Tillet, Eikan Wang, Xiaodong Wang, William Wen, Shunting Zhang, Xu~Zhao, Keren Zhou, Richard Zou, Ajit Mathews, Gregory Chanan, Peng Wu, and Soumith Chintala.
\newblock Pytorch 2: Faster machine learning through dynamic {P}ython bytecode transformation and graph compilation.
\newblock In \emph{ACM International Conference on Architectural Support for Programming Languages and Operating Systems}, 2024.

\bibitem[Bahdanau et~al.(2016)Bahdanau, Cho, and Bengio]{bahdanau2016neural}
Dzmitry Bahdanau, Kyunghyun Cho, and Yoshua Bengio.
\newblock Neural machine translation by jointly learning to align and translate.
\newblock In \emph{Proceedings of the International Conference on Learning Representations}, 2016.

\bibitem[Bai et~al.(2023)Bai, Chen, Wang, Xiong, and Mei]{bai2023transformers}
Yu~Bai, Fan Chen, Huan Wang, Caiming Xiong, and Song Mei.
\newblock Transformers as statisticians: Provable in-context learning with in-context algorithm selection.
\newblock In \emph{Advances in Neural Information Processing Systems}, 2023.

\bibitem[Barshan et~al.(2020)Barshan, Brunet, and Dziugaite]{barshan20a}
Elnaz Barshan, Marc-Etienne Brunet, and Gintare~Karolina Dziugaite.
\newblock Relatif: Identifying explanatory training samples via relative influence.
\newblock In \emph{Proceedings of the International Conference on Artificial Intelligence and Statistics}, 2020.

\bibitem[Basu et~al.(2021)Basu, Pope, and Feizi]{basu2021influence}
Samyadeep Basu, Philip Pope, and Soheil Feizi.
\newblock Influence functions in deep learning are fragile.
\newblock In \emph{Proceedings of the International Conference on Learning Representations}, 2021.

\bibitem[Bhattamishra et~al.(2024)Bhattamishra, Patel, Blunsom, and Kanade]{bhattamishra2023understanding}
Satwik Bhattamishra, Arkil Patel, Phil Blunsom, and Varun Kanade.
\newblock Understanding in-context learning in transformers and llms by learning to learn discrete functions.
\newblock In \emph{Proceedings of the International Conference on Learning Representations}, 2024.

\bibitem[Bohnet et~al.(2022)Bohnet, Tran, Verga, Aharoni, Andor, Soares, Ciaramita, Eisenstein, Ganchev, Herzig, Hui, Kwiatkowski, Ma, Ni, Saralegui, Schuster, Cohen, Collins, Das, Metzler, Petrov, and Webster]{bohnet2023attributed}
Bernd Bohnet, Vinh~Q. Tran, Pat Verga, Roee Aharoni, Daniel Andor, Livio~Baldini Soares, Massimiliano Ciaramita, Jacob Eisenstein, Kuzman Ganchev, Jonathan Herzig, Kai Hui, Tom Kwiatkowski, Ji~Ma, Jianmo Ni, Lierni~Sestorain Saralegui, Tal Schuster, William~W. Cohen, Michael Collins, Dipanjan Das, Donald Metzler, Slav Petrov, and Kellie Webster.
\newblock Attributed question answering: Evaluation and modeling for attributed large language models, 2022.
\newblock URL \url{https://arxiv.org/abs/2212.08037}.

\bibitem[Bommasani et~al.(2022)Bommasani, Hudson, Adeli, Altman, Arora, von Arx, Bernstein, Bohg, Bosselut, Brunskill, Brynjolfsson, Buch, Card, Castellon, Chatterji, Chen, Creel, Davis, Demszky, Donahue, Doumbouya, Durmus, Ermon, Etchemendy, Ethayarajh, Fei-Fei, Finn, Gale, Gillespie, Goel, Goodman, Grossman, Guha, Hashimoto, Henderson, Hewitt, Ho, Hong, Hsu, Huang, Icard, Jain, Jurafsky, Kalluri, Karamcheti, Keeling, Khani, Khattab, Koh, Krass, Krishna, Kuditipudi, Kumar, Ladhak, Lee, Lee, Leskovec, Levent, Li, Li, Ma, Malik, Manning, Mirchandani, Mitchell, Munyikwa, Nair, Narayan, Narayanan, Newman, Nie, Niebles, Nilforoshan, Nyarko, Ogut, Orr, Papadimitriou, Park, Piech, Portelance, Potts, Raghunathan, Reich, Ren, Rong, Roohani, Ruiz, Ryan, Ré, Sadigh, Sagawa, Santhanam, Shih, Srinivasan, Tamkin, Taori, Thomas, Tramèr, Wang, Wang, Wu, Wu, Wu, Xie, Yasunaga, You, Zaharia, Zhang, Zhang, Zhang, Zhang, Zheng, Zhou, and Liang]{bommasani2022opportunities}
Rishi Bommasani, Drew~A. Hudson, Ehsan Adeli, Russ Altman, Simran Arora, Sydney von Arx, Michael~S. Bernstein, Jeannette Bohg, Antoine Bosselut, Emma Brunskill, Erik Brynjolfsson, Shyamal Buch, Dallas Card, Rodrigo Castellon, Niladri Chatterji, Annie Chen, Kathleen Creel, Jared~Quincy Davis, Dora Demszky, Chris Donahue, Moussa Doumbouya, Esin Durmus, Stefano Ermon, John Etchemendy, Kawin Ethayarajh, Li~Fei-Fei, Chelsea Finn, Trevor Gale, Lauren Gillespie, Karan Goel, Noah Goodman, Shelby Grossman, Neel Guha, Tatsunori Hashimoto, Peter Henderson, John Hewitt, Daniel~E. Ho, Jenny Hong, Kyle Hsu, Jing Huang, Thomas Icard, Saahil Jain, Dan Jurafsky, Pratyusha Kalluri, Siddharth Karamcheti, Geoff Keeling, Fereshte Khani, Omar Khattab, Pang~Wei Koh, Mark Krass, Ranjay Krishna, Rohith Kuditipudi, Ananya Kumar, Faisal Ladhak, Mina Lee, Tony Lee, Jure Leskovec, Isabelle Levent, Xiang~Lisa Li, Xuechen Li, Tengyu Ma, Ali Malik, Christopher~D. Manning, Suvir Mirchandani, Eric Mitchell, Zanele Munyikwa, Suraj Nair,
  Avanika Narayan, Deepak Narayanan, Ben Newman, Allen Nie, Juan~Carlos Niebles, Hamed Nilforoshan, Julian Nyarko, Giray Ogut, Laurel Orr, Isabel Papadimitriou, Joon~Sung Park, Chris Piech, Eva Portelance, Christopher Potts, Aditi Raghunathan, Rob Reich, Hongyu Ren, Frieda Rong, Yusuf Roohani, Camilo Ruiz, Jack Ryan, Christopher Ré, Dorsa Sadigh, Shiori Sagawa, Keshav Santhanam, Andy Shih, Krishnan Srinivasan, Alex Tamkin, Rohan Taori, Armin~W. Thomas, Florian Tramèr, Rose~E. Wang, William Wang, Bohan Wu, Jiajun Wu, Yuhuai Wu, Sang~Michael Xie, Michihiro Yasunaga, Jiaxuan You, Matei Zaharia, Michael Zhang, Tianyi Zhang, Xikun Zhang, Yuhui Zhang, Lucia Zheng, Kaitlyn Zhou, and Percy Liang.
\newblock On the opportunities and risks of foundation models, 2022.

\bibitem[Bradbury et~al.(2018)Bradbury, Frostig, Hawkins, Johnson, Leary, Maclaurin, Necula, Paszke, Vander{P}las, Wanderman-{M}ilne, and Zhang]{jax2018github}
James Bradbury, Roy Frostig, Peter Hawkins, Matthew~James Johnson, Chris Leary, Dougal Maclaurin, George Necula, Adam Paszke, Jake Vander{P}las, Skye Wanderman-{M}ilne, and Qiao Zhang.
\newblock {JAX}: composable transformations of {P}ython+{N}um{P}y programs, 2018.
\newblock URL \url{http://github.com/google/jax}.

\bibitem[Brown et~al.(2020)Brown, Mann, Ryder, Subbiah, Kaplan, Dhariwal, Neelakantan, Shyam, Sastry, Askell, Agarwal, Herbert-Voss, Krueger, Henighan, Child, Ramesh, Ziegler, Wu, Winter, Hesse, Chen, Sigler, Litwin, Gray, Chess, Clark, Berner, McCandlish, Radford, Sutskever, and Amodei]{brown2020}
Tom Brown, Benjamin Mann, Nick Ryder, Melanie Subbiah, Jared~D Kaplan, Prafulla Dhariwal, Arvind Neelakantan, Pranav Shyam, Girish Sastry, Amanda Askell, Sandhini Agarwal, Ariel Herbert-Voss, Gretchen Krueger, Tom Henighan, Rewon Child, Aditya Ramesh, Daniel Ziegler, Jeffrey Wu, Clemens Winter, Chris Hesse, Mark Chen, Eric Sigler, Mateusz Litwin, Scott Gray, Benjamin Chess, Jack Clark, Christopher Berner, Sam McCandlish, Alec Radford, Ilya Sutskever, and Dario Amodei.
\newblock Language models are few-shot learners.
\newblock In \emph{Advances in Neural Information Processing Systems}, 2020.

\bibitem[Chang \& Jia(2023)Chang and Jia]{chang2023data}
Ting-Yun Chang and Robin Jia.
\newblock Data curation alone can stabilize in-context learning.
\newblock In \emph{Proceedings of the Annual Meeting of the Association for Computational Linguistics}, 2023.

\bibitem[Chen et~al.(2024)Chen, Sheen, Wang, and Yang]{chen2024training}
Siyu Chen, Heejune Sheen, Tianhao Wang, and Zhuoran Yang.
\newblock Training dynamics of multi-head softmax attention for in-context learning: Emergence, convergence, and optimality, 2024.

\bibitem[Chowdhery et~al.(2022)Chowdhery, Narang, Devlin, Bosma, Mishra, Roberts, Barham, Chung, Sutton, Gehrmann, Schuh, Shi, Tsvyashchenko, Maynez, Rao, Barnes, Tay, Shazeer, Prabhakaran, Reif, Du, Hutchinson, Pope, Bradbury, Austin, Isard, Gur-Ari, Yin, Duke, Levskaya, Ghemawat, Dev, Michalewski, Garcia, Misra, Robinson, Fedus, Zhou, Ippolito, Luan, Lim, Zoph, Spiridonov, Sepassi, Dohan, Agrawal, Omernick, Dai, Pillai, Pellat, Lewkowycz, Moreira, Child, Polozov, Lee, Zhou, Wang, Saeta, Diaz, Firat, Catasta, Wei, Meier-Hellstern, Eck, Dean, Petrov, and Fiedel]{chowdhery2022palm}
Aakanksha Chowdhery, Sharan Narang, Jacob Devlin, Maarten Bosma, Gaurav Mishra, Adam Roberts, Paul Barham, Hyung~Won Chung, Charles Sutton, Sebastian Gehrmann, Parker Schuh, Kensen Shi, Sasha Tsvyashchenko, Joshua Maynez, Abhishek Rao, Parker Barnes, Yi~Tay, Noam Shazeer, Vinodkumar Prabhakaran, Emily Reif, Nan Du, Ben Hutchinson, Reiner Pope, James Bradbury, Jacob Austin, Michael Isard, Guy Gur-Ari, Pengcheng Yin, Toju Duke, Anselm Levskaya, Sanjay Ghemawat, Sunipa Dev, Henryk Michalewski, Xavier Garcia, Vedant Misra, Kevin Robinson, Liam Fedus, Denny Zhou, Daphne Ippolito, David Luan, Hyeontaek Lim, Barret Zoph, Alexander Spiridonov, Ryan Sepassi, David Dohan, Shivani Agrawal, Mark Omernick, Andrew~M. Dai, Thanumalayan~Sankaranarayana Pillai, Marie Pellat, Aitor Lewkowycz, Erica Moreira, Rewon Child, Oleksandr Polozov, Katherine Lee, Zongwei Zhou, Xuezhi Wang, Brennan Saeta, Mark Diaz, Orhan Firat, Michele Catasta, Jason Wei, Kathy Meier-Hellstern, Douglas Eck, Jeff Dean, Slav Petrov, and Noah Fiedel.
\newblock Palm: Scaling language modeling with pathways, 2022.

\bibitem[Conneau \& Kiela(2018)Conneau and Kiela]{conneau2018senteval}
Alexis Conneau and Douwe Kiela.
\newblock Senteval: An evaluation toolkit for universal sentence representations.
\newblock In \emph{Proceedings of the International Conference on Language Resources and Evaluation}, 2018.

\bibitem[Cook(1977)]{cook1977}
R.~Dennis. Cook.
\newblock Detection of influential observation in linear regression.
\newblock \emph{Technometrics}, 1977.

\bibitem[Dai et~al.(2023)Dai, Sun, Dong, Hao, Ma, Sui, and Wei]{dai-etal-2023-gpt}
Damai Dai, Yutao Sun, Li~Dong, Yaru Hao, Shuming Ma, Zhifang Sui, and Furu Wei.
\newblock Why can {GPT} learn in-context? language models secretly perform gradient descent as meta-optimizers.
\newblock In \emph{Findings of the Association for Computational Linguistics}, 2023.

\bibitem[Dasgupta \& Gupta(2003)Dasgupta and Gupta]{dasgupta2003elementary}
Sanjoy Dasgupta and Anupam Gupta.
\newblock An elementary proof of a theorem of {J}ohnson and {L}indenstrauss.
\newblock \emph{Random Structures \& Algorithms}, 2003.

\bibitem[Deng(2012)]{deng2012mnist}
Li~Deng.
\newblock The {MNIST} database of handwritten digit images for machine learning research.
\newblock \emph{IEEE Signal Processing Magazine}, 2012.

\bibitem[Devlin et~al.(2019)Devlin, Chang, Lee, and Toutanova]{devlin-etal-2019-bert}
Jacob Devlin, Ming-Wei Chang, Kenton Lee, and Kristina Toutanova.
\newblock {BERT}: Pre-training of deep bidirectional transformers for language understanding.
\newblock In \emph{Proceedings of the 2019 Conference of the North {A}merican Chapter of the Association for Computational Linguistics: Human Language Technologies, Volume 1 (Long and Short Papers)}, 2019.

\bibitem[Garg et~al.(2023)Garg, Tsipras, Liang, and Valiant]{garg2023transformers}
Shivam Garg, Dimitris Tsipras, Percy Liang, and Gregory Valiant.
\newblock What can transformers learn in-context? {A} case study of simple function classes, 2023.

\bibitem[Ghorbani \& Zou(2019)Ghorbani and Zou]{ghorbani2019}
Amirata Ghorbani and James Zou.
\newblock Data {S}hapley: Equitable valuation of data for machine learning.
\newblock In \emph{Proceedings of the International Conference on Machine Learning}, 2019.

\bibitem[Grosse et~al.(2023)Grosse, Bae, Anil, Elhage, Tamkin, Tajdini, Steiner, Li, Durmus, Perez, Hubinger, Lukošiūtė, Nguyen, Joseph, McCandlish, Kaplan, and Bowman]{grosse2023studying}
Roger Grosse, Juhan Bae, Cem Anil, Nelson Elhage, Alex Tamkin, Amirhossein Tajdini, Benoit Steiner, Dustin Li, Esin Durmus, Ethan Perez, Evan Hubinger, Kamilė Lukošiūtė, Karina Nguyen, Nicholas Joseph, Sam McCandlish, Jared Kaplan, and Samuel~R. Bowman.
\newblock Studying large language model generalization with influence functions, 2023.

\bibitem[Gu \& Dao(2023)Gu and Dao]{gu2023mamba}
Albert Gu and Tri Dao.
\newblock Mamba: Linear-time sequence modeling with selective state spaces, 2023.

\bibitem[Hainmueller \& Hazlett(2014)Hainmueller and Hazlett]{HainmuellerJens2014KRLS}
Jens Hainmueller and Chad Hazlett.
\newblock Kernel regularized least squares: Reducing misspecification bias with a flexible and interpretable machine learning approach.
\newblock \emph{Political analysis}, 2014.

\bibitem[Hinton et~al.(2015)Hinton, Vinyals, and Dean]{hinton2015distillingknowledgeneuralnetwork}
Geoffrey Hinton, Oriol Vinyals, and Jeff Dean.
\newblock Distilling the knowledge in a neural network.
\newblock In \emph{Advances in Neural Information Processing Systems Deep Learning Workshop}, 2015.

\bibitem[Hsieh et~al.(2023)Hsieh, Li, Yeh, Nakhost, Fujii, Ratner, Krishna, Lee, and Pfister]{hsieh2023distillingstepbystepoutperforminglarger}
Cheng-Yu Hsieh, Chun-Liang Li, Chih-Kuan Yeh, Hootan Nakhost, Yasuhisa Fujii, Alexander Ratner, Ranjay Krishna, Chen-Yu Lee, and Tomas Pfister.
\newblock Distilling step-by-step! outperforming larger language models with less training data and smaller model sizes.
\newblock In \emph{Conference on Empirical Methods in Natural Language Processing}, 2023.

\bibitem[Hu et~al.(2022)Hu, Shen, Wallis, Allen-Zhu, Li, Wang, Wang, and Chen]{hu2022lora}
Edward~J Hu, Yelong Shen, Phillip Wallis, Zeyuan Allen-Zhu, Yuanzhi Li, Shean Wang, Lu~Wang, and Weizhu Chen.
\newblock Lo{RA}: Low-rank adaptation of large language models.
\newblock In \emph{Proceedings of the International Conference on Learning Representations}, 2022.

\bibitem[Hu et~al.(2024)Hu, Shu, Yu, Wu, Lin, Dai, Ng, and Low]{hu2024localizedzerothorderpromptoptimization}
Wenyang Hu, Yao Shu, Zongmin Yu, Zhaoxuan Wu, Xiangqiang Lin, Zhongxiang Dai, See-Kiong Ng, and Bryan Kian~Hsiang Low.
\newblock Localized zeroth-order prompt optimization.
\newblock In \emph{Proceedings of the International Conference on Machine Learning}, 2024.

\bibitem[Johnson \& Lindenstrauss(1984)Johnson and Lindenstrauss]{jllemma84}
William Johnson and Joram Lindenstrauss.
\newblock Extensions of {L}ipschitz maps into a {H}ilbert space.
\newblock \emph{Contemporary Mathematics}, 1984.

\bibitem[Koh \& Liang(2017)Koh and Liang]{koh2017}
Pang~Wei Koh and Percy Liang.
\newblock Understanding black-box predictions via influence functions.
\newblock In \emph{Proceedings of the International Conference on Machine Learning}, 2017.

\bibitem[Kwon et~al.(2024)Kwon, Wu, Wu, and Zou]{kwon2023datainf}
Yongchan Kwon, Eric Wu, Kevin Wu, and James Zou.
\newblock Data{I}nf: Efficiently estimating data influence in {LoRA}-tuned {LLMs} and diffusion models.
\newblock In \emph{Proceedings of the International Conference on Learning Representations}, 2024.

\bibitem[Lhoest et~al.(2021)Lhoest, Villanova~del Moral, Jernite, Thakur, von Platen, Patil, Chaumond, Drame, Plu, Tunstall, Davison, {\v{S}}a{\v{s}}ko, Chhablani, Malik, Brandeis, Le~Scao, Sanh, Xu, Patry, McMillan-Major, Schmid, Gugger, Delangue, Matussi{\`e}re, Debut, Bekman, Cistac, Goehringer, Mustar, Lagunas, Rush, and Wolf]{lhoest-etal-2021-datasets}
Quentin Lhoest, Albert Villanova~del Moral, Yacine Jernite, Abhishek Thakur, Patrick von Platen, Suraj Patil, Julien Chaumond, Mariama Drame, Julien Plu, Lewis Tunstall, Joe Davison, Mario {\v{S}}a{\v{s}}ko, Gunjan Chhablani, Bhavitvya Malik, Simon Brandeis, Teven Le~Scao, Victor Sanh, Canwen Xu, Nicolas Patry, Angelina McMillan-Major, Philipp Schmid, Sylvain Gugger, Cl{\'e}ment Delangue, Th{\'e}o Matussi{\`e}re, Lysandre Debut, Stas Bekman, Pierric Cistac, Thibault Goehringer, Victor Mustar, Fran{\c{c}}ois Lagunas, Alexander Rush, and Thomas Wolf.
\newblock Datasets: A community library for natural language processing.
\newblock In \emph{Proceedings of the Conference on Empirical Methods in Natural Language Processing}, 2021.

\bibitem[Li et~al.(2023)Li, Ildiz, Papailiopoulos, and Oymak]{li2023transformer}
Yingcong Li, M.~Emrullah Ildiz, Dimitris Papailiopoulos, and Samet Oymak.
\newblock Transformers as algorithms: generalization and stability in in-context learning.
\newblock In \emph{Proceedings of the International Conference on Machine Learning}, 2023.

\bibitem[Lin et~al.(2024{\natexlab{a}})Lin, Dai, Verma, Ng, Jaillet, and Low]{lin2024promptoptimizationhumanfeedback}
Xiaoqiang Lin, Zhongxiang Dai, Arun Verma, See-Kiong Ng, Patrick Jaillet, and Bryan Kian~Hsiang Low.
\newblock Prompt optimization with human feedback.
\newblock In \emph{ICML Workshop on Models of Human Feedback for AI Alignment}, 2024{\natexlab{a}}.

\bibitem[Lin et~al.(2024{\natexlab{b}})Lin, Wu, Dai, Hu, Shu, Ng, Jaillet, and Low]{lin2024useinstinctinstructionoptimization}
Xiaoqiang Lin, Zhaoxuan Wu, Zhongxiang Dai, Wenyang Hu, Yao Shu, See-Kiong Ng, Patrick Jaillet, and Bryan Kian~Hsiang Low.
\newblock Use your instinct: Instruction optimization for llms using neural bandits coupled with transformers.
\newblock In \emph{Proceedings of the International Conference on Machine Learning}, 2024{\natexlab{b}}.

\bibitem[Lin et~al.(2024{\natexlab{c}})Lin, Xu, Wu, Ng, and Low]{lin2024distributionally}
Xiaoqiang Lin, Xinyi Xu, Zhaoxuan Wu, See-Kiong Ng, and Bryan Kian~Hsiang Low.
\newblock Distributionally robust data valuation.
\newblock In \emph{Proceedings of the International Conference on Machine Learning}, 2024{\natexlab{c}}.

\bibitem[Liu et~al.(2023{\natexlab{a}})Liu, Mao, Xia, Lou, and Liu]{liu2023prompt}
Hanxi Liu, Xiaokai Mao, Haocheng Xia, Jian Lou, and Jinfei Liu.
\newblock Prompt valuation based on {S}hapley values, 2023{\natexlab{a}}.

\bibitem[Liu et~al.(2023{\natexlab{b}})Liu, Lin, Hewitt, Paranjape, Bevilacqua, Petroni, and Liang]{liu2023lost}
Nelson~F. Liu, Kevin Lin, John Hewitt, Ashwin Paranjape, Michele Bevilacqua, Fabio Petroni, and Percy Liang.
\newblock Lost in the middle: How language models use long contexts.
\newblock \emph{Transactions of the Association for Computational Linguistics}, 2023{\natexlab{b}}.

\bibitem[Liu et~al.(2023{\natexlab{c}})Liu, Zhang, and Liang]{liu2023evaluating}
Nelson~F. Liu, Tianyi Zhang, and Percy Liang.
\newblock Evaluating verifiability in generative search engines, 2023{\natexlab{c}}.

\bibitem[Lu et~al.(2022)Lu, Bartolo, Moore, Riedel, and Stenetorp]{lu2022fantastically}
Yao Lu, Max Bartolo, Alastair Moore, Sebastian Riedel, and Pontus Stenetorp.
\newblock Fantastically ordered prompts and where to find them: Overcoming few-shot prompt order sensitivity.
\newblock In \emph{Proceedings of the Annual Meeting of the Association for Computational Linguistics}, 2022.

\bibitem[Machiraju et~al.(2024)Machiraju, Derry, Desai, Guha, Karimi, Zou, Altman, Ré, and Mallick]{machiraju2024prospector}
Gautam Machiraju, Alexander Derry, Arjun Desai, Neel Guha, Amir-Hossein Karimi, James Zou, Russ Altman, Christopher Ré, and Parag Mallick.
\newblock Prospector heads: Generalized feature attribution for large models \& data, 2024.

\bibitem[Mitchell et~al.(2022)Mitchell, Cooper, Frank, and Holmes]{mitchellsobol22}
Rory Mitchell, Joshua Cooper, Eibe Frank, and Geoffrey Holmes.
\newblock Sampling permutations for {S}hapley value estimation.
\newblock \emph{Journal of Machine Learning Research}, 2022.

\bibitem[Murphy(2012)]{murphy22}
Kevin~P. Murphy.
\newblock \emph{Machine Learning: A Probabilistic Perspective}.
\newblock The MIT Press, 2012.

\bibitem[Nguyen \& Wong(2023)Nguyen and Wong]{nguyen2023incontext}
Tai Nguyen and Eric Wong.
\newblock In-context example selection with influences, 2023.

\bibitem[Olsson et~al.(2022)Olsson, Elhage, Nanda, Joseph, DasSarma, Henighan, Mann, Askell, Bai, Chen, Conerly, Drain, Ganguli, Hatfield-Dodds, Hernandez, Johnston, Jones, Kernion, Lovitt, Ndousse, Amodei, Brown, Clark, Kaplan, McCandlish, and Olah]{olsson2022incontext}
Catherine Olsson, Nelson Elhage, Neel Nanda, Nicholas Joseph, Nova DasSarma, Tom Henighan, Ben Mann, Amanda Askell, Yuntao Bai, Anna Chen, Tom Conerly, Dawn Drain, Deep Ganguli, Zac Hatfield-Dodds, Danny Hernandez, Scott Johnston, Andy Jones, Jackson Kernion, Liane Lovitt, Kamal Ndousse, Dario Amodei, Tom Brown, Jack Clark, Jared Kaplan, Sam McCandlish, and Chris Olah.
\newblock In-context learning and induction heads, 2022.

\bibitem[Pang \& Lee(2005)Pang and Lee]{pang05a}
Bo~Pang and Lillian Lee.
\newblock Seeing stars: Exploiting class relationships for sentiment categorization with respect to rating scales.
\newblock In \emph{Proceedings of the Annual Meeting of the Association for Computational Linguistics}, 2005.

\bibitem[Panwar et~al.(2024)Panwar, Ahuja, and Goyal]{panwar2024incontext}
Madhur Panwar, Kabir Ahuja, and Navin Goyal.
\newblock In-context learning through the bayesian prism.
\newblock In \emph{Proceedings of the International Conference on Learning Representations}, 2024.

\bibitem[Ribeiro et~al.(2016)Ribeiro, Singh, and Guestrin]{lime2016}
Marco~Tulio Ribeiro, Sameer Singh, and Carlos Guestrin.
\newblock "{W}hy should {I} trust you?": Explaining the predictions of any classifier.
\newblock In \emph{Proceedings of the {ACM} {SIGKDD} International Conference on Knowledge Discovery and Data Mining}, 2016.

\bibitem[S. et~al.(2024)S., Van, and Wu]{s2024incontext}
Vinay~M. S., Minh-Hao Van, and Xintao Wu.
\newblock In-context learning demonstration selection via influence analysis, 2024.

\bibitem[Sarti et~al.(2023)Sarti, Feldhus, Sickert, van~der Wal, Nissim, and Bisazza]{inseq2023}
Gabriele Sarti, Nils Feldhus, Ludwig Sickert, Oskar van~der Wal, Malvina Nissim, and Arianna Bisazza.
\newblock Inseq: An interpretability toolkit for sequence generation models.
\newblock In \emph{Proceedings of the Annual Meeting of the Association for Computational Linguistics}, 2023.

\bibitem[Sharma et~al.(2020)Sharma, Ramakrishnan, Prakash, Siew~Kei, and Srikanthan]{kuldeep2020}
Kuldeep Sharma, Nirmala Ramakrishnan, Alok Prakash, Lam Siew~Kei, and Thambipillai Srikanthan.
\newblock Evaluating the merits of ranking in structured network pruning.
\newblock In \emph{2020 IEEE 40th International Conference on Distributed Computing Systems (ICDCS)}, 2020.

\bibitem[Sim et~al.(2024)Sim, Fan, Tian, Jaillet, and Low]{pmlr-v235-sim24a}
Rachael Hwee~Ling Sim, Jue Fan, Xiao Tian, Patrick Jaillet, and Bryan Kian~Hsiang Low.
\newblock Deletion-anticipative data selection with a limited budget.
\newblock In \emph{Proceedings of the International Conference on Machine Learning}, 2024.

\bibitem[Socher et~al.(2013)Socher, Perelygin, Wu, Chuang, Manning, Ng, and Potts]{socher-etal-2013-recursive}
Richard Socher, Alex Perelygin, Jean Wu, Jason Chuang, Christopher~D. Manning, Andrew Ng, and Christopher Potts.
\newblock Recursive deep models for semantic compositionality over a sentiment treebank.
\newblock In \emph{Proceedings of the Conference on Empirical Methods in Natural Language Processing}, 2013.

\bibitem[Sundararajan et~al.(2017)Sundararajan, Taly, and Yan]{sundararajan17}
Mukund Sundararajan, Ankur Taly, and Qiqi Yan.
\newblock Axiomatic attribution for deep networks.
\newblock In \emph{Proceedings of the International Conference on Machine Learning}, 2017.

\bibitem[Tanaka et~al.(2020)Tanaka, Kunin, Yamins, and Ganguli]{tankapruning20}
Hidenori Tanaka, Daniel Kunin, Daniel L.~K. Yamins, and Surya Ganguli.
\newblock Pruning neural networks without any data by iteratively conserving synaptic flow.
\newblock In \emph{Advances in Neural Information Processing Systems}, 2020.

\bibitem[Touvron et~al.(2023)Touvron, Martin, Stone, Albert, Almahairi, Babaei, Bashlykov, Batra, Bhargava, Bhosale, Bikel, Blecher, Ferrer, Chen, Cucurull, Esiobu, Fernandes, Fu, Fu, Fuller, Gao, Goswami, Goyal, Hartshorn, Hosseini, Hou, Inan, Kardas, Kerkez, Khabsa, Kloumann, Korenev, Koura, Lachaux, Lavril, Lee, Liskovich, Lu, Mao, Martinet, Mihaylov, Mishra, Molybog, Nie, Poulton, Reizenstein, Rungta, Saladi, Schelten, Silva, Smith, Subramanian, Tan, Tang, Taylor, Williams, Kuan, Xu, Yan, Zarov, Zhang, Fan, Kambadur, Narang, Rodriguez, Stojnic, Edunov, and Scialom]{touvron2023llama}
Hugo Touvron, Louis Martin, Kevin Stone, Peter Albert, Amjad Almahairi, Yasmine Babaei, Nikolay Bashlykov, Soumya Batra, Prajjwal Bhargava, Shruti Bhosale, Dan Bikel, Lukas Blecher, Cristian~Canton Ferrer, Moya Chen, Guillem Cucurull, David Esiobu, Jude Fernandes, Jeremy Fu, Wenyin Fu, Brian Fuller, Cynthia Gao, Vedanuj Goswami, Naman Goyal, Anthony Hartshorn, Saghar Hosseini, Rui Hou, Hakan Inan, Marcin Kardas, Viktor Kerkez, Madian Khabsa, Isabel Kloumann, Artem Korenev, Punit~Singh Koura, Marie-Anne Lachaux, Thibaut Lavril, Jenya Lee, Diana Liskovich, Yinghai Lu, Yuning Mao, Xavier Martinet, Todor Mihaylov, Pushkar Mishra, Igor Molybog, Yixin Nie, Andrew Poulton, Jeremy Reizenstein, Rashi Rungta, Kalyan Saladi, Alan Schelten, Ruan Silva, Eric~Michael Smith, Ranjan Subramanian, Xiaoqing~Ellen Tan, Binh Tang, Ross Taylor, Adina Williams, Jian~Xiang Kuan, Puxin Xu, Zheng Yan, Iliyan Zarov, Yuchen Zhang, Angela Fan, Melanie Kambadur, Sharan Narang, Aurelien Rodriguez, Robert Stojnic, Sergey Edunov, and Thomas
  Scialom.
\newblock Llama 2: Open foundation and fine-tuned chat models, 2023.

\bibitem[Von~Oswald et~al.(2023)Von~Oswald, Niklasson, Randazzo, Sacramento, Mordvintsev, Zhmoginov, and Vladymyrov]{oswald2023}
Johannes Von~Oswald, Eyvind Niklasson, Ettore Randazzo, Jo\~{a}o Sacramento, Alexander Mordvintsev, Andrey Zhmoginov, and Max Vladymyrov.
\newblock Transformers learn in-context by gradient descent.
\newblock In \emph{Proceedings of the International Conference on Machine Learning}, 2023.

\bibitem[von Oswald et~al.(2023)von Oswald, Niklasson, Schlegel, Kobayashi, Zucchet, Scherrer, Miller, Sandler, y~Arcas, Vladymyrov, Pascanu, and Sacramento]{vonoswald2023uncovering}
Johannes von Oswald, Eyvind Niklasson, Maximilian Schlegel, Seijin Kobayashi, Nicolas Zucchet, Nino Scherrer, Nolan Miller, Mark Sandler, Blaise~Agüera y~Arcas, Max Vladymyrov, Razvan Pascanu, and João Sacramento.
\newblock Uncovering mesa-optimization algorithms in transformers, 2023.

\bibitem[Wang et~al.(2024)Wang, Lin, Qiao, Foo, and Low]{wang2024helpfulharmfuldatafinetuningfree}
Jingtan Wang, Xiaoqiang Lin, Rui Qiao, Chuan-Sheng Foo, and Bryan Kian~Hsiang Low.
\newblock Helpful or harmful data? fine-tuning-free shapley attribution for explaining language model predictions.
\newblock In \emph{Proceedings of the International Conference on Machine Learning}, 2024.

\bibitem[Wang et~al.(2023)Wang, Li, Dai, Chen, Zhou, Meng, Zhou, and Sun]{wang-etal-2023-label}
Lean Wang, Lei Li, Damai Dai, Deli Chen, Hao Zhou, Fandong Meng, Jie Zhou, and Xu~Sun.
\newblock Label words are anchors: An information flow perspective for understanding in-context learning.
\newblock In \emph{Proceedings of the Conference on Empirical Methods in Natural Language Processing}, 2023.

\bibitem[Wei et~al.(2023)Wei, Wang, Schuurmans, Bosma, Ichter, Xia, Chi, Le, and Zhou]{wei2023chainofthought}
Jason Wei, Xuezhi Wang, Dale Schuurmans, Maarten Bosma, Brian Ichter, Fei Xia, Ed~Chi, Quoc Le, and Denny Zhou.
\newblock Chain-of-thought prompting elicits reasoning in large language models.
\newblock In \emph{Advances in Neural Information Processing Systems}, 2023.

\bibitem[Wolf et~al.(2020)Wolf, Debut, Sanh, Chaumond, Delangue, Moi, Cistac, Rault, Louf, Funtowicz, Davison, Shleifer, von Platen, Ma, Jernite, Plu, Xu, Scao, Gugger, Drame, Lhoest, and Rush]{wolf-etal-2020-transformers}
Thomas Wolf, Lysandre Debut, Victor Sanh, Julien Chaumond, Clement Delangue, Anthony Moi, Pierric Cistac, Tim Rault, Rémi Louf, Morgan Funtowicz, Joe Davison, Sam Shleifer, Patrick von Platen, Clara Ma, Yacine Jernite, Julien Plu, Canwen Xu, Teven~Le Scao, Sylvain Gugger, Mariama Drame, Quentin Lhoest, and Alexander~M. Rush.
\newblock Transformers: State-of-the-art natural language processing.
\newblock In \emph{Proceedings of the Conference on Empirical Methods in Natural Language Processing}, 2020.

\bibitem[Wu et~al.(2024)Wu, Lin, Dai, Hu, Shu, Ng, Jaillet, and Low]{wu2024promptoptimizationeaseefficient}
Zhaoxuan Wu, Xiaoqiang Lin, Zhongxiang Dai, Wenyang Hu, Yao Shu, See-Kiong Ng, Patrick Jaillet, and Bryan Kian~Hsiang Low.
\newblock Prompt optimization with ease? efficient ordering-aware automated selection of exemplars.
\newblock In \emph{Advances in Neural Information Processing Systems}, 2024.

\bibitem[Xia et~al.(2024)Xia, Malladi, Gururangan, Arora, and Chen]{xia2024less}
Mengzhou Xia, Sadhika Malladi, Suchin Gururangan, Sanjeev Arora, and Danqi Chen.
\newblock Less: Selecting influential data for targeted instruction tuning, 2024.

\bibitem[Xie et~al.(2022)Xie, Raghunathan, Liang, and Ma]{xie2022incontext}
Sang~Michael Xie, Aditi Raghunathan, Percy Liang, and Tengyu Ma.
\newblock An explanation of in-context learning as implicit {B}ayesian inference.
\newblock In \emph{Proceedings of the International Conference on Learning Representations}, 2022.

\bibitem[Xu et~al.(2024{\natexlab{a}})Xu, Wang, Foo, Low, and Fanti]{xu2024datadistributionvaluation}
Xinyi Xu, Shuaiqi Wang, Chuan-Sheng Foo, Bryan Kian~Hsiang Low, and Giulia Fanti.
\newblock Data distribution valuation.
\newblock In \emph{Advances in Neural Information Processing Systems}, 2024{\natexlab{a}}.

\bibitem[Xu et~al.(2024{\natexlab{b}})Xu, Wu, Qiao, Verma, Shu, Wang, Niu, He, Chen, Zhou, Lau, Dao, Agussurja, Sim, Lin, Hu, Dai, Koh, and Low]{xu2024datacentricaiagelarge}
Xinyi Xu, Zhaoxuan Wu, Rui Qiao, Arun Verma, Yao Shu, Jingtan Wang, Xinyuan Niu, Zhenfeng He, Jiangwei Chen, Zijian Zhou, Gregory Kang~Ruey Lau, Hieu Dao, Lucas Agussurja, Rachael Hwee~Ling Sim, Xiaoqiang Lin, Wenyang Hu, Zhongxiang Dai, Pang~Wei Koh, and Bryan Kian~Hsiang Low.
\newblock Data-centric ai in the age of large language models.
\newblock In \emph{Conference on Empirical Methods in Natural Language Processing}, 2024{\natexlab{b}}.

\bibitem[Yao et~al.(2023)Yao, Yu, Zhao, Shafran, Griffiths, Cao, and Narasimhan]{yao2023}
Shunyu Yao, Dian Yu, Jeffrey Zhao, Izhak Shafran, Thomas~L. Griffiths, Yuan Cao, and Karthik Narasimhan.
\newblock {Tree of Thoughts}: Deliberate problem solving with large language models.
\newblock In \emph{Advances in Neural Information Processing Systems}, 2023.

\bibitem[Yue et~al.(2023)Yue, Wang, Zhang, Chen, Su, and Sun]{yue2023automatic}
Xiang Yue, Boshi Wang, Kai Zhang, Ziru Chen, Yu~Su, and Huan Sun.
\newblock Automatic evaluation of attribution by large language models.
\newblock \emph{arXiv preprint arXiv:2305.06311}, 2023.

\bibitem[Zhang et~al.(2024{\natexlab{a}})Zhang, Frei, and Bartlett]{zhang2024icl}
Ruiqi Zhang, Spencer Frei, and Peter~L. Bartlett.
\newblock Trained transformers learn linear models in-context.
\newblock \emph{Journal of Machine Learning Research}, 2024{\natexlab{a}}.

\bibitem[Zhang et~al.(2024{\natexlab{b}})Zhang, Xia, Wang, Chen, Liu, Wu, and Liu]{zhang2024ideal}
Shaokun Zhang, Xiaobo Xia, Zhaoqing Wang, Ling-Hao Chen, Jiale Liu, Qingyun Wu, and Tongliang Liu.
\newblock Ideal: Influence-driven selective annotations empower in-context learners in large language models.
\newblock In \emph{Proceedings of the International Conference on Learning Representations}, 2024{\natexlab{b}}.

\bibitem[Zhang et~al.(2020)Zhang, Kishore, Wu, Weinberger, and Artzi]{Zhang2019BERTScoreET}
Tianyi Zhang, Varsha Kishore, Felix Wu, Kilian~Q. Weinberger, and Yoav Artzi.
\newblock Bertscore: Evaluating text generation with bert.
\newblock In \emph{Proceedings of the International Conference on Learning Representations}, 2020.

\bibitem[Zhang et~al.(2015)Zhang, Zhao, and LeCun]{Zhang2015CharacterlevelCN}
Xiang Zhang, Junbo~Jake Zhao, and Yann LeCun.
\newblock Character-level convolutional networks for text classification.
\newblock In \emph{Advances in Neural Information Processing Systems}, 2015.

\bibitem[Zheng et~al.(2023)Zheng, Chiang, Sheng, Zhuang, Wu, Zhuang, Lin, Li, Li, Xing, Zhang, Gonzalez, and Stoica]{zheng2023judging}
Lianmin Zheng, Wei-Lin Chiang, Ying Sheng, Siyuan Zhuang, Zhanghao Wu, Yonghao Zhuang, Zi~Lin, Zhuohan Li, Dacheng Li, Eric.~P Xing, Hao Zhang, Joseph~E. Gonzalez, and Ion Stoica.
\newblock Judging {LLM}-as-a-judge with {MT}-bench and chatbot arena.
\newblock In \emph{Advances in Neural Information Processing Systems}, 2023.

\bibitem[Zhou et~al.(2023)Zhou, Xu, Sim, Foo, and Low]{zhou2022probablyapproximateshapleyfairness}
Zijian Zhou, Xinyi Xu, Rachael Hwee~Ling Sim, Chuan~Sheng Foo, and Kian~Hsiang Low.
\newblock Probably approximate shapley fairness with applications in machine learning.
\newblock In \emph{AAAI Conference on Artificial Intelligence}, 2023.

\end{thebibliography}
}

\appendix
\onecolumn

\section{Computational Resources} \label{app:compute}

\paragraph{Hardware.} All our experiments about 7B white-box models are conducted on a single L40 GPU. All experiments involving 13B white-box models are conducted on a single H100 GPU. Total budget for GPT-3.5 API calls to conduct the transferability experiments in \cref{sec:transfer} is estimated to be around US\$500 (at the rate of US\$0.5/1mln tokens for input and US\$1.50/1mln tokens for output).

\paragraph{Software. } All our experiments are conducted using Python3.10 on a Ubuntu 22.04.4 LTS distribution. We use Jax~\citep{jax2018github} for the experiments in \cref{sec:eval_custom_tf} and use PyTorch 2.1.0~\citep{pytorch2024} for the experiments in \cref{sec:eval_llm} and \cref{sec:exp_application}. We adopt the implementations of the LLMs (transformer-based and SSM-based) provided in the Huggingface's ``transformers''~\citep{wolf-etal-2020-transformers} system throughout this work. The NLP datasets are also obtained from Huggingface's ``datasets'' API~\citep{lhoest-etal-2021-datasets}. The precise repository references and other dependencies can be found in the code provided in the supplemental materials. 

\section{Algorithmic Implementation} \label{app:algo}

We provide an implementation of computing our \texttt{DETAIL} score for LLM in \cref{algo:infl}. Line 6 shows the (optional) random projection where a random matrix $P$ is multiplied by the embeddings. In line 8, $\beta$ has a closed-form expression as the solution to a regularized ridge regression. In line 10 and line 14, we select the embeddings of the target positions. Then, depending on whether we use self-influence, we use different embeddings for computing $\nabla_\beta L$ as shown in lines 15-19. Line 20 computes the inverse hessian $H_\beta^{-1} = (K_{\cI} + \lambda I)^{-1}$ before finally calculating $\cI_i$ in line 21.

\begin{algorithm}[!ht]
\caption
{\texttt{DETAIL}}\label{algo:infl}
\begin{algorithmic}[1]
\State {\bfseries Input: } model $M$, prompt tokens $x_{[1:n]}$,  label tokens $y_{[1:t]}$, target positions $p_{[1:t]}$, total number of transformer layers $L$, transformer layer to compute \texttt{DETAIL} score $l$, regularization constant $\lambda$, projection matrix $P \in \mathbb{R}^{d \times d'}$ (default $I$)
\State $\cI_i \gets 0$ for $i \in \{1,2,\cdots,t-1\}$
\State $h_1, h_2, \cdots, h_L \gets M(x_{[1:n]})$ 
\State $p_{\text{demo}} \gets p_{[1:t-1]}$ \Comment{Remove the last target which is the test query}
\State $y_{\text{demo}} \gets \text{one\_hot}(y_{[1:t-1]})$ \Comment{Remove the last label and convert to one-hot}
\State $m_{\text{demo}} \gets h_l[p_{\text{demo}}]P$ \Comment{Optional dimensionality reduction}
\State $K_{\beta} \gets m_{\text{demo}}m_{\text{demo}}^\top$ \Comment{$K_{\beta} \in \mathbb{R}^{(t-1) \times (t-1)}$ for speed-up as $t \ll d'$}
\State $\beta \gets [(K_{\beta} + \lambda{I})^{-1}m_{\text{demo}}]^\top y_{\text{demo}}$
\State $p_{\text{test}} \gets p_{[t-1:t]}$
\State $m_{\text{test}} \gets h_l[p_{\text{test}}]$
\State $y_{\text{test}} \gets \text{one\_hot}(y_{[t-1:t]})$
\State $K_{\cI} \gets m_{\text{demo}}^\top m_{\text{demo}}$ \Comment{$K_{\cI} \in \mathbb{R}^{d' \times d'}$}
\For{$i \in \{1,2,\cdots,t-1\}$}
\State $m_i \gets h_l[p_{[i:i+1]}]P$
\If{self influence}
\State $\nabla_{\beta}L \gets m_{i}^\top(m_{i} \beta - y_{\text{demo}}[i]) + \lambda \beta$
\Else
\State $\nabla_{\beta}L \gets m_{\text{test}}^\top(m_{\text{test}} \beta - y_{\text{test}}) + \lambda \beta$
\EndIf
\State $\cI_{\text{reg}} \gets (K_{\cI} + \lambda {I})^{-1}[m_i^\top (m_i\beta - y_{\text{demo}}[i]) + \lambda \beta]$ \Comment{\cref{eq:infl_reg} with the constant dropped}
\State $\cI_i \gets \cI_i + (\nabla_{\beta}L)^\top \cI_{\text{reg}}$ \Comment{\cref{eq:infl_kernel}}
\EndFor
\State {\bfseries Return} $\cI$
\end{algorithmic}
\end{algorithm}

\section{Additional Discussion} \label{app:discussion}

\paragraph{Potential societal impact.}

We propose an attribution technique for improving the interpretability of in-context learning. We believe our research has potential positive societal impacts in improving the safety of LLMs via filtering out corrupted/harmful demonstrations as demonstrated by our experiments as well as saving energy by curating the demonstration, hence reducing the cost of querying LLMs. We do not find any direct negative societal impact posed by our research contribution. 

\paragraph{Setting $\lambda$.} Generally, there is no golden rule for the most appropriate $\lambda$ that regularizes the ridge parameters $\beta$. Intuitively, a larger $\lambda$ likely works better when the dimension of $\beta$ is large since the model tends to be over-parameterized (i.e., in a LLM). Therefore, we set a relatively large $\lambda=1.0$ for LLMs and a relatively small $\lambda=0.01$ for our custom transformer. When detecting noisy demonstrations, we may not want to regularize $\beta$ too much because we wish to retain the information captured by the eigenvalues of the hessian $H$ which can be eroded with a larger $\lambda$. As such, for the noisy demonstration detection task, we set a very small $\lambda=10^{-9}$ to retain most of the information captured by $H$ while ensuring that it is invertible.

\paragraph{Random projection matrix.} We recall the Johnson-Lindenstrauss (JL) lemma~\citep{dasgupta2003elementary,jllemma84}.
\begin{theorem}[Johnson-Lindenstrauss Lemma] \label{thm:wang2016}
For any $0 < \epsilon < 1$ and any integer $n$, let $d'$ be a positive integer such that
  \begin{equation*}
      d' \geq \frac{24}{3\epsilon^2 - 2\epsilon^3}\log{n}\ ,
  \end{equation*}
then for any set $A$ of $n$ points $\in \mathbb{R}^d$, there exists a mapping $f: \mathbb{R}^d \to \mathbb{R}^{d'}$ such that for all $x_i, x_j \in A$,
\begin{equation*}
    (1 - \epsilon) \Vert x_i - x_j \Vert^2 \leq \Vert f(x_i) - f(x_j) \Vert^2 \leq (1+\epsilon) \Vert x_i - x_j \Vert^2\ .
\end{equation*}
\end{theorem}
A specific constructive proof is by setting $A \coloneqq \frac{1}{\sqrt{d'}}R$ where $R_{i,j} \overset{\text{i.i.d.}}{\sim} \cN(0, 1)$.\footnote{\href{https://cs.stanford.edu/people/mmahoney/cs369m/Lectures/lecture1.pdf}{Lecture notes.}} In our work, we treat each embedding $m$ as $x$ and the projected embedding $mP$ as $f(x)$. The specific construction follows the abovementioned constructive proof defining 
\[P \coloneqq \frac{1}{\sqrt{d'}}R \sim \cN(\boldsymbol{0}, \frac{1}{d'} \boldsymbol{I}) \ . \] 
Empirically, our threshold $d' = 1000$ corresponds to $\epsilon \lessapprox 0.164$, ensuring a good preservation of (Euclidean) distance between points.

\paragraph{ICL prompt example.} We include a visualization of a prompt for ICL below. Each input-output pair consists of a task demonstration. The query is appended at the end of the prompt with only the input and the output header.

\begin{mycolorbox}{Prompt}{Subj}
\label{colorbox:subj}
    Input: tsai may be ploughing the same furrow once too often .\\
    Output: B\\\\
    Input: equilibrium the movie , as opposed to the manifesto , is really , really stupid .\\
    Output: B\\\\
    <More demonstrations...> \\\\
    Input: a friendly vacation for four old friends - two couples from college - turns ugly . . . then\\
    Output: A\\\\
    Input: he meets god and is given all the powers of god .\\
    Output:
\end{mycolorbox}

\paragraph{Additional potential future directions.} Apart from applying \texttt{DETAIL} to more generalized prompting settings, we think it is also an interesting direction to research whether \texttt{DETAIL} can provide meaningful attribution for pruned~\citep{kuldeep2020,tankapruning20} (or distilled~\citep{hinton2015distillingknowledgeneuralnetwork,hsieh2023distillingstepbystepoutperforminglarger}) networks. This is especially useful since, compared to very large models, pruned small networks can be deployed privately, admitting a straightforward application of \texttt{DETAIL}. Moreover, the size of the hidden states tends to be smaller, allowing for faster computation of \texttt{DETAIL} score.

\section{Additional Experiments}\label{app:exp}

\subsection{Additional Results for Demonstration Perturbation Task} \label{app:exp_sample_perturb_additional}

\begin{figure}[H]
\centering
\begin{subfigure}[t]{0.22\textwidth}
    \includegraphics[width=\textwidth]{figs/cls_llm_ag_news_vicuna-7b_acc_corrupt.pdf}
    \end{subfigure}
\hfill
\begin{subfigure}[t]{0.23\textwidth}
    \includegraphics[width=\textwidth]{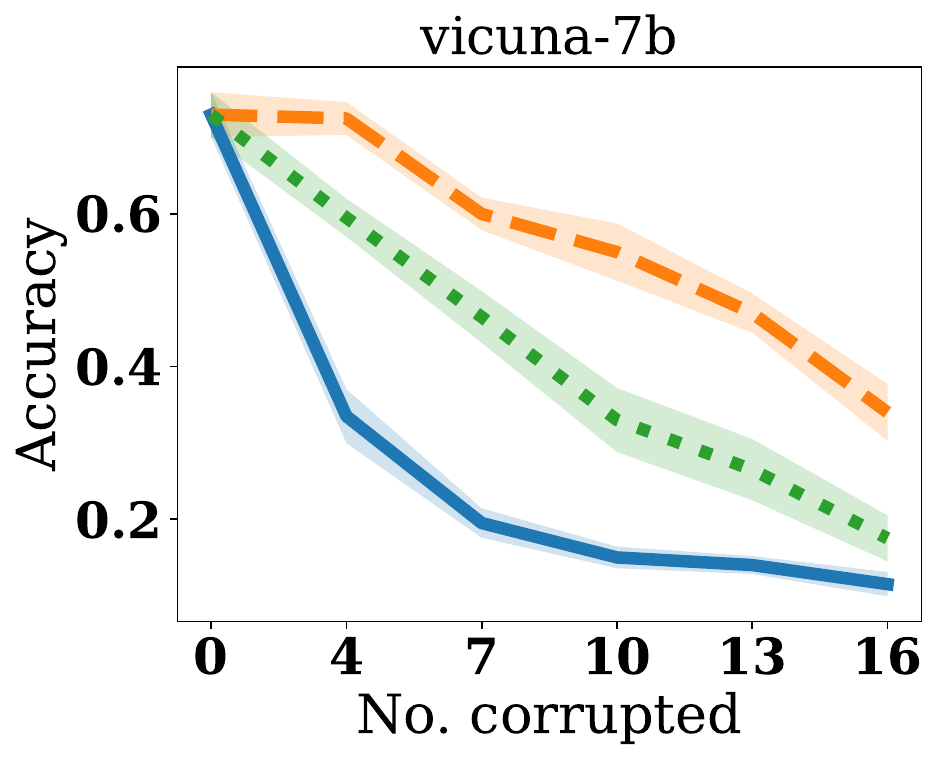}
    \end{subfigure}
\hfill
\begin{subfigure}[t]{0.23\textwidth}
    \includegraphics[width=\textwidth]{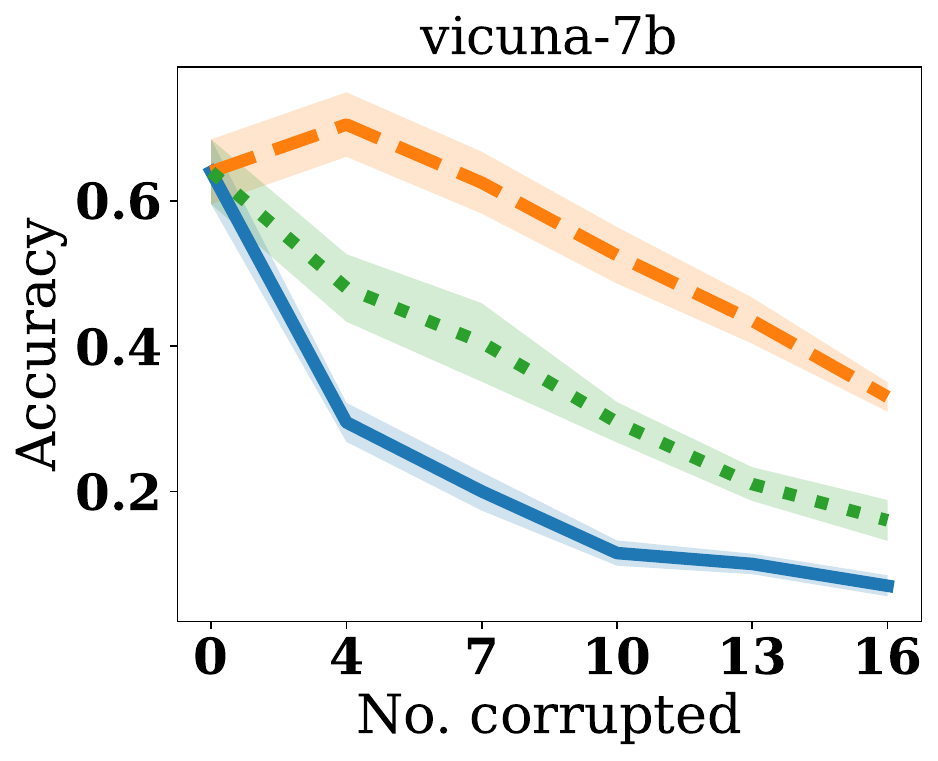}
    \end{subfigure}
\hfill
\begin{subfigure}[t]{0.23\textwidth}
    \includegraphics[width=\textwidth]{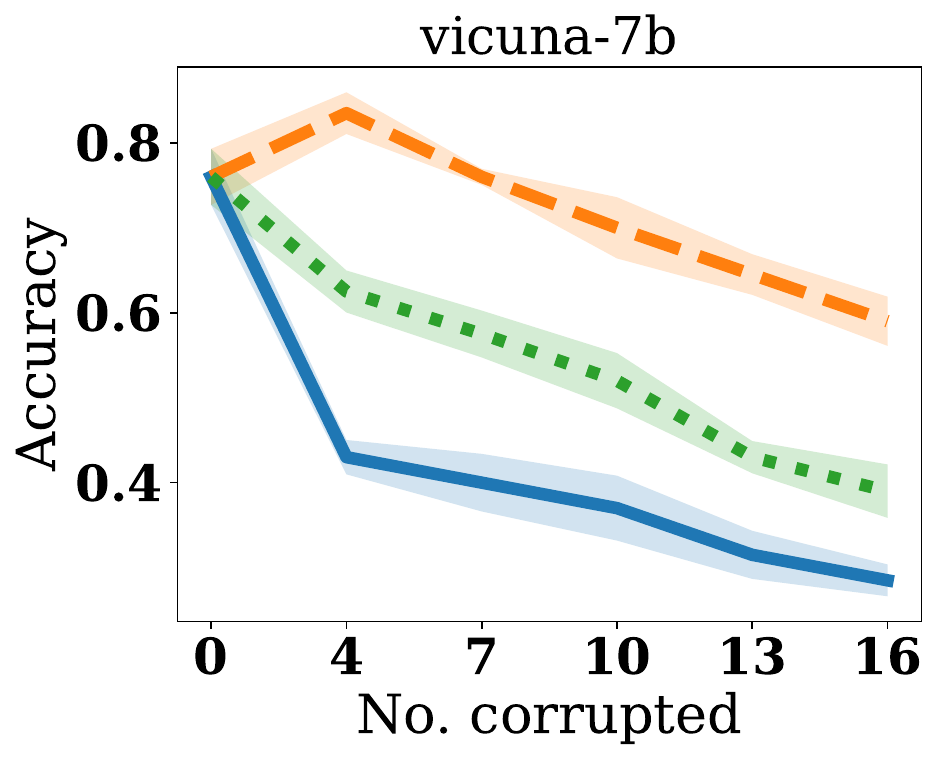}
    \end{subfigure}
\hfill
\begin{subfigure}[t]{0.23\textwidth}
    \includegraphics[width=\textwidth]{figs/cls_llm_ag_news_vicuna-7b_acc_remove.pdf}
    \end{subfigure}
\hfill
\begin{subfigure}[t]{0.23\textwidth}
    \includegraphics[width=\textwidth]{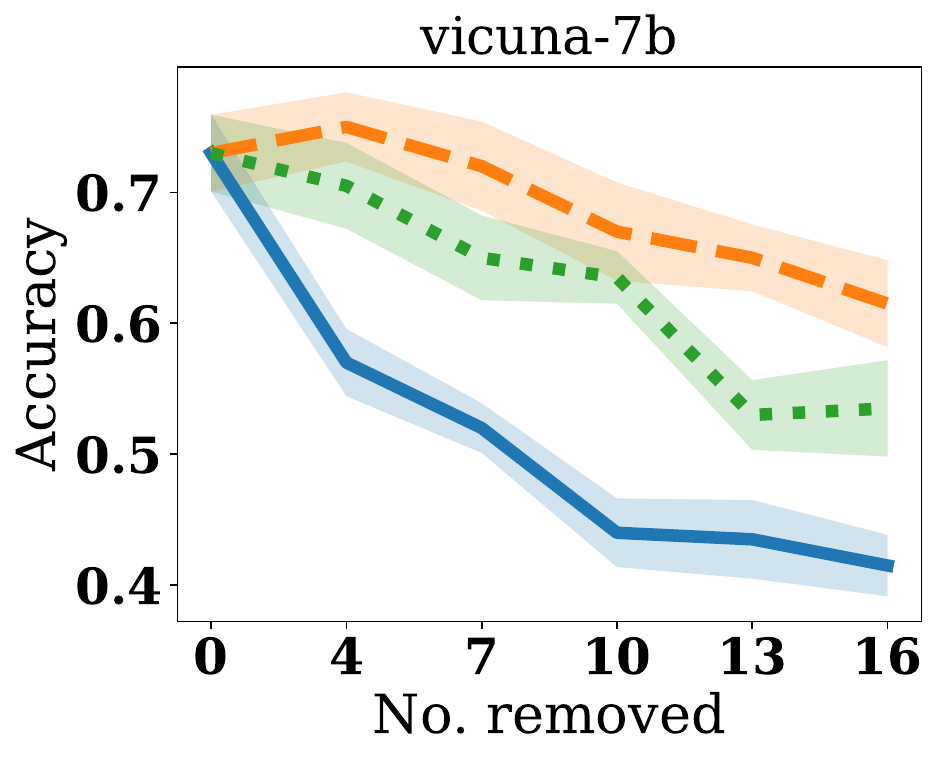}
    \end{subfigure}
\hfill
\begin{subfigure}[t]{0.23\textwidth}
    \includegraphics[width=\textwidth]{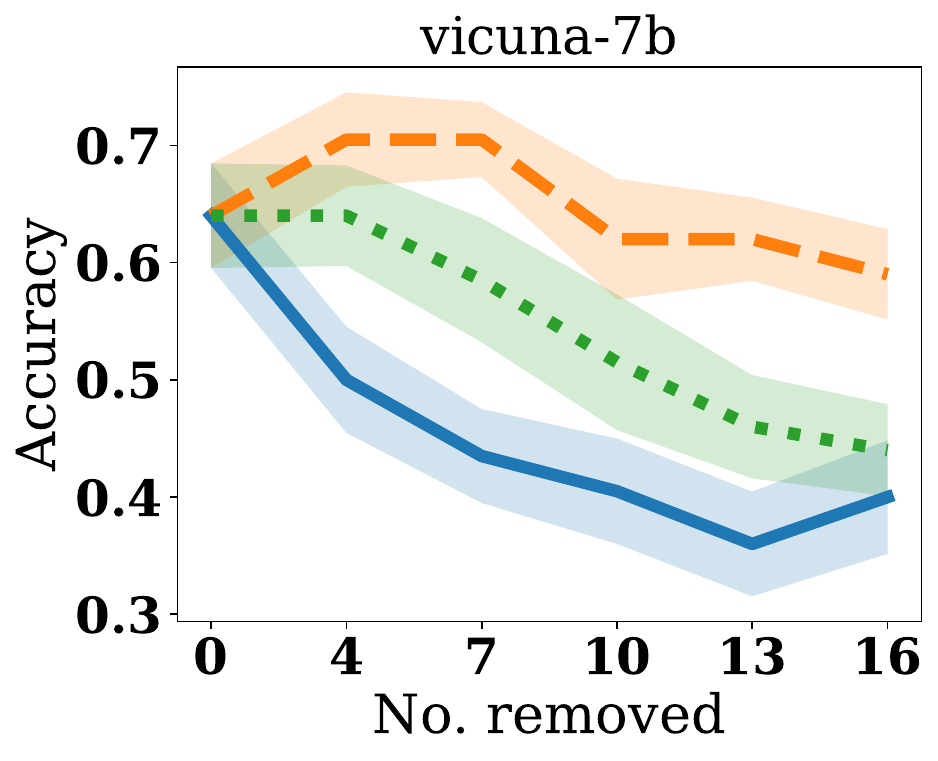}
    \end{subfigure}
\hfill
\begin{subfigure}[t]{0.23\textwidth}
    \includegraphics[width=\textwidth]{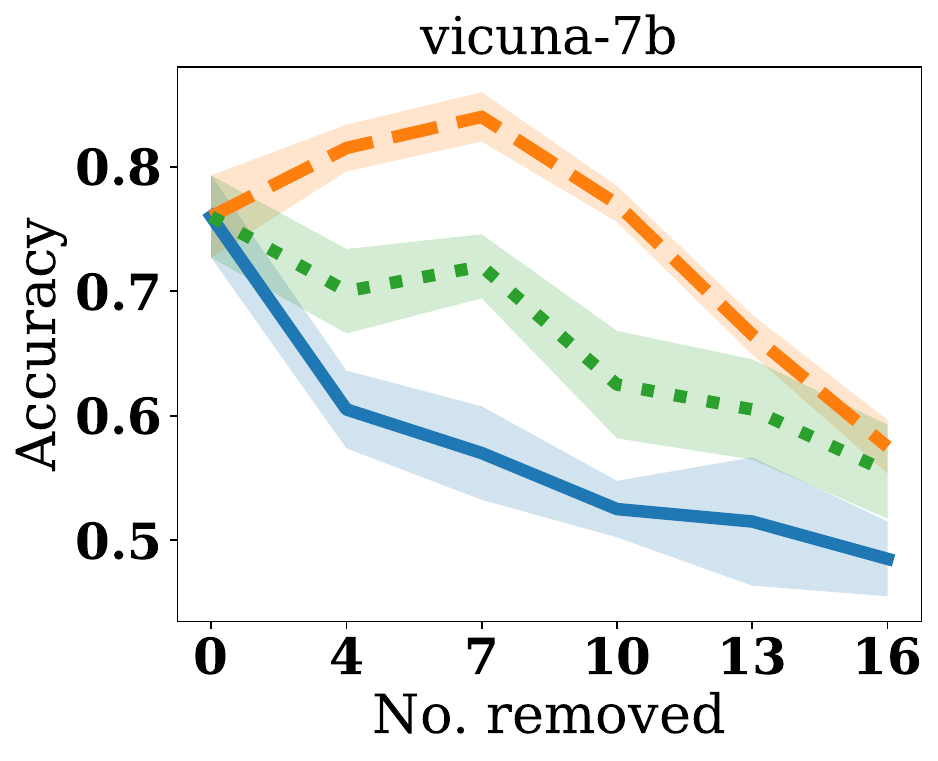}
    \end{subfigure}
\hfill
\caption{Corrupting and removing demonstration on datasets affects the model predictive power differently on AG News and SST-2, Rotten Tomatoes, and Subj from left to right using Vicuna-7b. Corrupting/removing demonstrations with high \texttt{DETAIL} scores results in lower model accuracy and \textit{vice versa}. Corrupting/removing demonstrations randomly results in an accuracy in the middle as expected. All experiments are repeated with $10$ independent trials. $\lambda=1.0$. Lines and shades represent the mean and standard error respectively.}
\label{fig:llm_label_perturbation_additional}
\end{figure}

\begin{figure}[H]
\centering
\begin{subfigure}[t]{0.23\textwidth}
    \includegraphics[width=\textwidth]{figs/cls_llm_ag_news_Llama-2-13b_acc_remove.pdf}
    \caption{AG News Removal}
    \end{subfigure}
\hfill
\begin{subfigure}[t]{0.23\textwidth}
    \includegraphics[width=\textwidth]{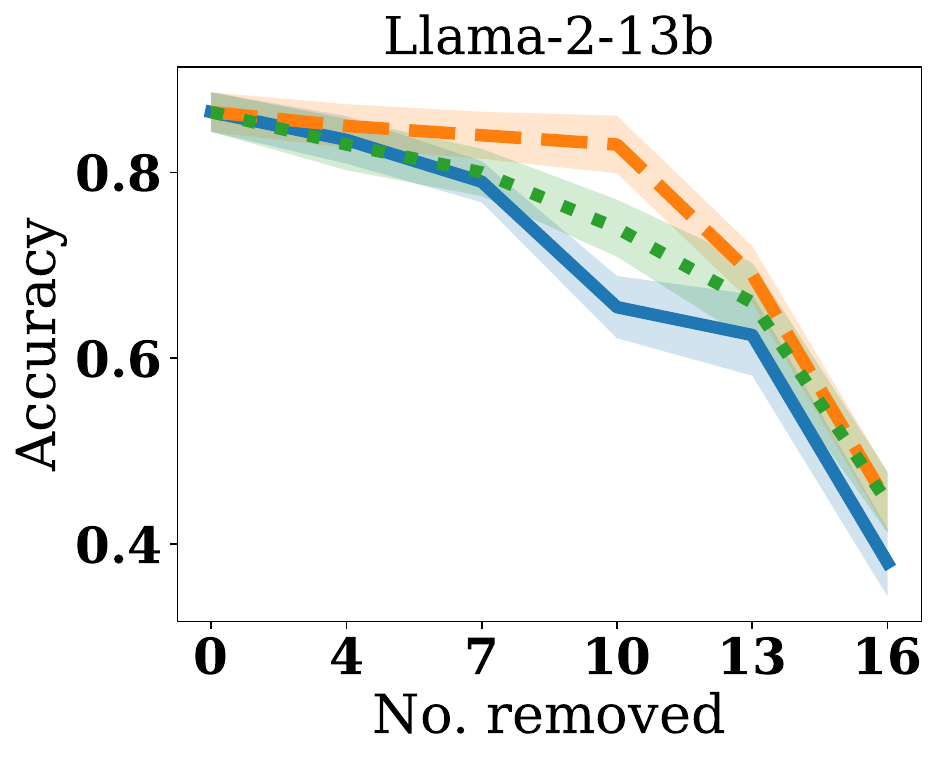}
    \caption{SST-2 Removal}
    \end{subfigure}
\hfill
\begin{subfigure}[t]{0.23\textwidth}
    \includegraphics[width=\textwidth]{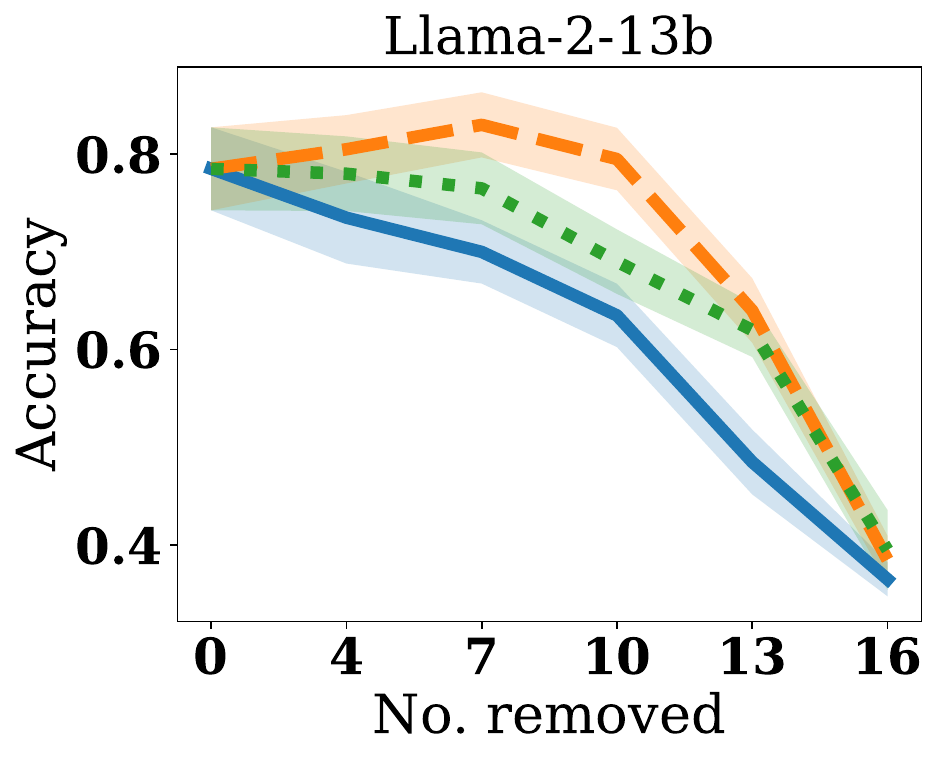}
    \caption{Rotten Tomatoes Removal}
    \end{subfigure}
\hfill
\begin{subfigure}[t]{0.23\textwidth}
    \includegraphics[width=\textwidth]{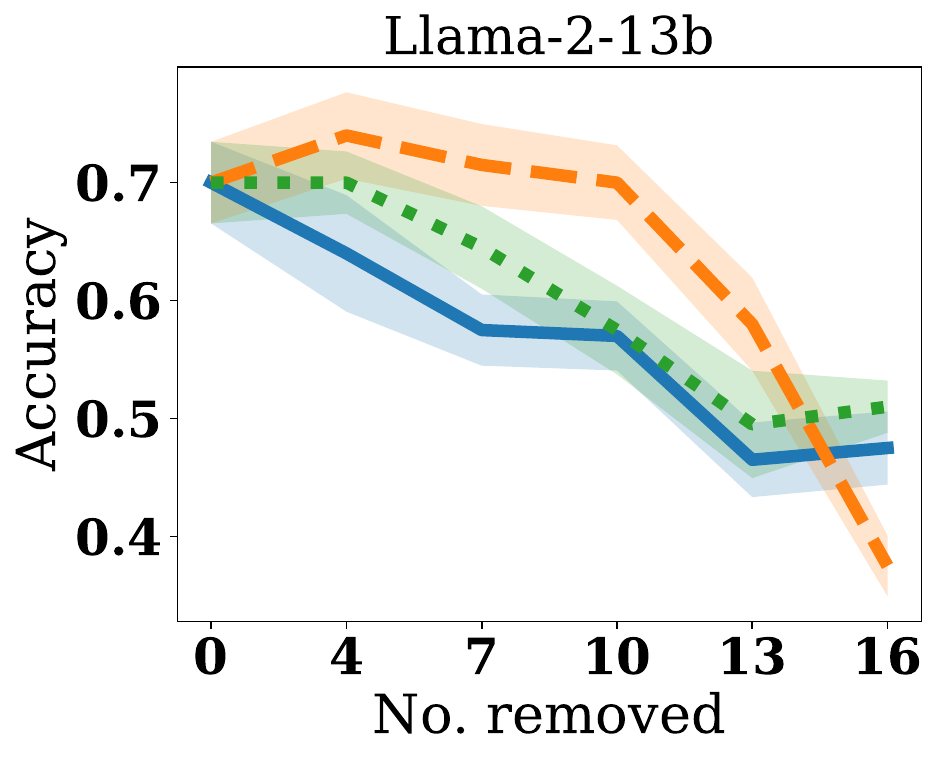}
    \caption{Subj Removal}
    \end{subfigure}
\hfill

\begin{subfigure}[t]{0.23\textwidth}
    \includegraphics[width=\textwidth]{figs/cls_llm_ag_news_Llama-2-13b_acc_corrupt.pdf}
    \caption{AG News Corrupt}
    \end{subfigure}
\hfill
\begin{subfigure}[t]{0.23\textwidth}
    \includegraphics[width=\textwidth]{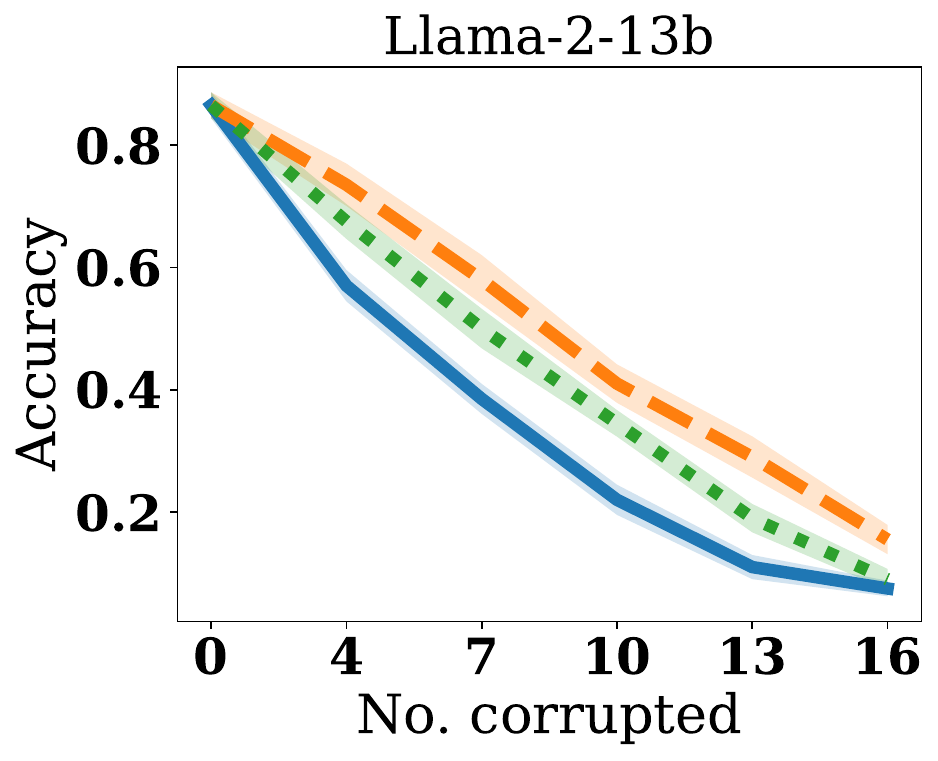}
    \caption{SST-2 Corrupt}
    \end{subfigure}
\hfill
\begin{subfigure}[t]{0.23\textwidth}
    \includegraphics[width=\textwidth]{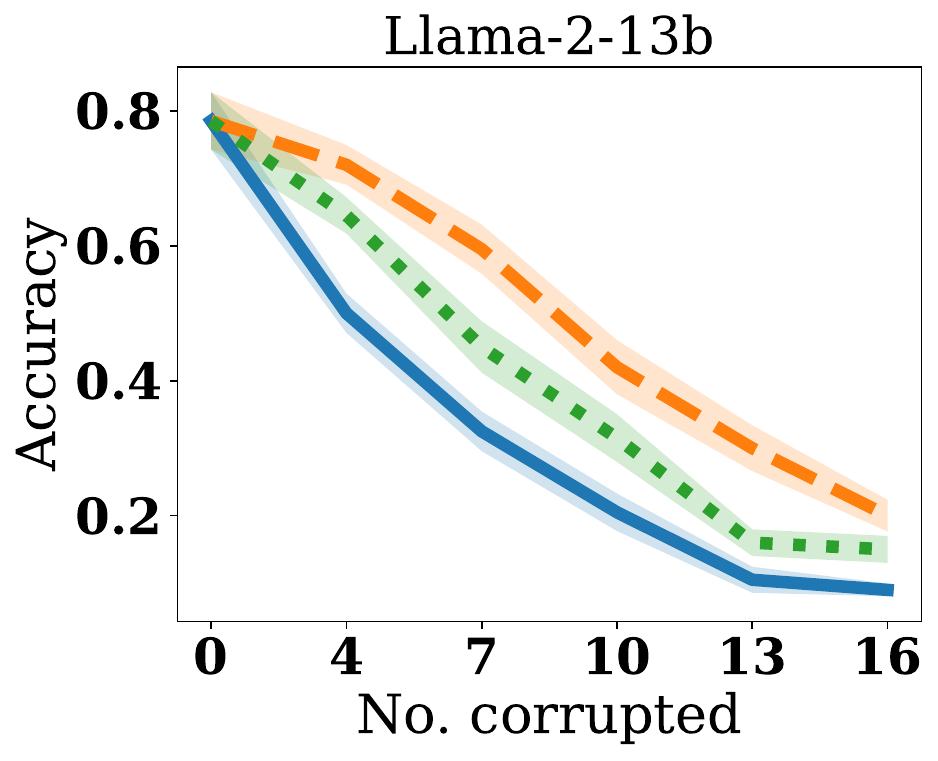}
    \caption{Rotten Tomatoes Corrupt}
    \end{subfigure}
\hfill
\begin{subfigure}[t]{0.23\textwidth}
    \includegraphics[width=\textwidth]{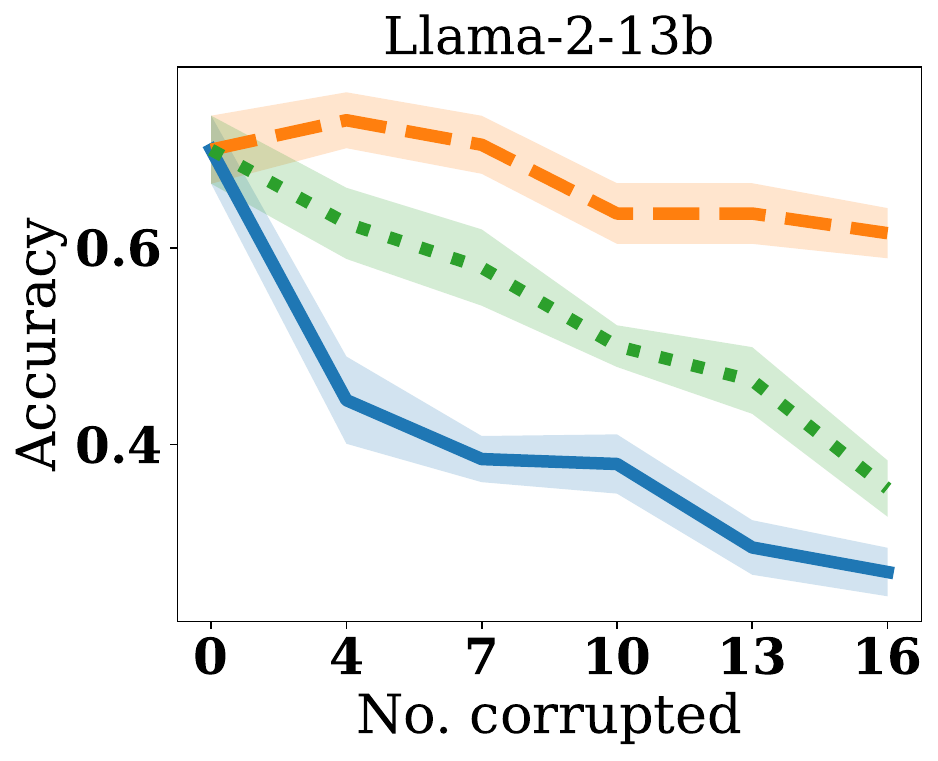}
    \caption{Subj Corrupt}
    \end{subfigure}
\hfill
\caption{Results of model prediction accuracy vs.~number of demonstrations removed/corrupted on Llama-2-13b model. $\lambda=1.0$. Lines and shades represent the mean and standard error respectively.}
\label{fig:llama13b-cls}
\end{figure}

We include the full results for the demonstration perturbation task using a Vicuna-7b v1.3 model in \cref{fig:llm_label_perturbation_additional} and using a Llama-2-13b model in \cref{fig:llama13b-cls}. A consistent trend can be observed across different datasets using both models.

We compare (in addition to Vicuna-7b) a total of $3$ LLMs: Llama-2-7b~\citep{touvron2023llama}, Llama-2-13b~\citep{touvron2023llama} and Falcon-7b~\citep{almazrouei2023falcon}. We reuse the same experimental setup as the demonstration label perturbation task and compare the accuracy by removing and corrupting $10$ among $20$ ICL data with high/low \texttt{DETAIL} scores computed in different models. The results are tabulated in \cref{tab:compare_models}. A similar trend can be observed across these models where removing/corrupting demonstrations with high $\cI_{\text{test}}$ results in lower accuracy and \textit{vice versa}. The results demonstrate that our method is robust against model pre-training/fine-tuning data as well as model size.

\begin{table}[H]
	\centering
    \setlength{\tabcolsep}{5pt}
    \caption{Performance on demonstration perturbation task across different models. The mean and standard error (in bracket) of predictive accuracy after removal or corruption $10$ out of $20$ demonstrations of $20$ randomly drawn ICL datasets is shown. All experiments are independently repeated $20$ times.}
    \resizebox{\linewidth}{!}{
    \footnotesize
    \begin{tabular}{@{}l|ccc|ccc@{}}
    \toprule
         & remove high $\downarrow$ & remove low $\uparrow$ & remove random & corrupt high $\downarrow$ & corrupt low $\uparrow$ & corrupt random \\ 
    \midrule
     \textbf{Subj}\\
        Llama-2-7b  & \textcolor{red!50!black}{0.380 (0.034)}  & \textcolor{green!50!black}{0.675 (0.025)}    & 0.535 (0.035) & \textcolor{red!50!black}{0.250 (0.036)}  & \textcolor{green!50!black}{0.690 (0.031)}    & 0.445 (0.019)  \\
        Llama-2-13b  & \textcolor{red!50!black}{0.570 (0.029)}  & \textcolor{green!50!black}{0.700 (0.032)}    & 0.575 (0.038) & \textcolor{red!50!black}{0.380 (0.030)}  & \textcolor{green!50!black}{0.635 (0.031)}    & 0.500 (0.021)  \\
        Falcon-7b   & \textcolor{red!50!black}{0.450 (0.026)}    & \textcolor{green!50!black}{0.690 (0.028)}    & 0.580 (0.042) & \textcolor{red!50!black}{0.280 (0.026)} & \textcolor{green!50!black}{0.630 (0.023)} & 0.445 (0.043)    \\
      \midrule
         \textbf{SST-2}\\
        Llama-2-7b  & \textcolor{red!50!black}{0.480 (0.031)}  & \textcolor{green!50!black}{0.670 (0.030)}    & 0.540 (0.042) & \textcolor{red!50!black}{0.145 (0.018)} & \textcolor{green!50!black}{0.445 (0.042)} & 0.275 (0.039)  \\
        Llama-2-13b  & \textcolor{red!50!black}{0.655 (0.034)}  & \textcolor{green!50!black}{0.830 (0.031)}    & 0.740 (0.031) & \textcolor{red!50!black}{0.220 (0.025)} & \textcolor{green!50!black}{0.410 (0.031)} & 0.345 (0.022)   \\
        Falcon-7b   & \textcolor{red!50!black}{0.560 (0.043)}     & \textcolor{green!50!black}{0.775 (0.028)}     & 0.680 (0.032) & \textcolor{red!50!black}{0.225 (0.031)} & \textcolor{green!50!black}{0.545 (0.030)} & 0.340 (0.023)    \\
      \midrule
       \textbf{Rotten toamtoes}\\
        Llama-2-7b  & \textcolor{red!50!black}{0.435 (0.034)}  & \textcolor{green!50!black}{0.670 (0.060)}   & 0.540 (0.045) & \textcolor{red!50!black}{0.120 (0.025)} & \textcolor{green!50!black}{0.420 (0.039)} & 0.235 (0.026)   \\
        Llama-2-13b  & \textcolor{red!50!black}{0.635 (0.032)}  & \textcolor{green!50!black}{0.795 (0.032)}    & 0.690 (0.033) & \textcolor{red!50!black}{0.205 (0.028)} & \textcolor{green!50!black}{0.420 (0.040)} & 0.315 (0.035)  \\
        Falcon-7b   & \textcolor{red!50!black}{0.475 (0.037)}     & \textcolor{green!50!black}{0.780 (0.024)}     & 0.620 (0.024) & \textcolor{red!50!black}{0.225 (0.023)} & \textcolor{green!50!black}{0.590 (0.031)} & 0.445 (0.025)    \\
      \midrule
       \textbf{AG News}\\
       Llama-2-7b   & \textcolor{red!50!black}{0.145 (0.018)}     & \textcolor{green!50!black}{0.525 (0.036)}     & 0.360 (0.026) & \textcolor{red!50!black}{0.150 (0.016)} & \textcolor{green!50!black}{0.520 (0.030)} & 0.325 (0.037)    \\
       Llama-2-13b   & \textcolor{red!50!black}{0.260 (0.018)}     & \textcolor{green!50!black}{0.600 (0.041)}     & 0.500 (0.032) & \textcolor{red!50!black}{0.175 (0.026)} & \textcolor{green!50!black}{0.565 (0.049)} & 0.385 (0.031)    \\
        Falcon-7b   & \textcolor{red!50!black}{0.155 (0.019)}     & \textcolor{green!50!black}{0.460 (0.021)}     & 0.335 (0.025) & \textcolor{red!50!black}{0.085 (0.020)} & \textcolor{green!50!black}{0.465 (0.017)} & 0.265 (0.027)   \\
      \bottomrule                          
    \end{tabular}
    \label{tab:compare_models}
    }
\end{table}

\subsection{Additional Results for Noisy Demonstration Detection} \label{app:exp_noisy}
\begin{figure}[H]
    \centering
    \includegraphics[width=0.23\linewidth]{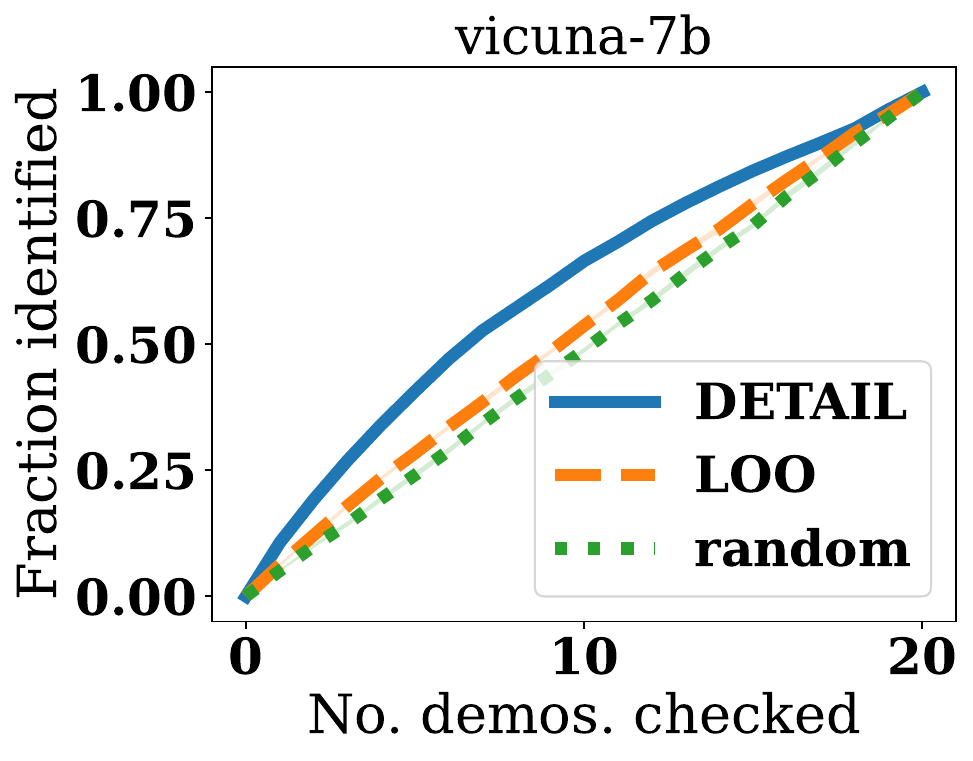}
    \hfill
    \includegraphics[width=0.23\linewidth]{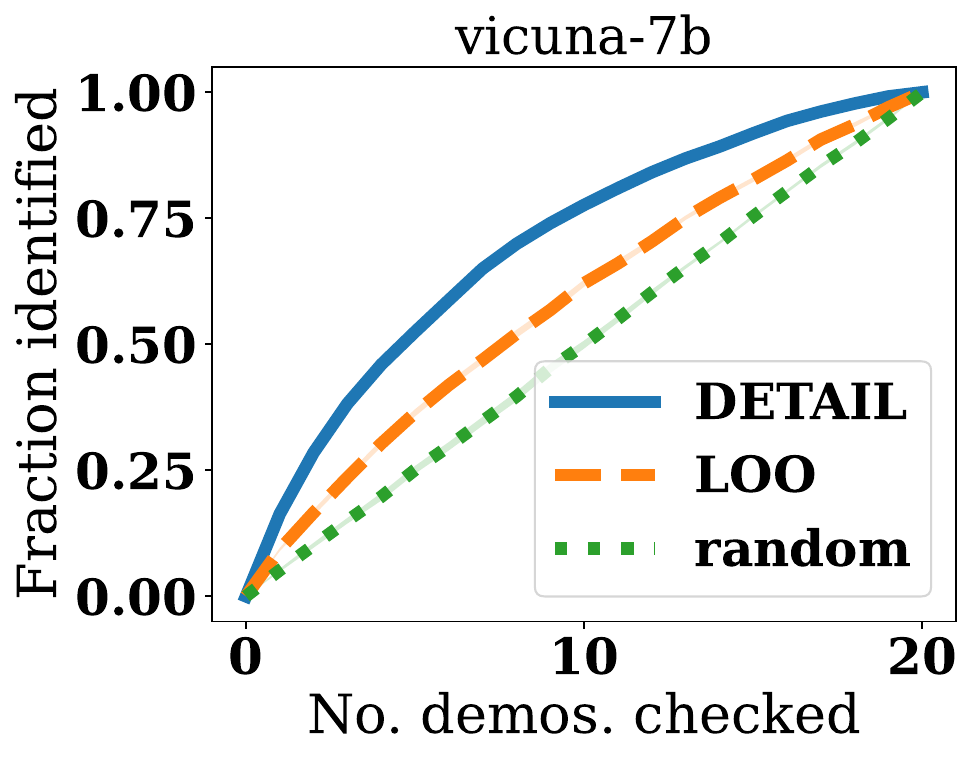}
    \hfill
    \includegraphics[width=0.23\linewidth]{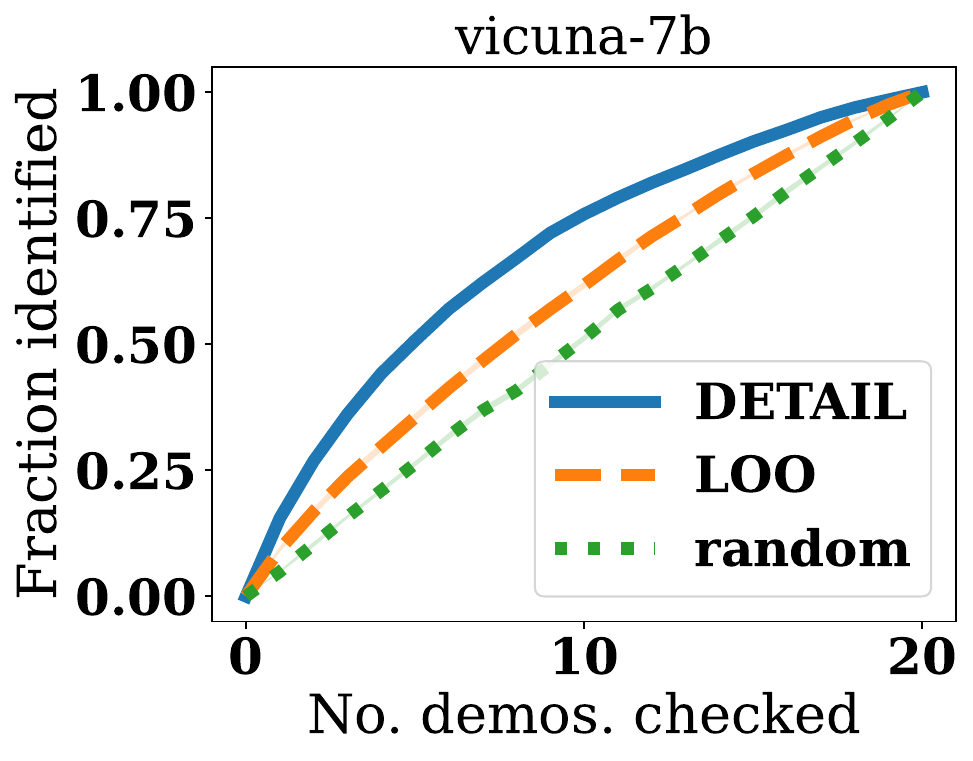}
    \hfill
    \includegraphics[width=0.23\linewidth]{figs/detect_llm_subj_vicuna-7b_fraction_checked.pdf}
    \hfill
    \caption{(Left to right) Fraction of all noisy labels identified vs.~the number of demonstrations ranked by our method (with projection down to $1000$ dimension) and LOO checked respectively on AG News, SST-2, Rotten Tomatoes, and Subj datasets. $\lambda=10^{-9}$. All experiments are repeated with $10$ independent trials. Lines and shades represent the mean and standard error respectively.}
    \label{fig:noisy_label_detection_additional}
\end{figure}

\begin{figure}[H]
\centering
\begin{subfigure}[t]{0.19\textwidth}
    \includegraphics[width=\textwidth]{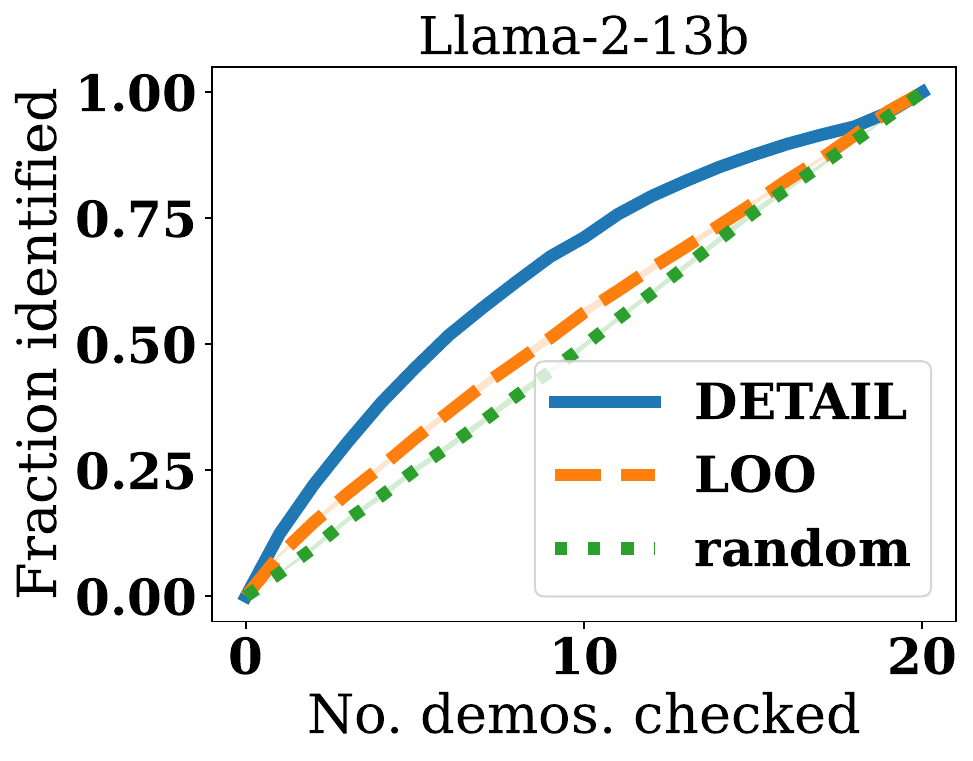}
    \caption{Detecting noisy label on AG News}
    \end{subfigure}
\hfill
\begin{subfigure}[t]{0.19\textwidth}
    \includegraphics[width=\textwidth]{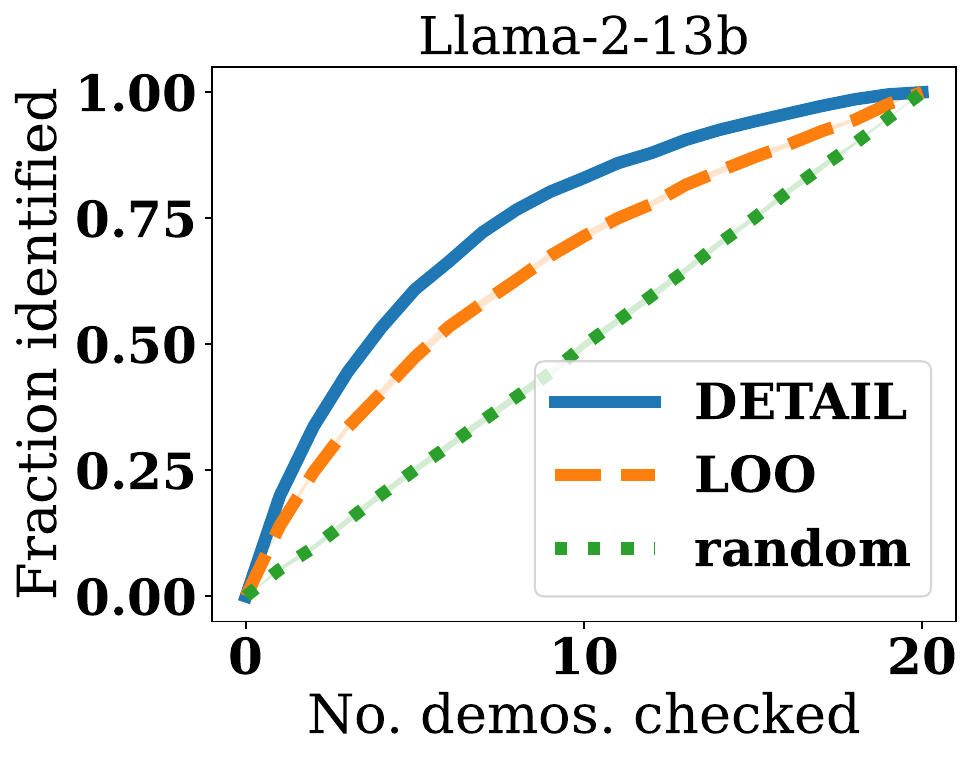}
    \caption{Detecting noisy label on SST-2}
    \end{subfigure}
\hfill
\begin{subfigure}[t]{0.19\textwidth}
    \includegraphics[width=\textwidth]{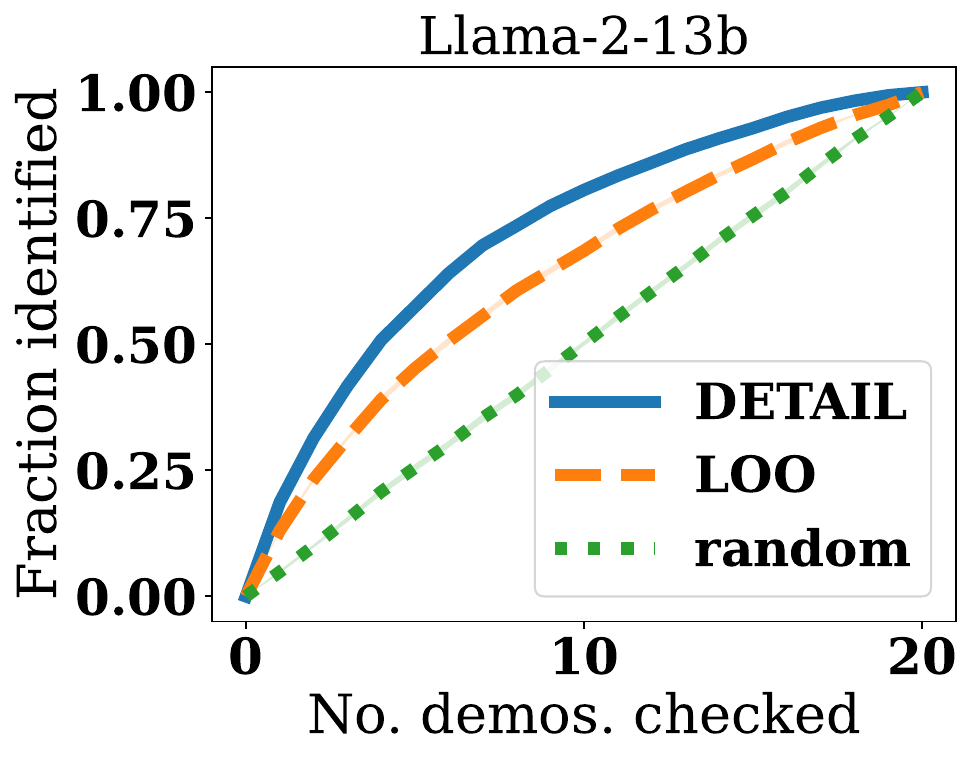}
    \caption{Detecting noisy label on Rotten Tomatoes}
    \end{subfigure}
\hfill
\begin{subfigure}[t]{0.19\textwidth}
    \includegraphics[width=\textwidth]{figs/detect_llm_subj_Llama-2-13b_fraction_checked.pdf}
    \caption{Detecting noisy label on Subj}
    \end{subfigure}
\hfill
\begin{subfigure}[t]{0.19\textwidth}
    \includegraphics[width=\textwidth]{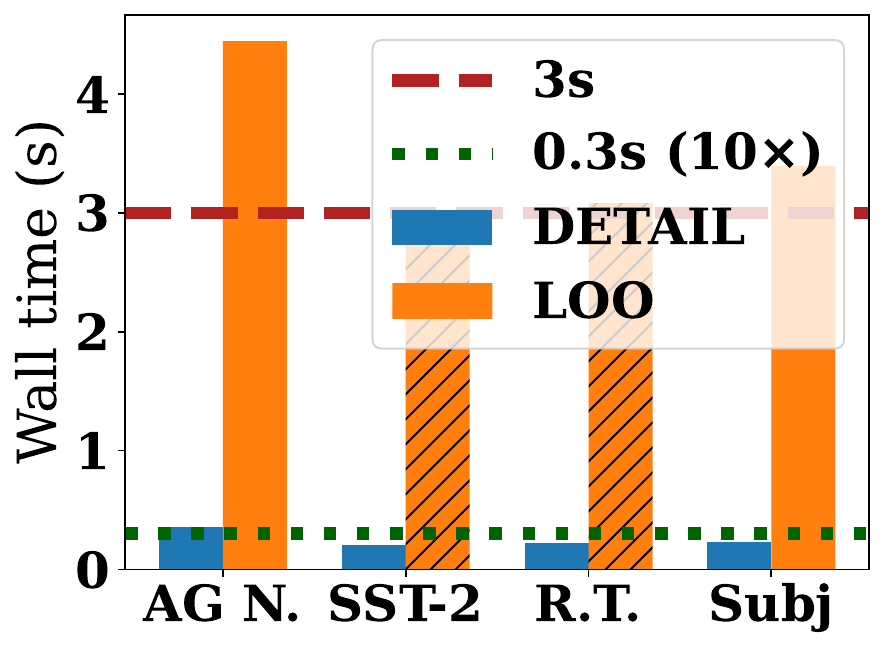}
    \caption{Wall time comparison}
    \end{subfigure}
\hfill
\caption{(a-d) Fraction of all noisy labels identified vs.~the number of demonstrations ranked by our method (with projection down to $1000$ dimension) and LOO checked respectively. (e) Wall time comparison across all datasets. $\lambda=10^{-9}$.  All experiments are repeated with $10$ independent trials using a Llama-2-13b model. Lines and shades represent the mean and standard error respectively.}
\label{fig:llm_noisy_label_detection_llama_13b}
\end{figure}

We include the results on AG News, SST-2, Rotten Tomatoes, and Subj datasets using a Vicuna-7b model in \cref{fig:noisy_label_detection_additional}. Similar trends as in the main text is observed. A counterpart experiment using Llama-2-13b is in \cref{fig:llm_noisy_label_detection_llama_13b}, where a similar trend is observed.

\subsection{Additional Results for Dimension Reduction}
\label{app:exp_dim_reduction_additional}

\begin{figure}[H]
\centering
\begin{subfigure}[t]{0.23\textwidth}
    \includegraphics[width=\textwidth]{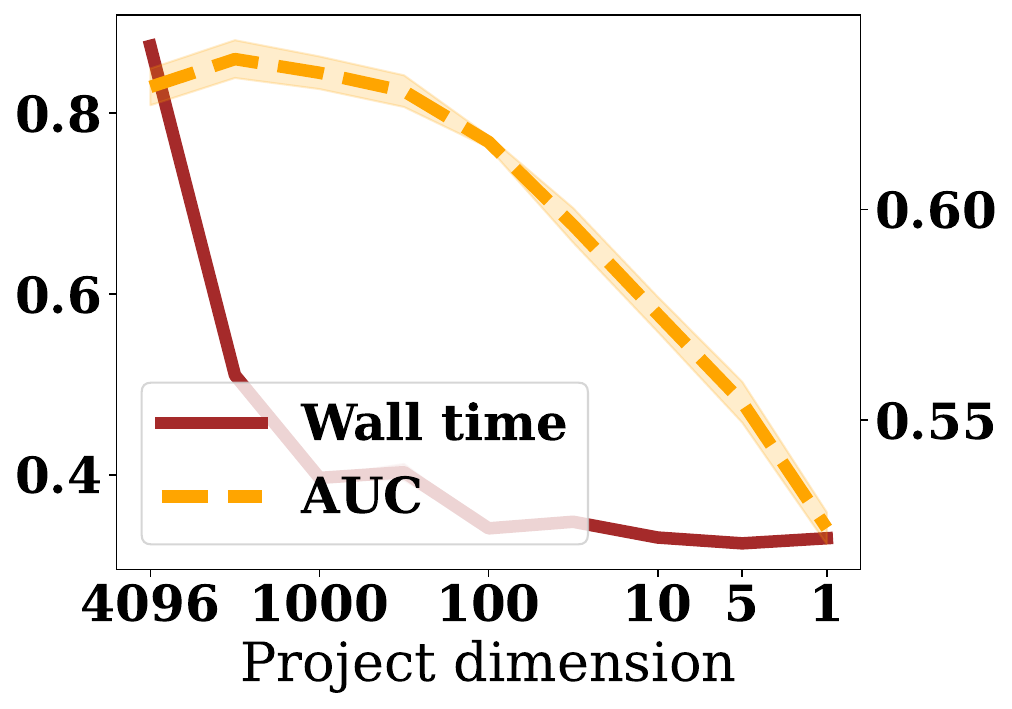}
    \end{subfigure}
\hfill
\begin{subfigure}[t]{0.23\textwidth}
    \includegraphics[width=\textwidth]{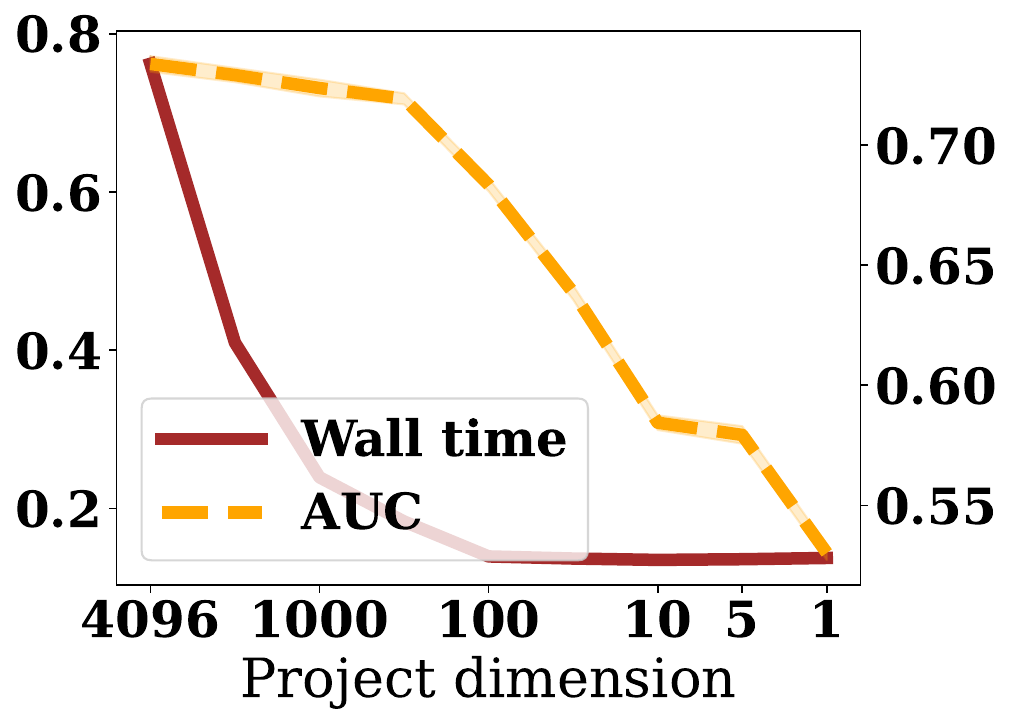}
    \end{subfigure}
\hfill
\begin{subfigure}[t]{0.23\textwidth}
    \includegraphics[width=\textwidth]{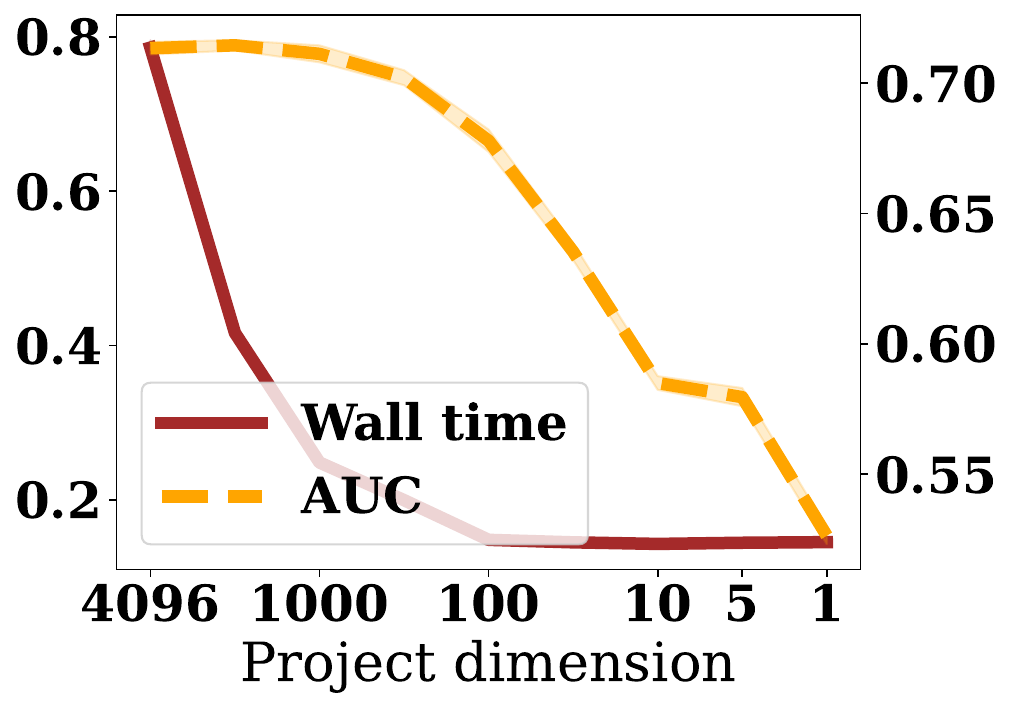}
    \end{subfigure}
\hfill
\begin{subfigure}[t]{0.23\textwidth}
    \includegraphics[width=\textwidth]{figs/detect_llm_proj_subj_vicuna-7b_fraction_checked_proj.pdf}
    \end{subfigure}
\hfill
\caption{(Left to right) wall time in seconds (left $y$-axis) and  AUCROC (right $y$-axis) vs.~projection dimension $d'$ on AG news, SST-2, Rotten Tomatoes, and Subj datasets. Experiments are repeated with $10$ trials. Lines and shades represent the mean and standard error respectively.}
\label{fig:llm_proj_noisy_label_detection}
\end{figure}

The experiments using Vicuna-7b on AG News, SST-2, Rotten Tomatoes, and Subj can be found in \cref{fig:llm_proj_noisy_label_detection}. It can be observed that the trend is consistent across different datasets.

\subsection{Additional Results for Demonstration Reordering Task}
\label{app:demo_reorder}
We conduct additional experiments on the demonstration reordering task by perturbing $6$ demonstrations in each ICL dataset of $20$ demonstrations. The results are shown in \cref{tab:position_additional}. It can be observed that reordering with $\cI_{\text{self}}$ still achieves an improvement in test accuracy, demonstrating the robustness of our method.

\begin{table}[H]
    \centering
    \captionof{table}{Predictive accuracy of demonstrations permuted randomly and based on $\cI_{\text{self}}$ respectively. The mean and standard error (in bracket) with $80$ repeated trials is shown.}
    \resizebox{0.8\linewidth}{!}{
    \footnotesize
    \begin{tabular}{@{}lccc@{}}
    \toprule
         & Subj & SST-2 & Rotten Tomatoes \\ 
    \midrule
     \textbf{Corrupt $6$ demonstrations}\\
        Baseline (random)  & \textcolor{red!50!black}{0.588 (7.96e-03)}  & \textcolor{red!50!black}{0.487 (9.45e-03)} &  \textcolor{red!50!black}{0.398 (1.12e-02)} \\
        Reorder (\texttt{DETAIL}) & \textcolor{green!50!black}{0.604 (7.39e-03)} &  \textcolor{green!50!black}{0.520 (1.01e-02)} & \textcolor{green!50!black}{0.425 (1.37-02)}  \\
        Difference $\uparrow$ & \textbf{0.0164 (7.05e-03)} & \textbf{0.0323 (8.13e-03)} & \textbf{0.0267 (1.01e-02)} \\
      \bottomrule                          
    \end{tabular}
    }
    \label{tab:position_additional}
\end{table}

\subsection{Additional Results for Demonstration Curation Task}
\label{app:other_attr_additional}

\begin{figure}[H]
\centering
\begin{subfigure}[t]{0.23\textwidth}
    \includegraphics[width=\textwidth]{figs/cls_llm_comp_ag_news_vicuna-7b_acc_remove.pdf}
    \end{subfigure}
\hfill
\begin{subfigure}[t]{0.23\textwidth}
    \includegraphics[width=\textwidth]{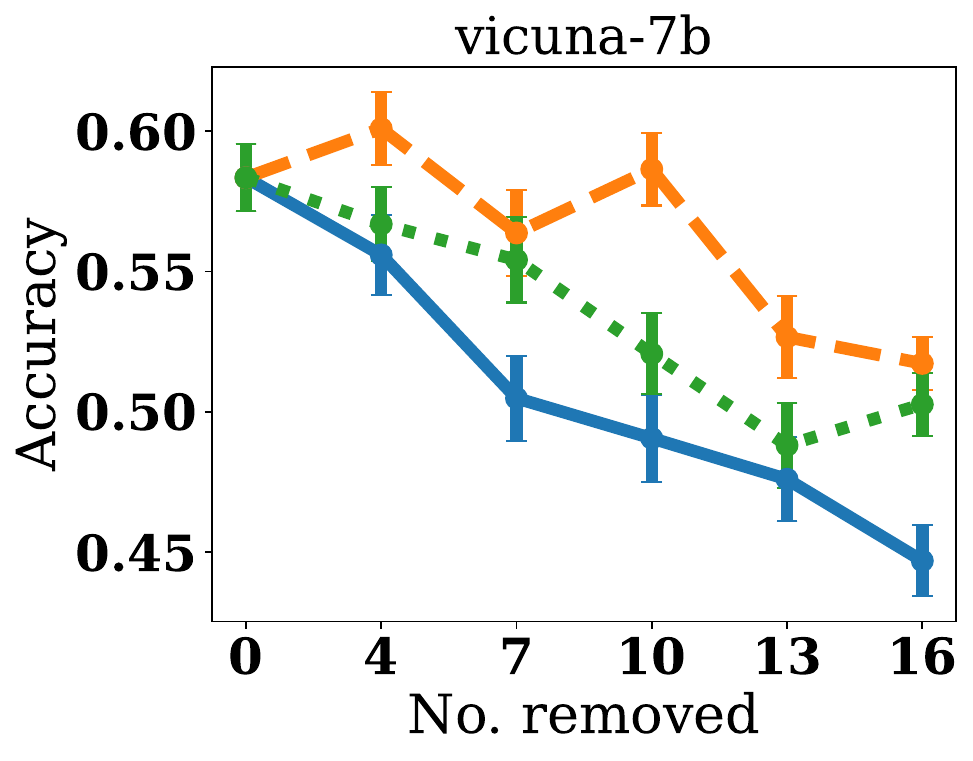}
    \end{subfigure}
\hfill
\begin{subfigure}[t]{0.23\textwidth}
    \includegraphics[width=\textwidth]{figs/cls_llm_comp_rotten_tomatoes_vicuna-7b_acc_remove.pdf}
    \end{subfigure}
\hfill
\begin{subfigure}[t]{0.23\textwidth}
    \includegraphics[width=\textwidth]{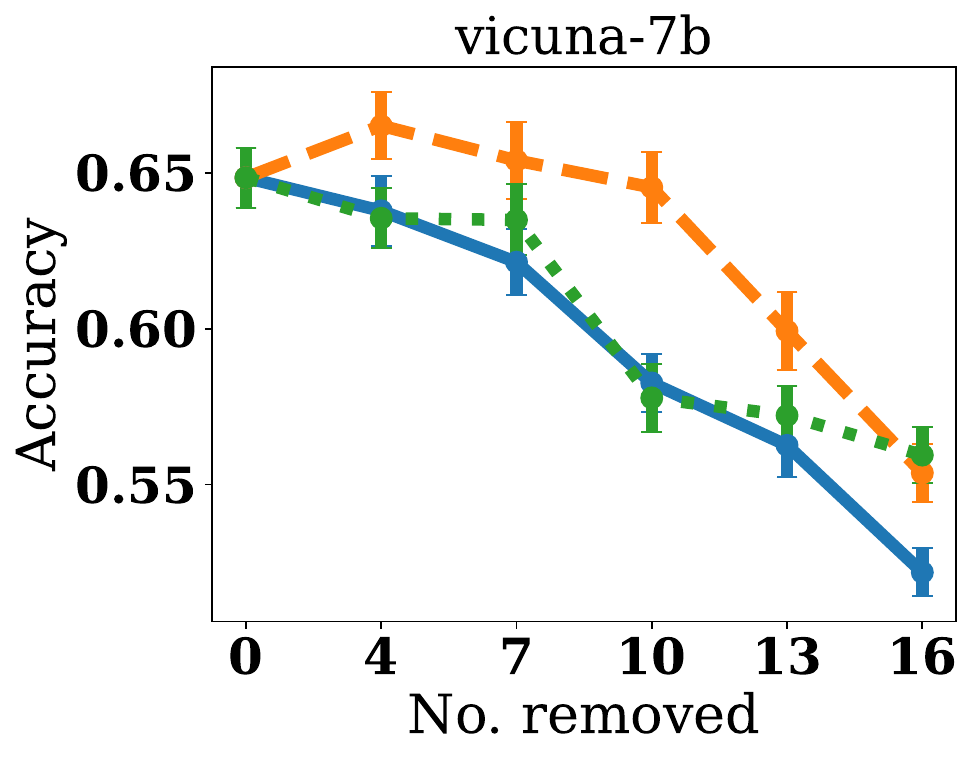}
    \end{subfigure}
\hfill
\begin{subfigure}[t]{0.23\textwidth}
    \includegraphics[width=\textwidth]{figs/cls_llm_comp_ag_news_Llama-2-13b_acc_remove.pdf}
    \end{subfigure}
\hfill
\begin{subfigure}[t]{0.23\textwidth}
    \includegraphics[width=\textwidth]{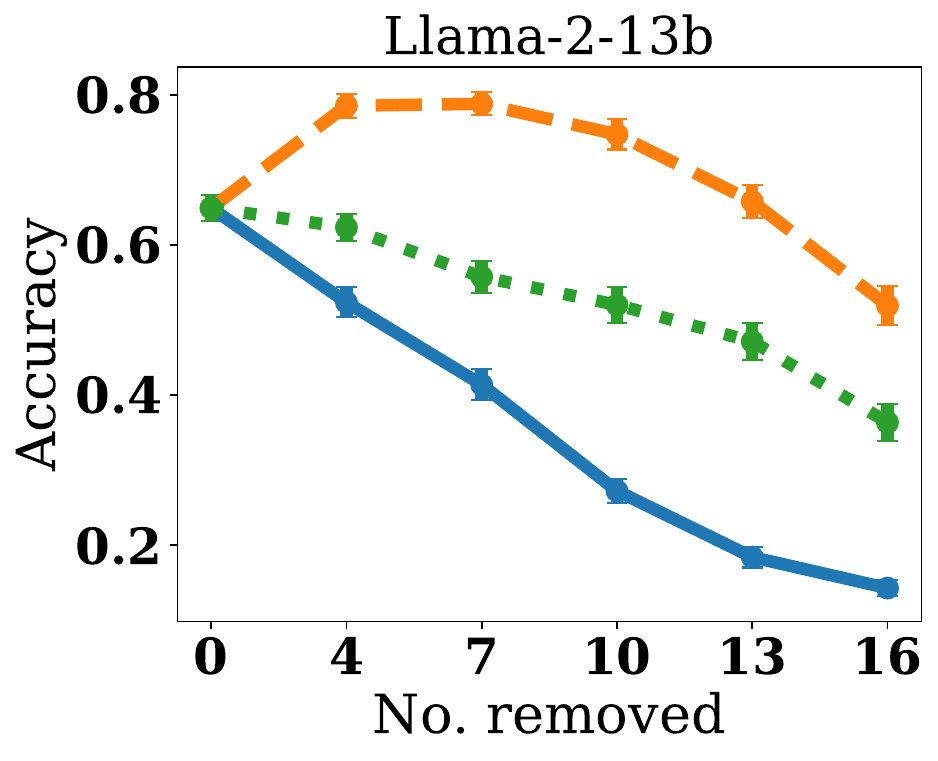}
    \end{subfigure}
\hfill
\begin{subfigure}[t]{0.23\textwidth}
    \includegraphics[width=\textwidth]{figs/cls_llm_comp_rotten_tomatoes_Llama-2-13b_acc_remove.pdf}
    \end{subfigure}
\hfill
\begin{subfigure}[t]{0.23\textwidth}
    \includegraphics[width=\textwidth]{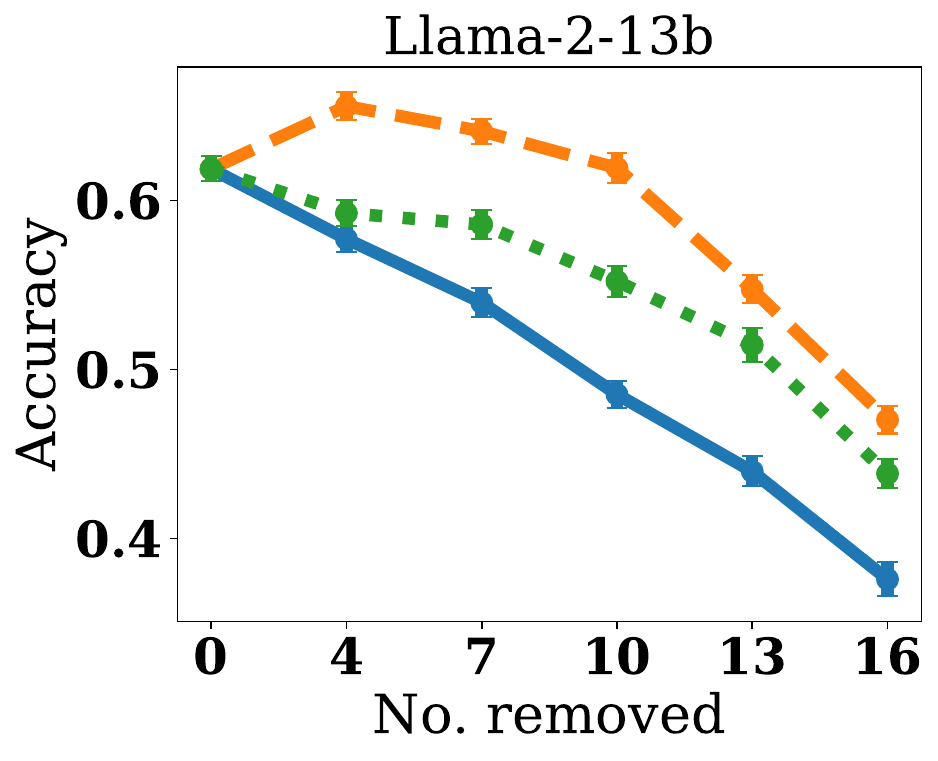}
    \end{subfigure}
\hfill
\caption{(Left to right) test accuracy vs.~number of demonstrations removed using $\cI_{\text{test}}$ on AG news, SST-2, Rotten Tomatoes, and Subj datasets using (top) Vicuna-7b and (bottom) Llama-2-13b. All experiments are repeated with $80$ independent trials. Lines and bars represent the mean and standard error respectively.}
\label{fig:icl_data_curation_all}
\end{figure}

We include the full results for all datasets on both Vicuna-7b and Llama-2-13b in \cref{fig:icl_data_curation_all}. It can be observed that the gap between removing demonstrations with high/low $\cI_{\text{test}}$ is wider with Llama-2-13b. We believe this is because Llama-2-13b being a larger model possesses better capability of formulating the internal optimizer as compared to Vicuna-7b which is smaller.

We additionally provide the result for the demonstration curation task with Attention~\citep{bahdanau2016neural} and \citep{s2024incontext} using $10$ iterations of LiSSA update for computing the hessian vector product~\citep{agarwal17lissa} and compare it with the case with $5$ iterations. The result is shown in \cref{tab:other_attr_additional}. With $10$ iterations, the wall time is much higher (over $20 \times$) but accuracy is comparable to the case with $5$ iterations. For Attention, the performance is comparable to random removal as shown in \cref{tab:other_attr}.

\begin{table}[H]
    \centering
    \setlength{\tabcolsep}{5pt}
    \caption{Test accuracy after curating the ICL dataset and the incurred wall time (in seconds on one L40 GPU). The mean and std. error (in bracket) is shown with $20$ repeated trials.}
    \resizebox{\linewidth}{!}{
    \footnotesize
    \begin{tabular}{@{}lccc@{}}
    \toprule
         Metric & Attention~\citep{bahdanau2016neural} & \citet{s2024incontext} (\#5) & \citet{s2024incontext} (\#10)  \\ 
    \midrule
     \textbf{Subj (curate $10$ demonstrations)}\\
        Accuracy $\uparrow$ & 0.627 (2.01e-02) & 0.556 (1.38e-02) & 0.597 (3.23e-02)  \\
        Wall time $\downarrow$ & 37.6 (1.09e+00) &  9.37 (4.19e-01) & 245 (8.80e-01)  \\
      \midrule
      \textbf{SST-2 (curate $10$ demonstrations)} \\
        Accuracy $\uparrow$ & 0.460 (2.60e-02) & 0.493 (1.34e-02) & 0.499 (1.64e-02) \\
        Wall time $\downarrow$ & 24.9 (7.26e-01) &  10.6 (7.80e-01) & 241 (9.63e-01)  \\
      \midrule
      \textbf{Rotten Tomatoes (curate $10$ demonstrations)} \\
        Accuracy $\uparrow$ & 0.488 (2.05e-02) & 0.498 (1.72e-02) & 0.494 (1.74e-02)  \\
        Wall time $\downarrow$ & 36.8 (1.09e+00) & 9.74 (5.57e-01) & 240 (4.61e-01)  \\
      \midrule
      \textbf{AG News (curate $10$ demonstrations)} \\
        Accuracy $\uparrow$ & 0.350 (1.35e-02) & 0.416 (2.10e-02) & 0.346 (2.07e-02) \\
        Wall time $\downarrow$ & 138 (3.28e+00) & 11.9 (5.78e-01) & 441 (4.0e-01)  \\
      \bottomrule                          
    \end{tabular}
    \label{tab:other_attr_additional}
    }
\end{table}

Given that BERT-score is light-weight, it is intuitive to think about whether it is possible to combine \texttt{DETAIL} and BERT score and reap the benefits of both worlds. We conduct a preliminary experiment in this direction. Specifically, we consider applying our formulation of \texttt{DETAIL} (\cref{eq:influence}) using a hidden state modified by the BERT embedding. Specifically, we consider a weighted average of the BERT embedding and the transformer's hidden state as well as a direct concatenation of the two embeddings. The results are shown in \cref{tab:other_attr_additional_bert}. Surprisingly, we discover that using a weighted average of the transformer's hidden state and the BERT embedding improves the attribution accuracy.

\begin{table}[H]
    \centering
    \setlength{\tabcolsep}{5pt}
    \caption{Test accuracy after curating the ICL dataset and the incurred wall time (in seconds on one L40 GPU). The mean and std. error (in bracket) is shown with $20$ repeated trials.}
    \resizebox{\linewidth}{!}{
    \footnotesize
    \begin{tabular}{@{}lccc@{}}
    \toprule
         Metric & BERT Embedding Only & Weighted Average (equal weight) & Concatenation  \\ 
    \midrule
     \textbf{Subj (curate $10$ demonstrations)}\\
        Accuracy $\uparrow$ & 0.665 (2.36e-02) & \textbf{0.758 (1.97e-02)} & 0.719 (2.75e-02)  \\
      \midrule
      \textbf{SST-2 (curate $10$ demonstrations)} \\
        Accuracy $\uparrow$ & 0.475 (1.31e-02) & 0.579 (2.34e-02) & \textbf{0.596 (1.96e-02)} \\
      \midrule
      \textbf{Rotten Tomatoes (curate $10$ demonstrations)} \\
        Accuracy $\uparrow$ & 0.510 (2.10e-02) & \textbf{0.575 (2.46e-02)} & 0.570 (2.75e-02)  \\
      \midrule
      \textbf{AG News (curate $10$ demonstrations)} \\
        Accuracy $\uparrow$ & 0.408 (2.10e-02) & \textbf{0.425 (1.76e-02)} & 0.412 (1.50e-02) \\
      \bottomrule                          
    \end{tabular}
    \label{tab:other_attr_additional_bert}
    }
\end{table}

\subsection{Ablation of Different Transformer Layers for Computing \texttt{DETAIL} scores.}\label{app:exp_layer}

\begin{figure}[H]
\centering
\begin{subfigure}[t]{0.3\textwidth}
    \includegraphics[width=\textwidth]{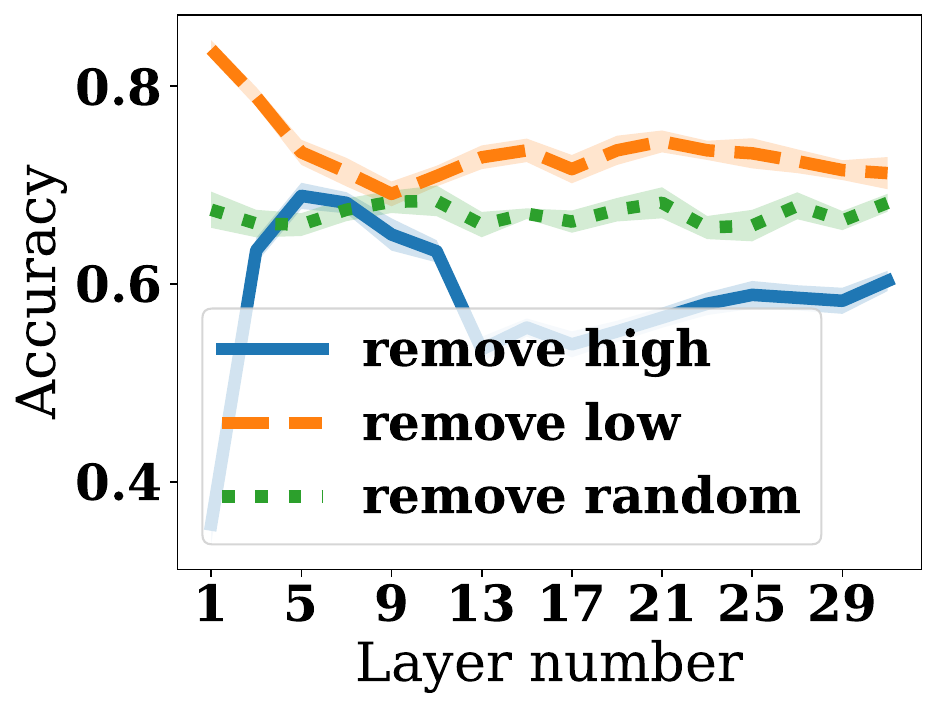}
    \caption{Remove demonstration on Subj}
    \end{subfigure}
\hfill
\begin{subfigure}[t]{0.3\textwidth}
    \includegraphics[width=\textwidth]{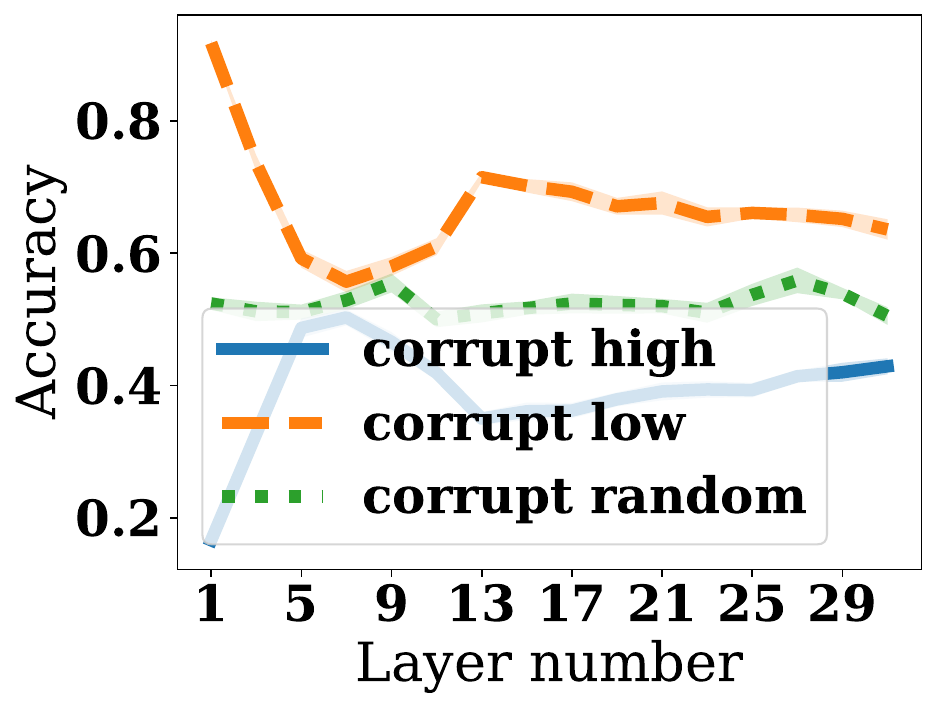}
    \caption{Corrupt demonstration on Subj}
    \end{subfigure}
\hfill
\begin{subfigure}[t]{0.3\textwidth}
    \includegraphics[width=\textwidth]{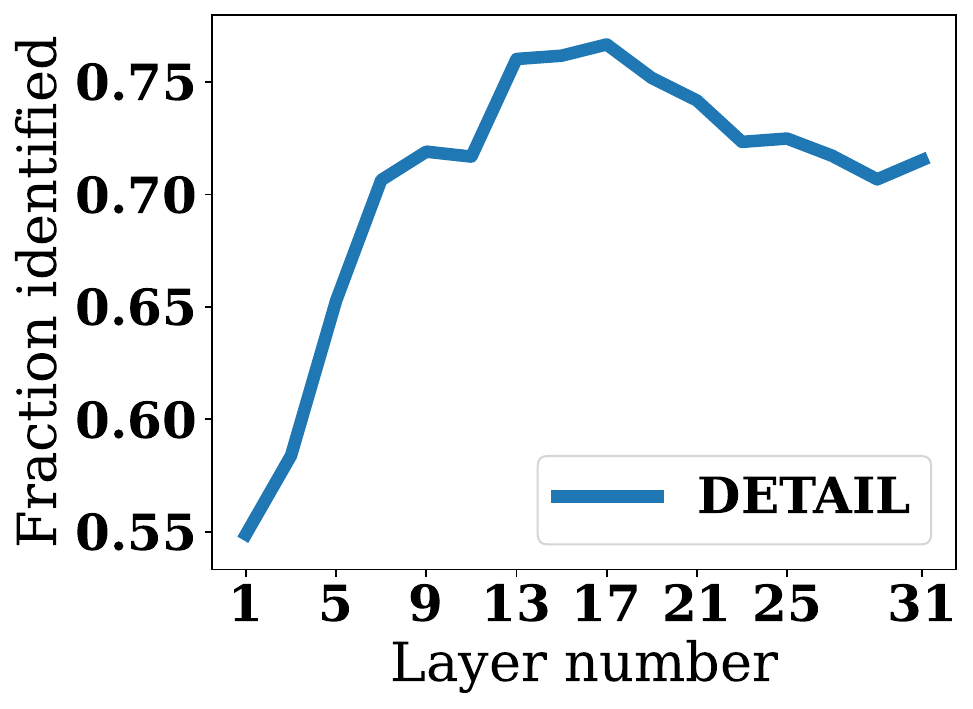}
    \caption{Detecting noisy label on Subj}
    \end{subfigure}
\hfill
\begin{subfigure}[t]{0.3\textwidth}
    \includegraphics[width=\textwidth]{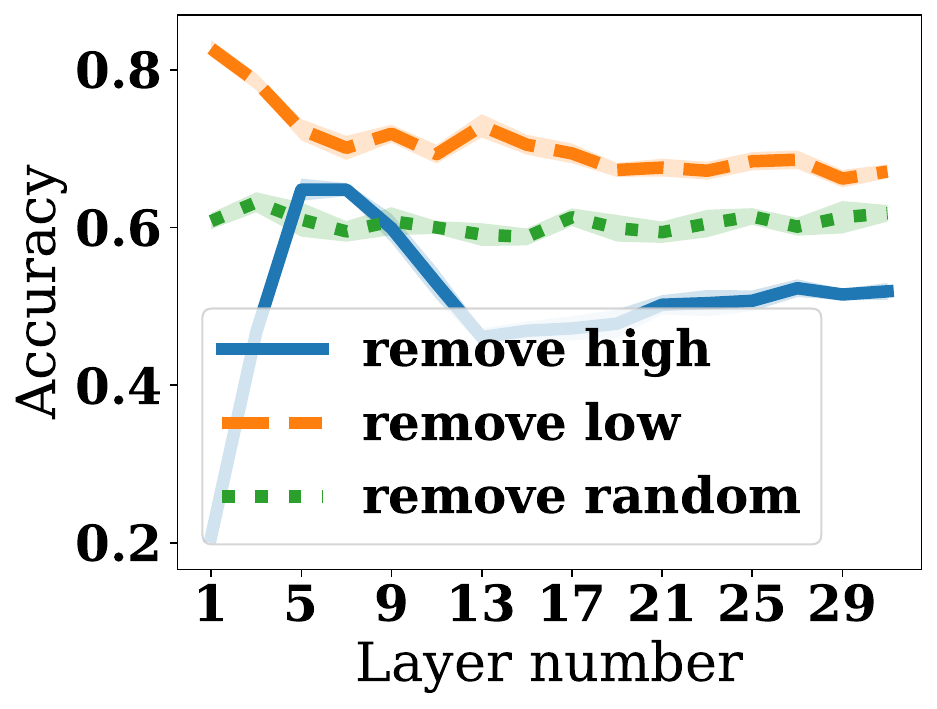}
    \caption{Remove demonstration on SST-2}
    \end{subfigure}
\hfill
\begin{subfigure}[t]{0.3\textwidth}
    \includegraphics[width=\textwidth]{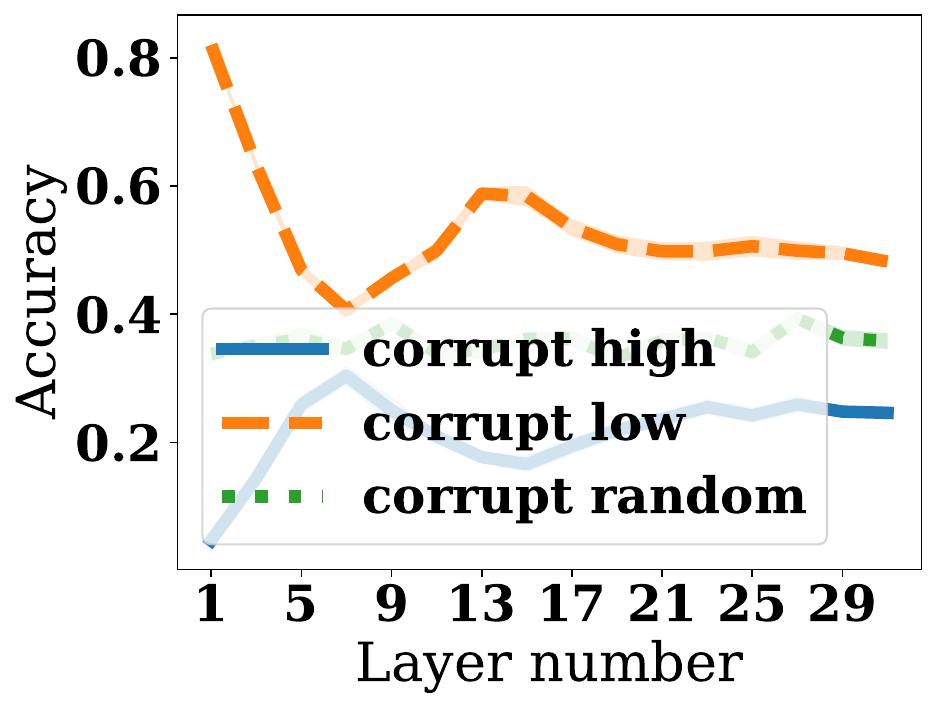}
    \caption{Corrupt demonstration on SST-2}
    \end{subfigure}
\hfill
\begin{subfigure}[t]{0.3\textwidth}
    \includegraphics[width=\textwidth]{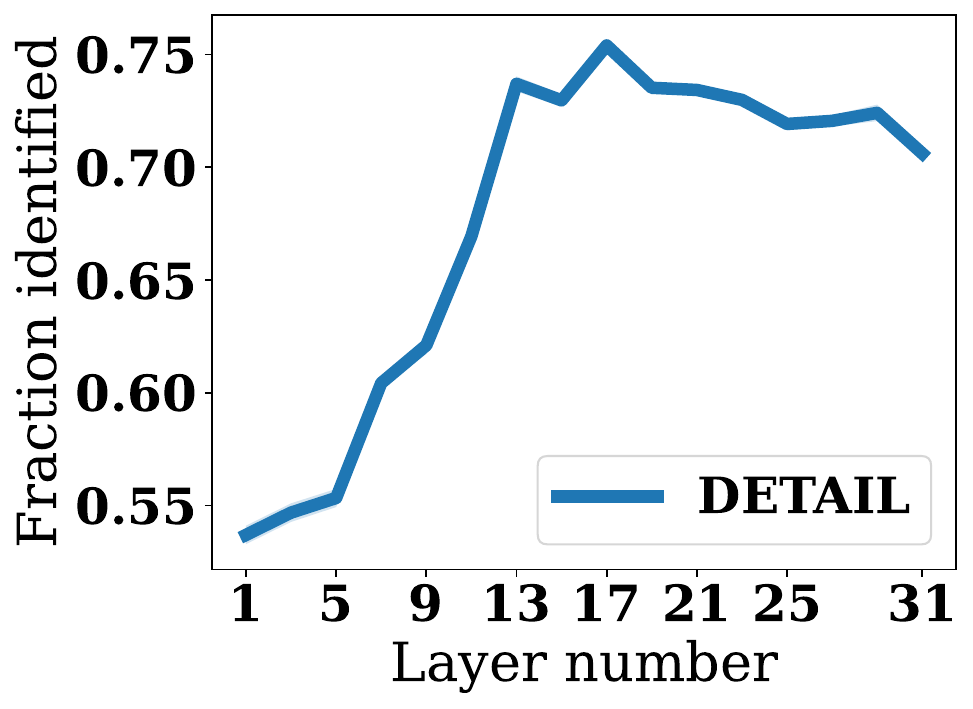}
    \caption{Detecting noisy label on SST-2}
    \end{subfigure}
\hfill
\caption{Results of different task performance vs.~the layer number in a Vicuna-7b model which consists of $31$ layers. Experiments are repeated with $10$ trials. $\lambda = 1.0$ for (a,b,d,e) and $\lambda=10^{-9}$ for (c,f). Lines and shades represent the mean and standard error respectively.}
\label{fig:llm_layers}
\end{figure}

We experiment with the difference in the effectiveness \texttt{DETAIL} using the embeddings of different layers. We conduct experiments on demonstration removal, demonstration perturbation, and noisy label detection tasks. The results are shown in \cref{fig:llm_layers}. It can be observed that obtaining the \texttt{DETAIL} scores from the later layers of the model consistently produces desirable results.

\subsection{Ablation of Target Position for Computing \texttt{DETAIL}.}\label{app:exp_target_pos}

\begin{figure}[H]
\centering
\begin{subfigure}[t]{0.23\textwidth}
    \includegraphics[width=\textwidth]{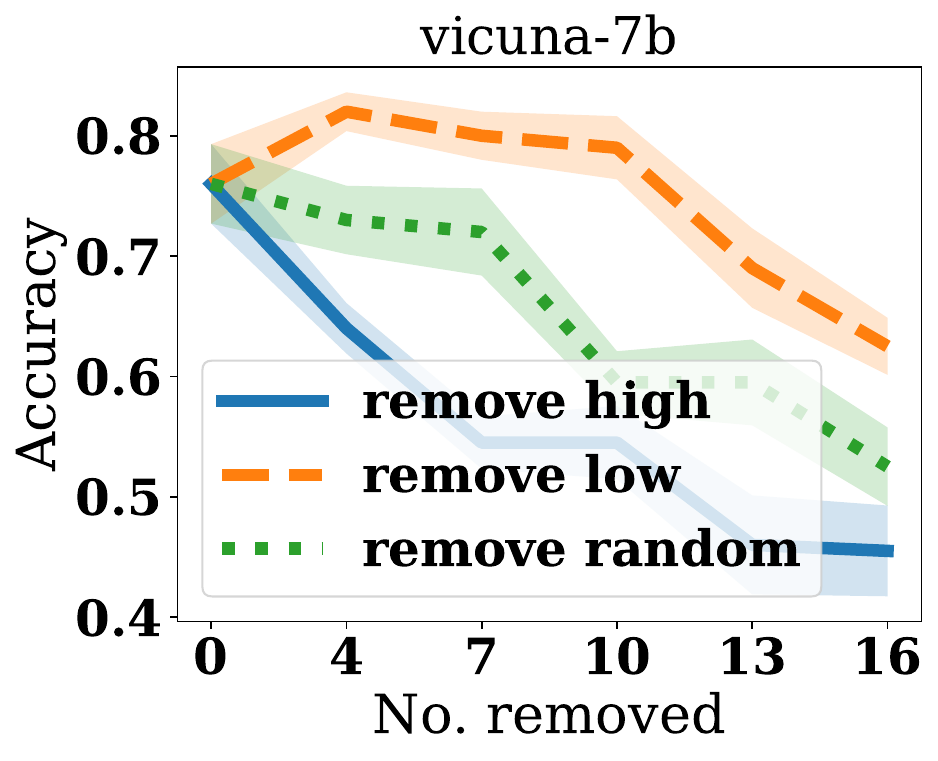}
    \caption{Subj using label position}
    \end{subfigure}
\hfill
\begin{subfigure}[t]{0.23\textwidth}
    \includegraphics[width=\textwidth]{figs/cls_llm_subj_vicuna-7b_acc_remove.pdf}
    \caption{Subj using column position}
    \end{subfigure}
\hfill
\begin{subfigure}[t]{0.23\textwidth}
    \includegraphics[width=\textwidth]{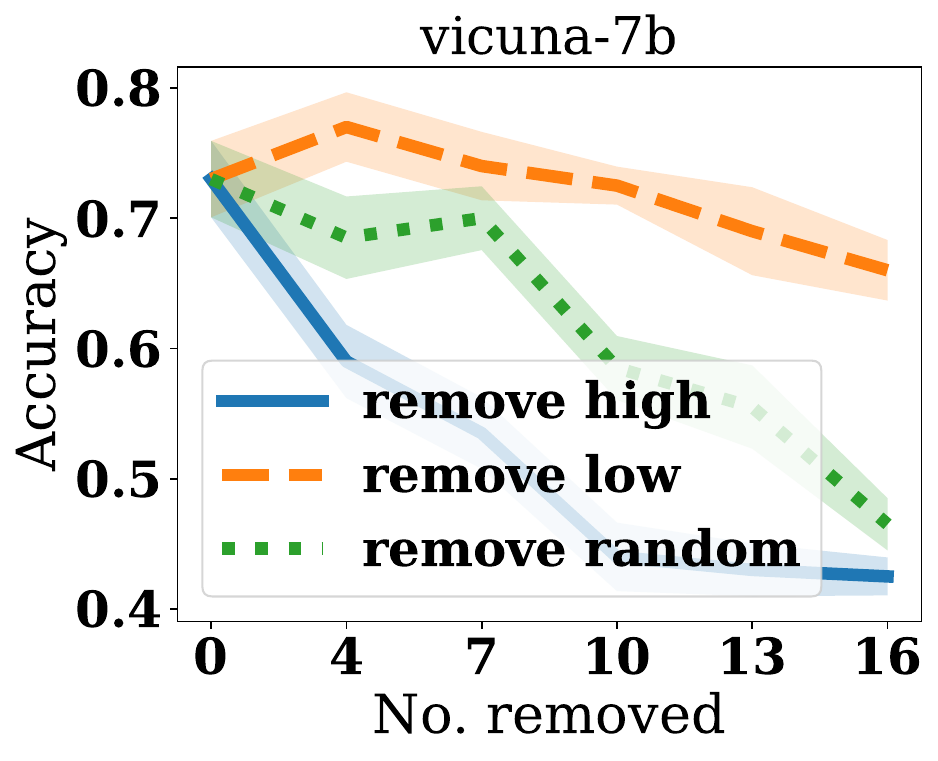}
    \caption{SST-2 using label position}
    \end{subfigure}
\hfill
\begin{subfigure}[t]{0.23\textwidth}
    \includegraphics[width=\textwidth]{figs/cls_llm_sst2_vicuna-7b_acc_remove.pdf}
    \caption{SST-2 using column position}
    \end{subfigure}
\hfill
\caption{Results of model prediction accuracy vs.~number of demonstrations removed using different positions for taking embeddings. $\lambda=1.0$. Lines and shades represent the mean and standard error respectively.}
\label{fig:llm_label_pos}
\end{figure}

As a rule of thumb, for each demonstration, we generally want to take the embedding of its last few tokens because of the causal nature of inference and because information generally flows toward the end of the sequence~\citep{wang-etal-2023-label}. We compare two possible choices of target position: the column position (immediately before the label) and the label position. We experiment on the demonstration removal task with these two choices of embeddings. The results are shown in \cref{fig:llm_label_pos}. Using embeddings of both positions achieves decent task performance as reflected by the clear distinction in accuracy between removing demonstrations with high/low \texttt{DETAIL} scores, demonstrating that our method is robust against the choice of token embeddings. In our experiments, we adopt the column position to isolate information about the label from the embedding.

\subsection{Experiment on State-Space Model Architecture} \label{app:exp_ssm}

We consider experiments on a popular state-space model (SSM) architecture, a Mamba-2.8b model~\citep{gu2023mamba}, which consists of $64$ layers with $d=2560$. The tasks are described in the main text in detail. While we find that \texttt{DETAIL} can still successfully attribute Mamba on certain tasks and datasets, the performance is inconsistent. One hypothesis is that the largest currently available Mamba model (2.8B) is still significantly smaller than the 7B LLMs we conduct experiments on in the main text. A smaller model size reduces the inductive power of the model to formulate the ``internal optimizer'', leading to less interpretive \text{DETAIL} scores. We would also like to note that \texttt{DETAIL} is not designed to work on SSMs.

\begin{figure}[H]
\centering
\begin{subfigure}[t]{0.23\textwidth}
    \includegraphics[width=\textwidth]{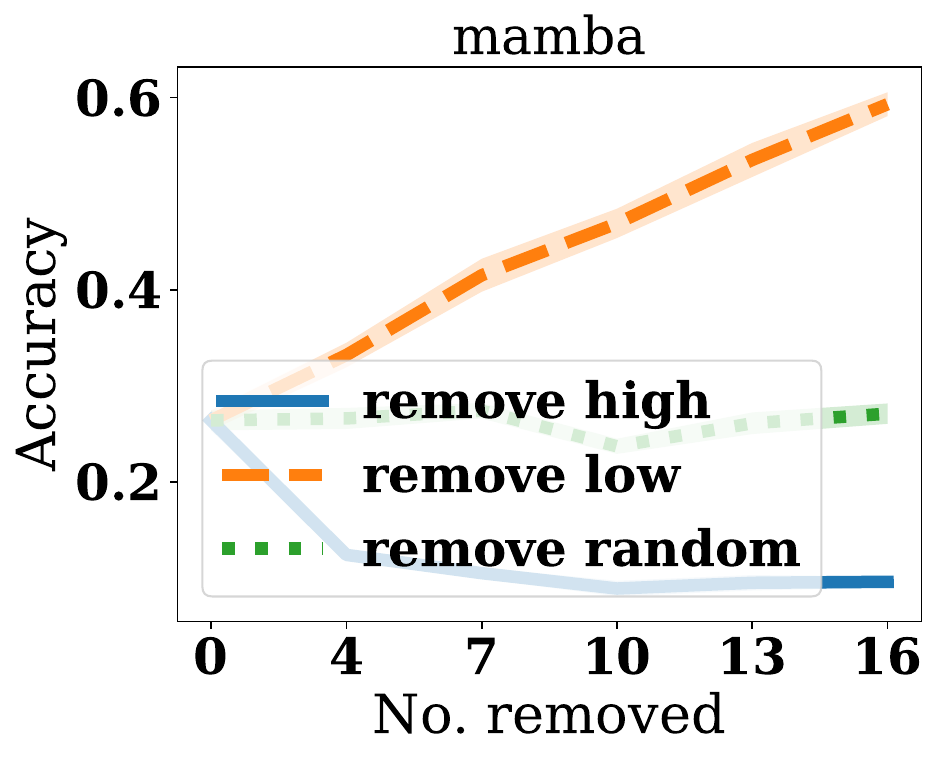}
    \caption{AG News removal}
    \end{subfigure}
\hfill
\begin{subfigure}[t]{0.23\textwidth}
    \includegraphics[width=\textwidth]{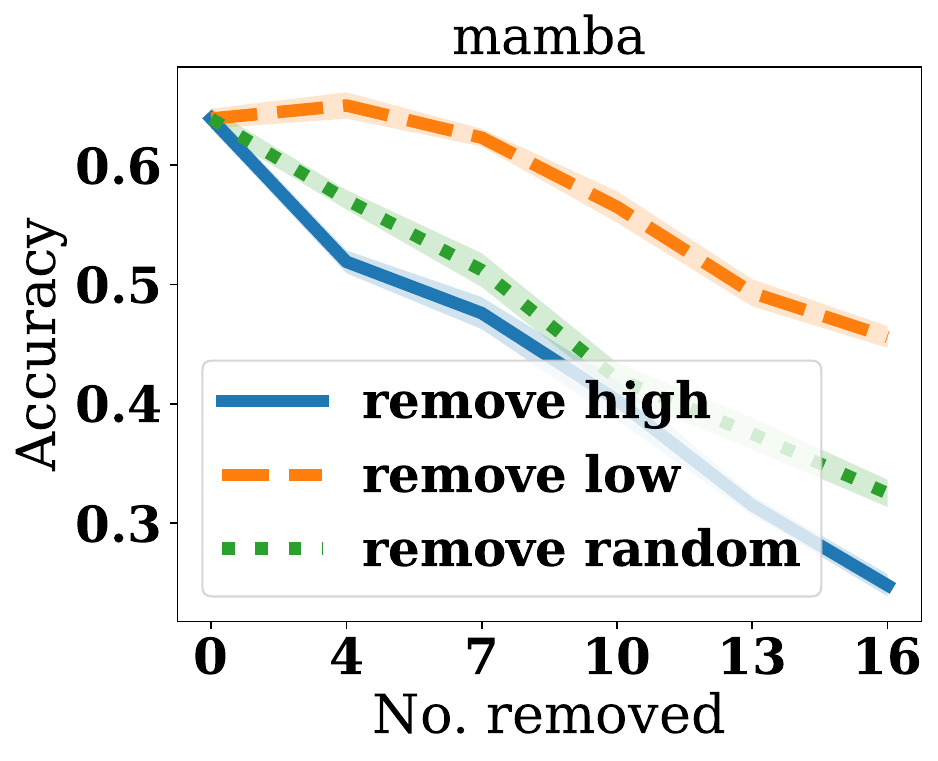}
    \caption{SST-2 removal}
    \end{subfigure}
\hfill
\begin{subfigure}[t]{0.23\textwidth}
    \includegraphics[width=\textwidth]{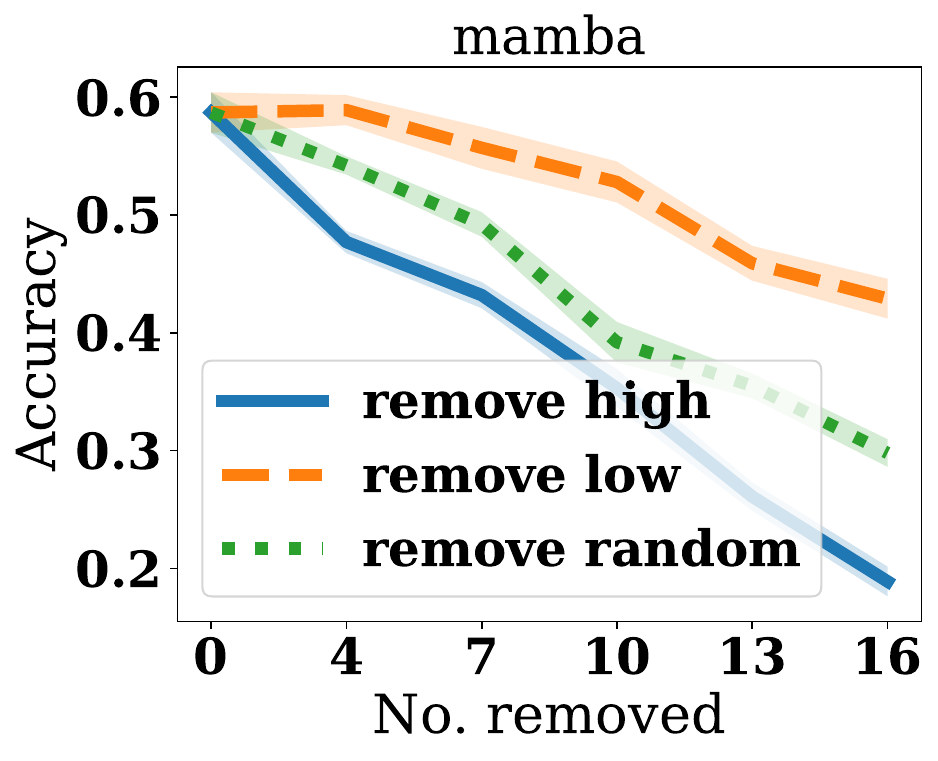}
    \caption{Rotten Tomatoes removal}
    \end{subfigure}
\hfill
\begin{subfigure}[t]{0.23\textwidth}
    \includegraphics[width=\textwidth]{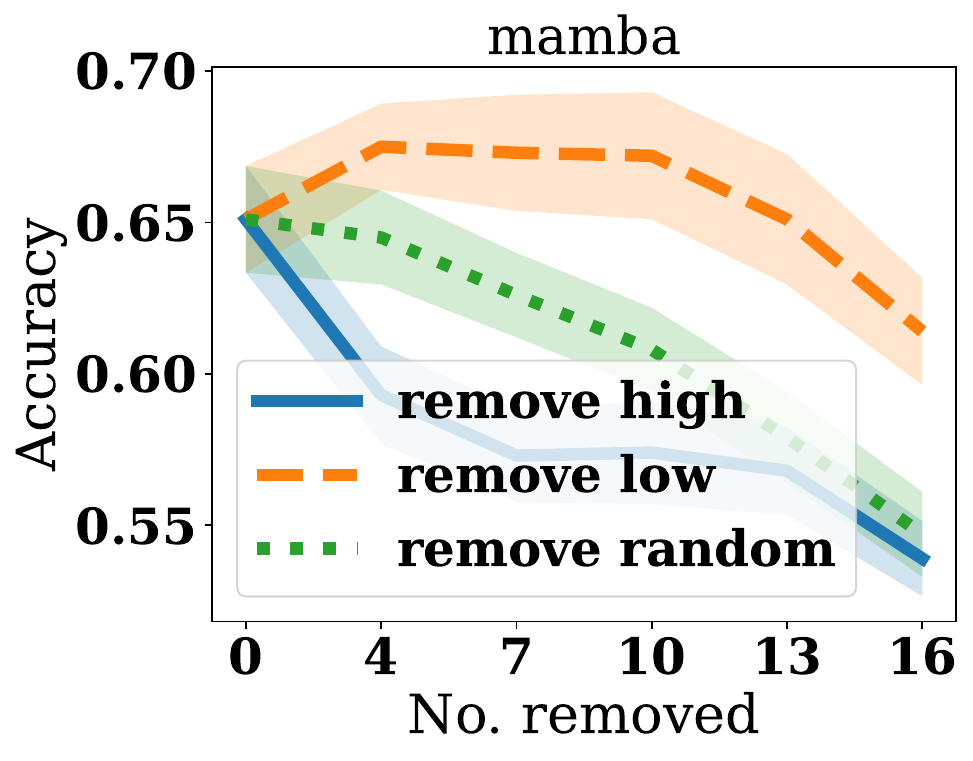}
    \caption{Subj removal}
    \end{subfigure}
\hfill
\caption{Results of model prediction accuracy vs.~number of demonstrations removed on Mamba. $\lambda=1.0$. Lines and shades represent the mean and standard error respectively.}
\label{fig:ssm-cls}
\end{figure}

\paragraph{Demonstration removal.} The demonstration removal experiment follows the same setup as \cref{sec:eval_llm} but uses a Mamba-2.8b model instead, which is the largest model officially open-sourced. The results are shown in \cref{fig:ssm-cls}. Interestingly, removing demonstrations according to \texttt{DETAIL} scores still can influence predictive performance in the desirable manner where removing demonstrations with high $\cI_{\text{test}}$ leads to lower accuracy and \textit{vice versa}.

\begin{figure}[H]
\centering
\begin{subfigure}[t]{0.19\textwidth}
    \includegraphics[width=\textwidth]{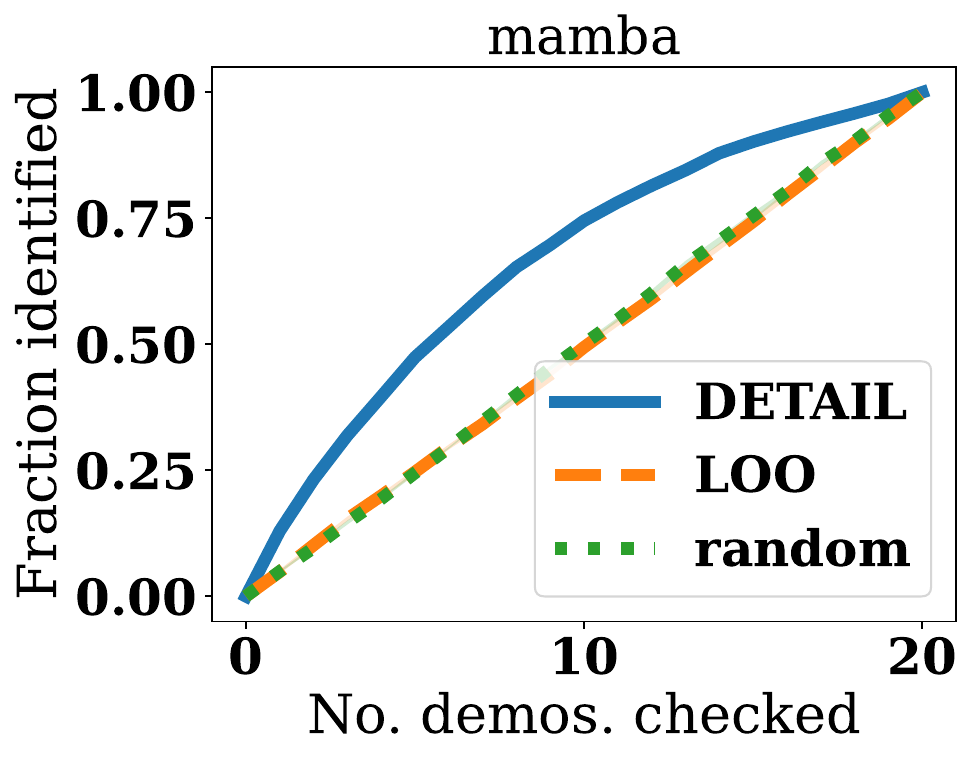}
    \caption{Detecting noisy label on AG News}
    \end{subfigure}
\hfill
\begin{subfigure}[t]{0.19\textwidth}
    \includegraphics[width=\textwidth]{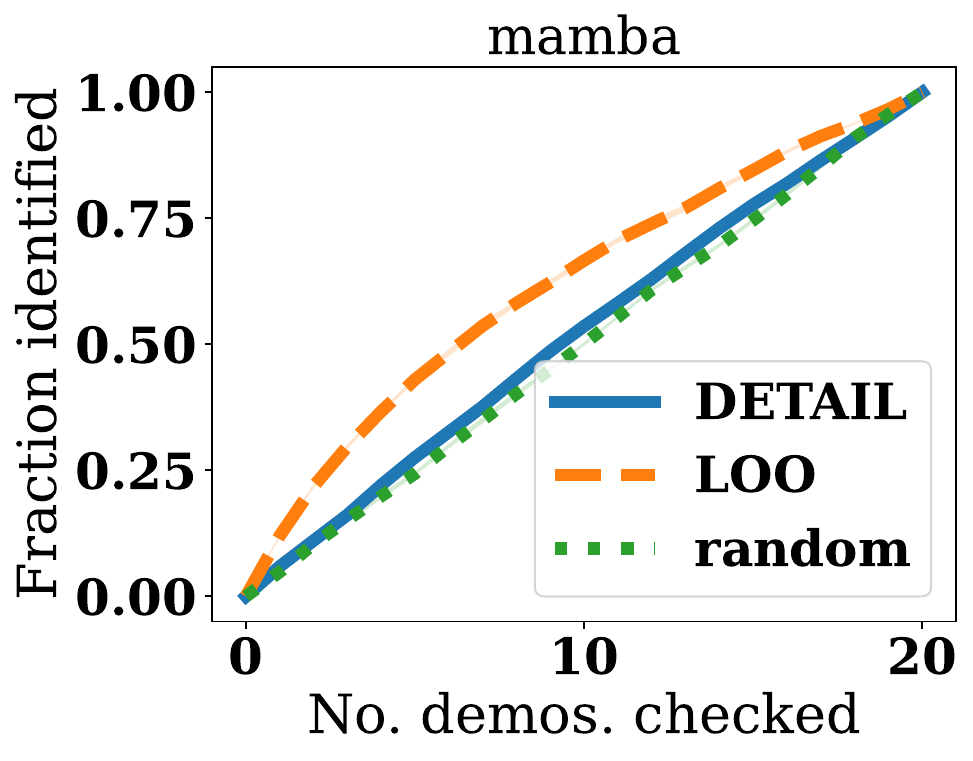}
    \caption{Detecting noisy label on SST-2}
    \end{subfigure}
\hfill
\begin{subfigure}[t]{0.19\textwidth}
    \includegraphics[width=\textwidth]{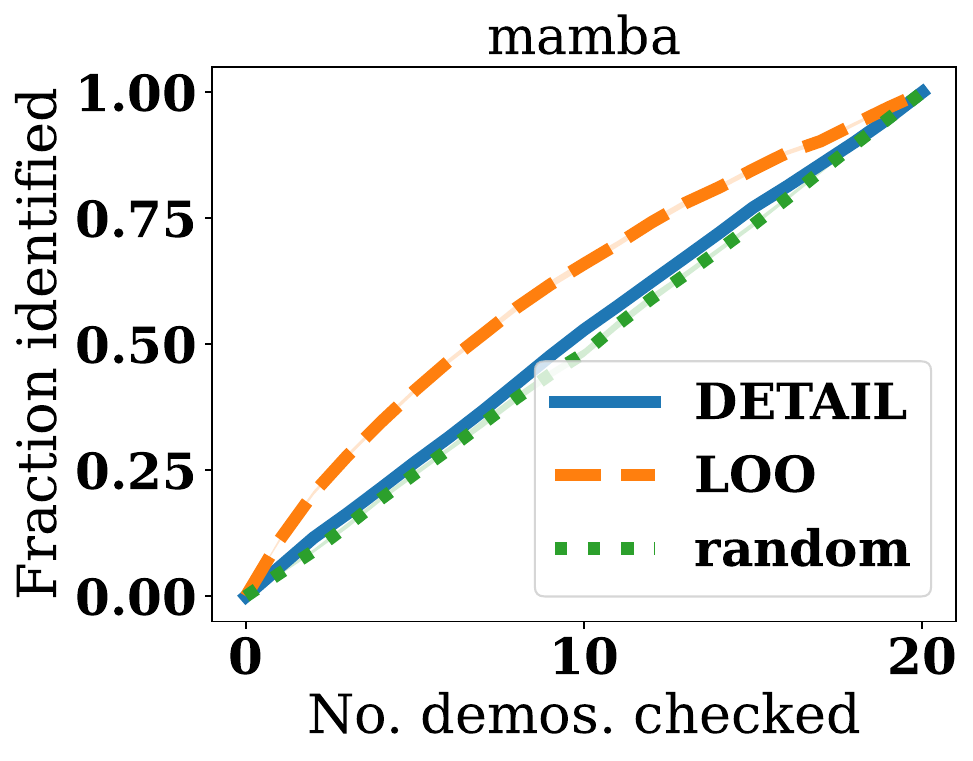}
    \caption{Detecting noisy label on Rotten Tomatoes}
    \end{subfigure}
\hfill
\begin{subfigure}[t]{0.19\textwidth}
    \includegraphics[width=\textwidth]{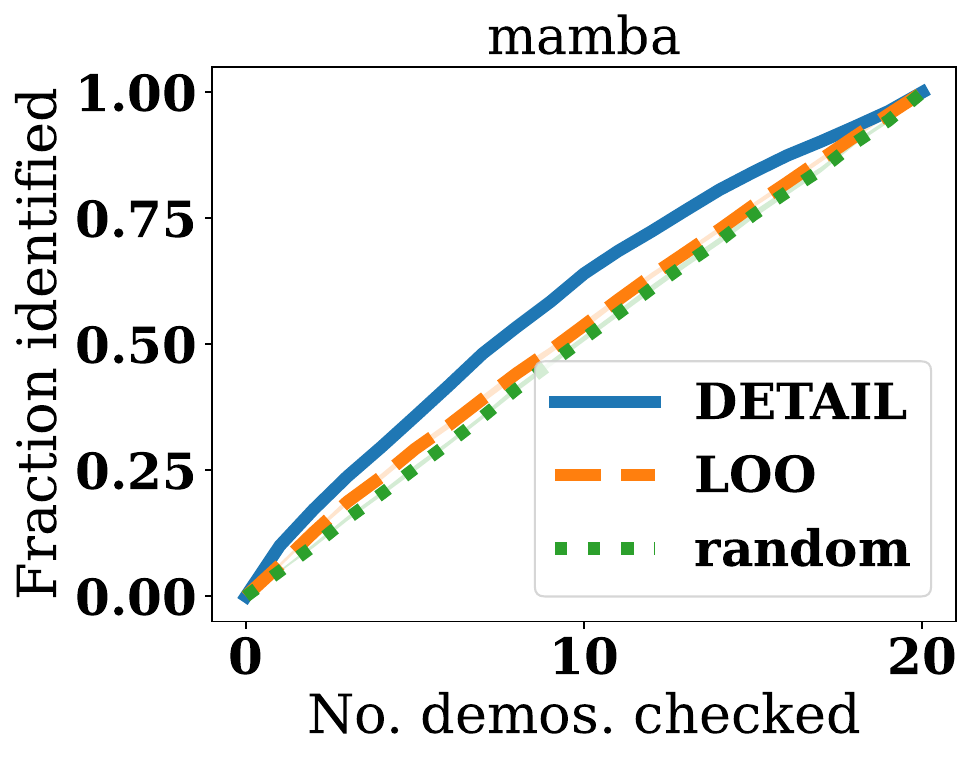}
    \caption{Detecting noisy label on Subj}
    \end{subfigure}
\hfill
\begin{subfigure}[t]{0.19\textwidth}
    \includegraphics[width=\textwidth]{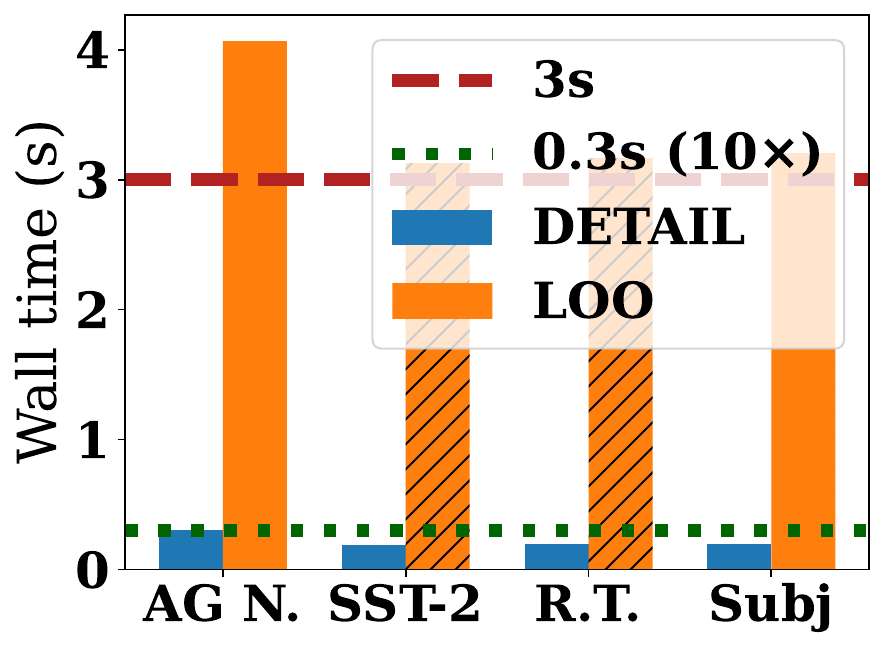}
    \caption{Wall time comparison}
    \end{subfigure}
\hfill
\caption{(a-d) Fraction of all noisy labels identified vs.~the number of demonstrations ranked by our method (with projection down to $1000$ dimension) and LOO checked respectively. (e) Wall time comparison across all datasets. $\lambda=10^{-9}$. All experiments are repeated with $10$ independent trials. Lines and shades represent the mean and standard error respectively.}
\label{fig:llm_noisy_label_detection_mamba}
\end{figure}

\paragraph{Noisy demonstration detection.} We further consider the noisy demonstration detection task as described in \cref{sec:eval_llm} on Mamba. Unfortunately, the performance is not consistent across datasets, as shown in \cref{fig:llm_noisy_label_detection_mamba}: detecting demonstrations with high $\cI_{\text{self}}$ performs close to random selection on SST-2 and Rotten Tomatoes datasets, although the inference speedup is still significant. We leave the analysis of these failure cases to future work.

\begin{figure}[H]
\centering
\begin{subfigure}[t]{0.23\textwidth}
    \includegraphics[width=\textwidth]{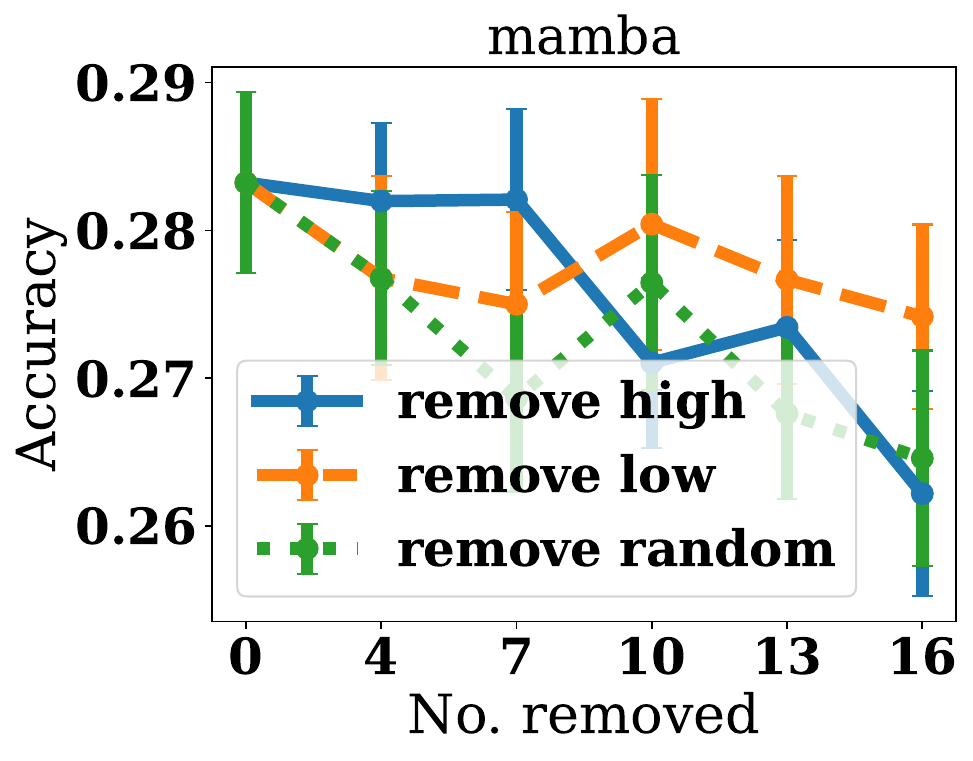}
    \end{subfigure}
\hfill
\begin{subfigure}[t]{0.23\textwidth}
    \includegraphics[width=\textwidth]{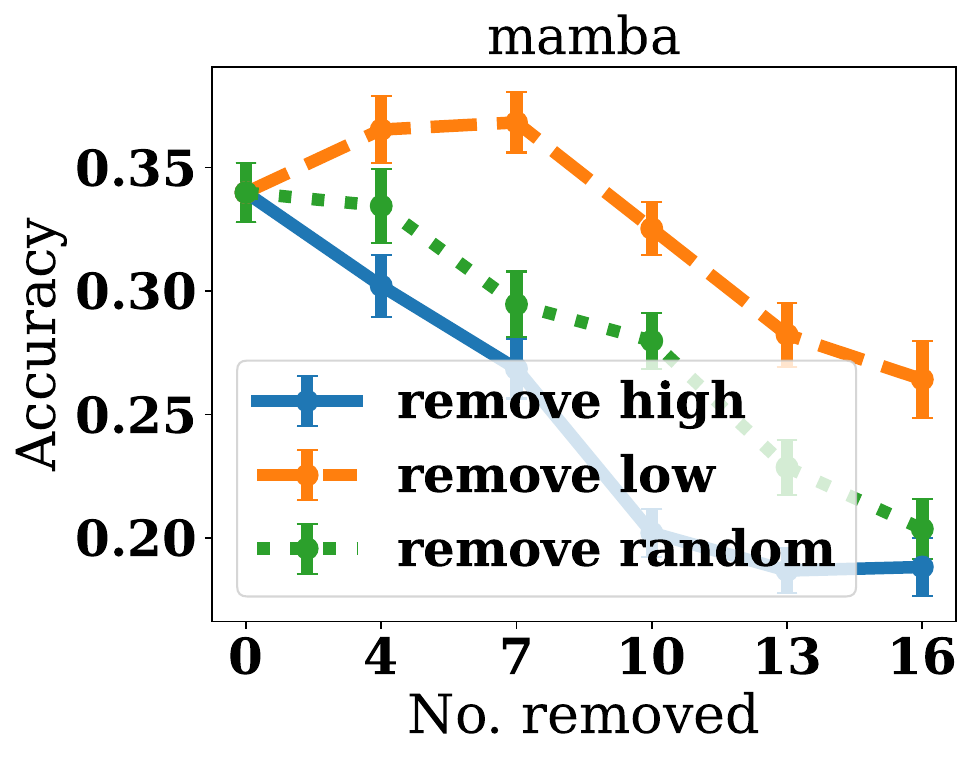}
    \end{subfigure}
\hfill
\begin{subfigure}[t]{0.23\textwidth}
    \includegraphics[width=\textwidth]{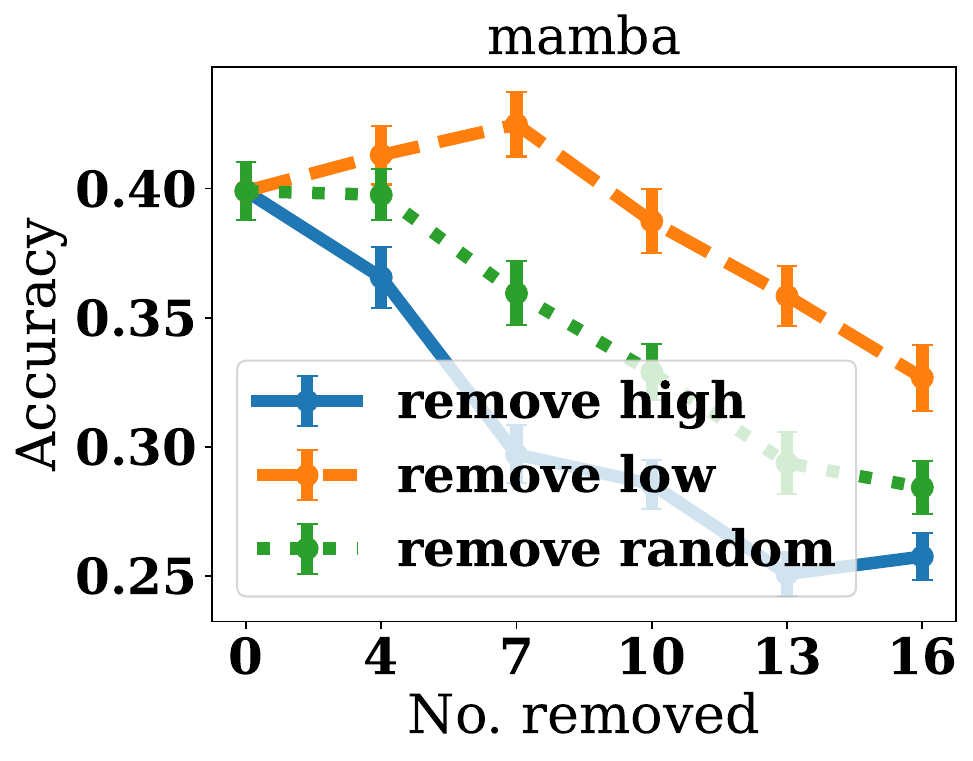}
    \end{subfigure}
\hfill
\begin{subfigure}[t]{0.23\textwidth}
    \includegraphics[width=\textwidth]{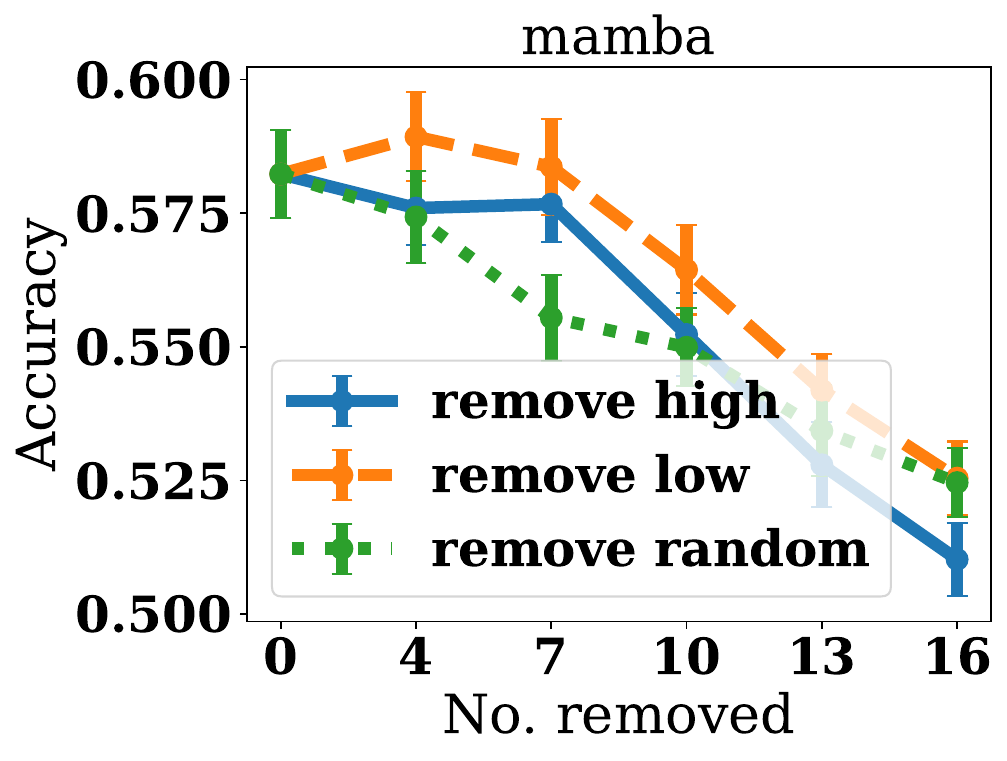}
    \end{subfigure}
\hfill
\caption{(Left to right) test accuracy vs.~number of demonstrations removed using $\cI_{\text{test}}$ on AG news, SST-2, Rotten Tomatoes, and Subj datasets using Mamba-2.8b. All experiments are repeated with $80$ independent trials. Lines and bars represent the mean and standard error respectively.}
\label{fig:icl_data_curation_mamba}
\end{figure}

\paragraph{Demonstration curation.} As \texttt{DETAIL} performs well using Mamba on the demonstration removal task, it is reasonable to hope that it works well on the demonstration curation task as well. As it turns out, \texttt{DETAIL} performs well on binary classification tasks as shown in \cref{fig:icl_data_curation_mamba} but performs poorly on AG News which is $4$-way classification. We hypothesize that this is due to Mamba's worse inductive power to formulate an internal algorithm successfully.

\clearpage
\newpage
\section*{NeurIPS Paper Checklist}

\begin{enumerate}

\item {\bf Claims}
    \item[] Question: Do the main claims made in the abstract and introduction accurately reflect the paper's contributions and scope?
    \item[] Answer: \answerYes{} 
    \item[] Justification: We propose an attribution method for demonstrations via the viewpoint of treating the transformer as implementing an internal algorithm and demonstrate its effectiveness. Our abstract and introduction accurately reflect this claim with elaboration.

\item {\bf Limitations}
    \item[] Question: Does the paper discuss the limitations of the work performed by the authors?
    \item[] Answer: \answerYes{} 
    \item[] Justification: A limitation of our work is discussed in \cref{sec:conclusion}.

\item {\bf Theory Assumptions and Proofs}
    \item[] Question: For each theoretical result, does the paper provide the full set of assumptions and a complete (and correct) proof?
    \item[] Answer: \answerNA{} 
    \item[] Justification: We demonstrate the effectiveness of our method primarily by conducting experiments under different scenarios. There are theoretical insights that lead to our formulation, for which we have properly referenced.

    \item {\bf Experimental Result Reproducibility}
    \item[] Question: Does the paper fully disclose all the information needed to reproduce the main experimental results of the paper to the extent that it affects the main claims and/or conclusions of the paper (regardless of whether the code and data are provided or not)?
    \item[] Answer: \answerYes{} 
    \item[] Justification: We have included the specific details of our algorithmic implementation in \cref{app:algo} and an illustration in \cref{fig:illustration}. Python code for reproducibility is also included in the supplemental materials. The datasets used in the experiments are open-source, commonly used, and well-specified.

\item {\bf Open access to data and code}
    \item[] Question: Does the paper provide open access to the data and code, with sufficient instructions to faithfully reproduce the main experimental results, as described in supplemental material?
    \item[] Answer: \answerYes{} 
    \item[] Justification: Python code with the conda environment file is included in the supplemental materials with running instructions in the README file. Commands for reproducing experimental results are also included in a bash file. All datasets are freely downloadable with downloading code snippets written in the code.

\item {\bf Experimental Setting/Details}
    \item[] Question: Does the paper specify all the training and test details (e.g., data splits, hyperparameters, how they were chosen, type of optimizer, etc.) necessary to understand the results?
    \item[] Answer: \answerYes{} 
    \item[] Justification: Important experimental settings are specified in each subsection in \cref{sec:eval_custom_tf}, \cref{sec:eval_llm}, and \cref{sec:exp_application}. Additional details are provided in \cref{app:exp}.

\item {\bf Experiment Statistical Significance}
    \item[] Question: Does the paper report error bars suitably and correctly defined or other appropriate information about the statistical significance of the experiments?
    \item[] Answer: \answerYes{} 
    \item[] Justification: Average and standard errors are included in the experimental results either in shades (for figures) or in brackets (for tables).

\item {\bf Experiments Compute Resources}
    \item[] Question: For each experiment, does the paper provide sufficient information on the computer resources (type of compute workers, memory, time of execution) needed to reproduce the experiments?
    \item[] Answer: \answerYes{} 
    \item[] Justification: Computing hardware (in terms of GPU) is specified in \cref{app:compute}. Wall time for our approach on different tasks (and the corresponding GPU) is provided in \cref{sec:eval_llm} and \cref{sec:exp_application}.
    
\item {\bf Code Of Ethics}
    \item[] Question: Does the research conducted in the paper conform, in every respect, with the NeurIPS Code of Ethics \url{https://neurips.cc/public/EthicsGuidelines}?
    \item[] Answer: \answerYes{} 

\item {\bf Broader Impacts}
    \item[] Question: Does the paper discuss both potential positive societal impacts and negative societal impacts of the work performed?
    \item[] Answer: \answerYes{} 
    \item[] Justification: We discuss the need for interpretability in in-context learning which our work addresses in the main text. Additional discussion is in \cref{app:discussion}.
    
\item {\bf Safeguards}
    \item[] Question: Does the paper describe safeguards that have been put in place for responsible release of data or models that have a high risk for misuse (e.g., pre-trained language models, image generators, or scraped datasets)?
    \item[] Answer: \answerNA{} 
    \item[] Justification: Our contribution is mainly improving the interpretability of in-context learning. Data and models used are freely and readily available online.

\item {\bf Licenses for existing assets}
    \item[] Question: Are the creators or original owners of assets (e.g., code, data, models), used in the paper, properly credited and are the license and terms of use explicitly mentioned and properly respected?
    \item[] Answer: \answerYes{} 
    \item[] Justification: The datasets used in our work are all properly cited.

\item {\bf New Assets}
    \item[] Question: Are new assets introduced in the paper well documented and is the documentation provided alongside the assets?
    \item[] Answer: \answerNA{} 

\item {\bf Crowdsourcing and Research with Human Subjects}
    \item[] Question: For crowdsourcing experiments and research with human subjects, does the paper include the full text of instructions given to participants and screenshots, if applicable, as well as details about compensation (if any)? 
    \item[] Answer: \answerNA{} 

\item {\bf Institutional Review Board (IRB) Approvals or Equivalent for Research with Human Subjects}
    \item[] Question: Does the paper describe potential risks incurred by study participants, whether such risks were disclosed to the subjects, and whether Institutional Review Board (IRB) approvals (or an equivalent approval/review based on the requirements of your country or institution) were obtained?
    \item[] Answer: \answerNA{} 
\end{enumerate}

\end{document}